\documentclass[runningheads]{llncs}

\usepackage[mobile]{eccv}

\usepackage{eccvabbrv}

\usepackage{graphicx}
\usepackage{booktabs}
\usepackage{array}
\newcolumntype{M}[1]{>{\centering\arraybackslash}m{#1}}
\newcolumntype{C}[1]{>{\centering\arraybackslash}m{#1}} %
\usepackage[export]{adjustbox}

\usepackage[accsupp]{axessibility}  %

\definecolor{lavenderrow}{RGB}{229,229,253}
\definecolor{lightgrayrow}{RGB}{240,240,240}
\definecolor{darkgrayrow}{RGB}{200,200,200}
\usepackage[table]{xcolor}
\definecolor{psnrgray}{gray}{0.80} %
\definecolor{psnrtext}{gray}{0.35} %

\usepackage{hyperref}

\usepackage{orcidlink}

\begin{document}

\raggedbottom
\title{SyncFix: Fixing 3D Reconstructions via Multi-View Synchronization
} 

\titlerunning{SyncFix}

\author{Deming Li\inst{1} \quad\quad
Abhay Yadav\inst{1} \quad\quad
Cheng Peng\inst{2} \\\vspace{3pt}
Rama Chellappa\inst{1} \quad\quad
Anand Bhattad\inst{1}}
\authorrunning{Deming Li et al.}

\renewcommand{\inst}[1]{$^{#1}$}

\institute{\inst{1}Johns Hopkins University \quad\quad
\inst{2}University of Virginia \\[5pt]
\tt{Project page: \href{https://syncfix.github.io/}{https://syncfix.github.io/}}
}

\maketitle

\newcommand{\cmpver}{compare_teaser}

\noindent\makebox[\textwidth][c]{%
\begin{minipage}{\textwidth}
\centering
    \setlength{\tabcolsep}{0.5pt}
    \renewcommand{\arraystretch}{1.0}
    \small
\vspace{5pt}
    \newlength{\imgH}
    \settoheight{\imgH}{\includegraphics[width=0.235\linewidth]{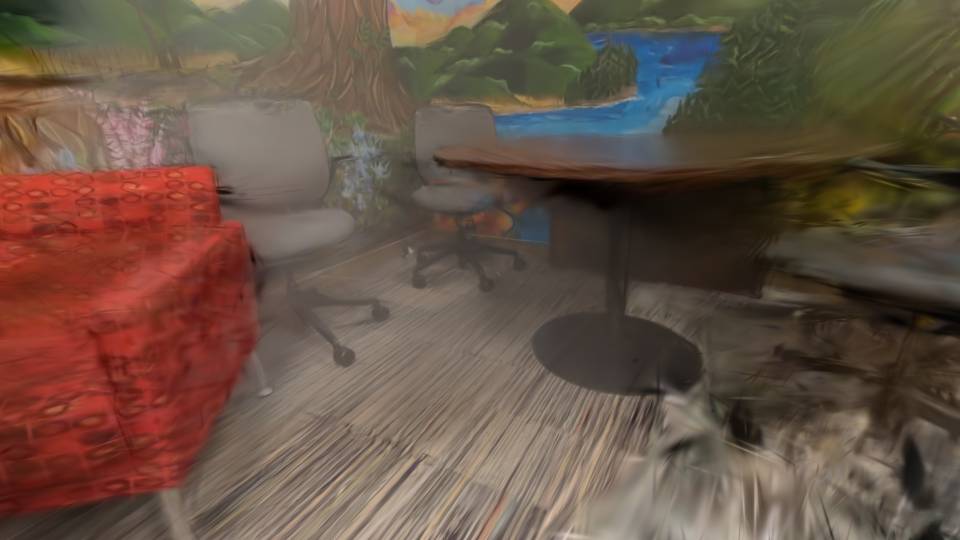}}

    \newcommand{\viewlabel}[1]{%
      \parbox[b][\imgH][c]{2.2em}{\centering\rotatebox{90}{\text{#1}}}%
    }

    \newcommand{\teaserannot}[2]{%
    \includegraphics[width=0.235\linewidth]{figures/teaser_abhay_arxiv_v2/#1/#2_\cmpver/#2_\cmpver_annotated.png}%
    }

    \newcommand{\teasercropone}[2]{%
    \includegraphics[width=0.235\linewidth]{figures/teaser_abhay_arxiv_v2/#1/#2_\cmpver/#2_\cmpver_crop1_boxed.png}%
    }
    \begin{tabular}{c c c c c}

    \viewlabel{\small View 1} &
    \includegraphics[width=0.235\linewidth]{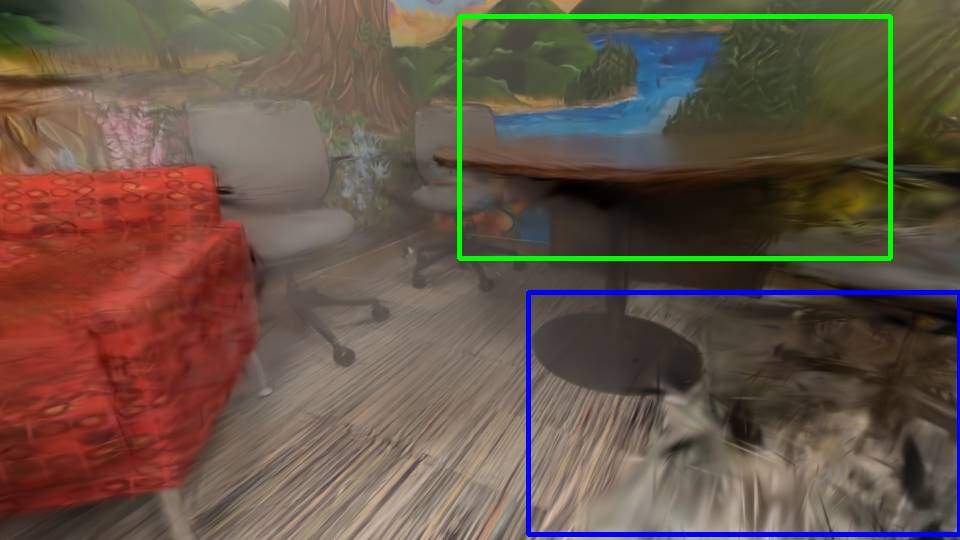} &
    \includegraphics[width=0.235\linewidth]{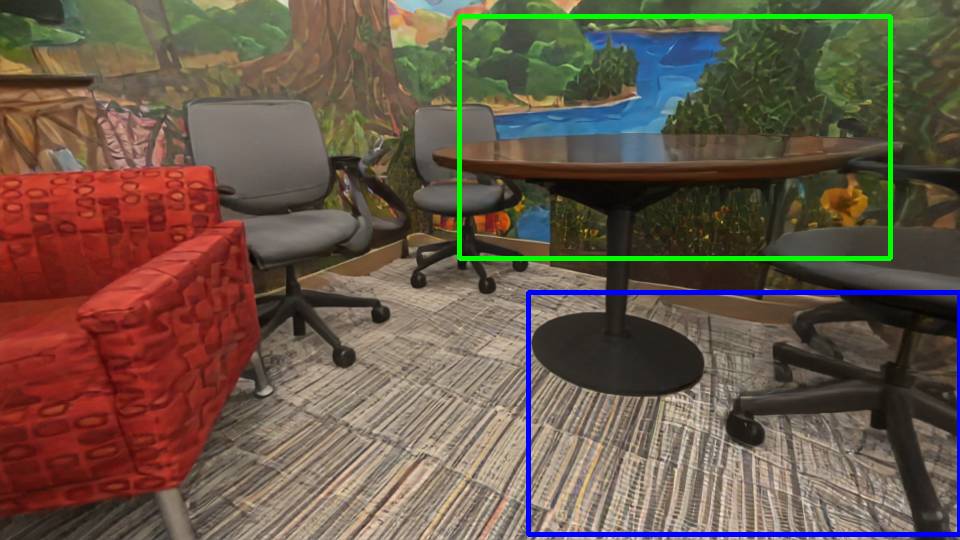} &
    \includegraphics[width=0.235\linewidth]{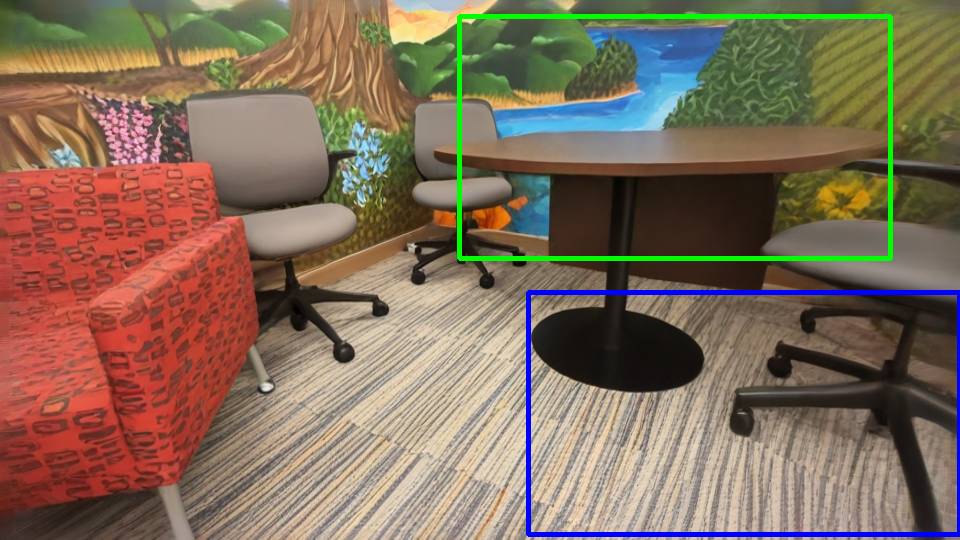} &
    \includegraphics[width=0.235\linewidth]{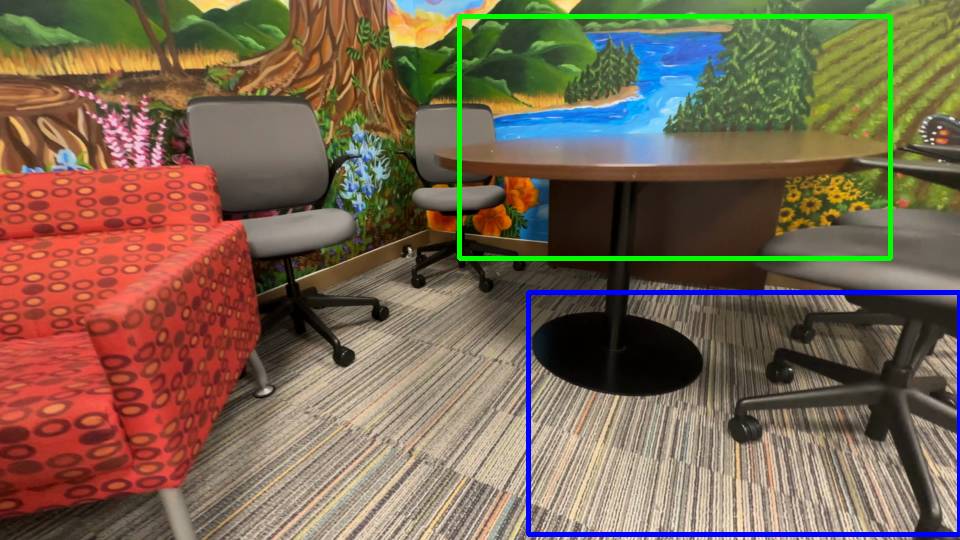} \\

    \viewlabel{\small View 2} &
    \includegraphics[width=0.235\linewidth]{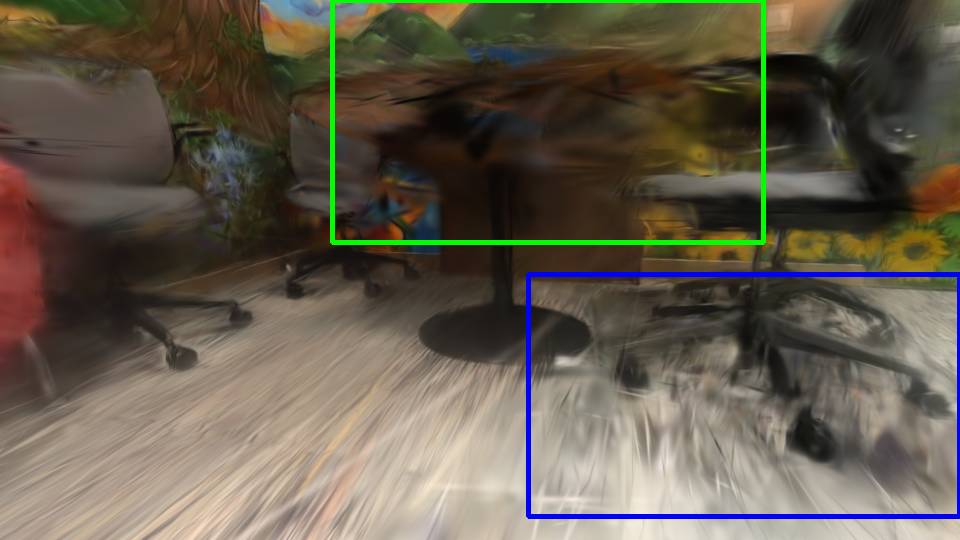} &
    \includegraphics[width=0.235\linewidth]{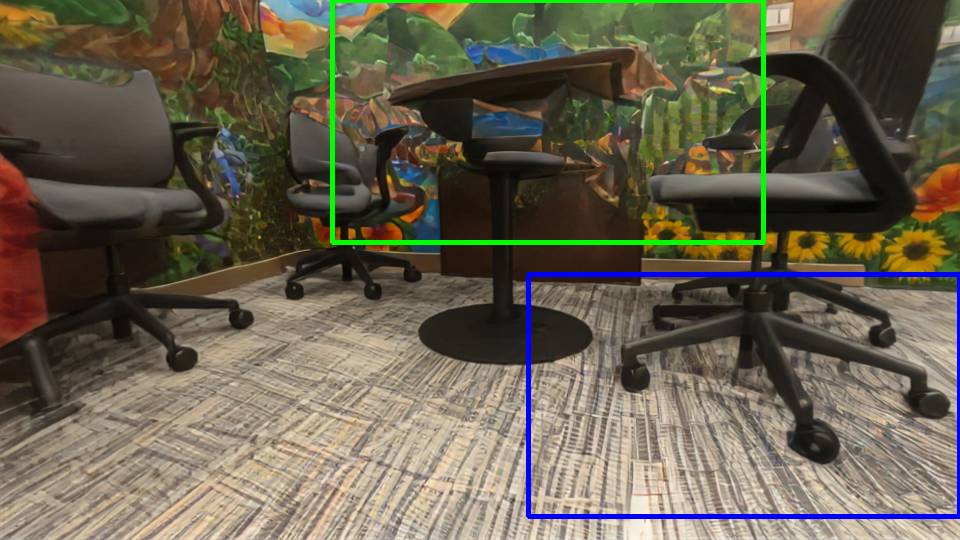} &
    \includegraphics[width=0.235\linewidth]{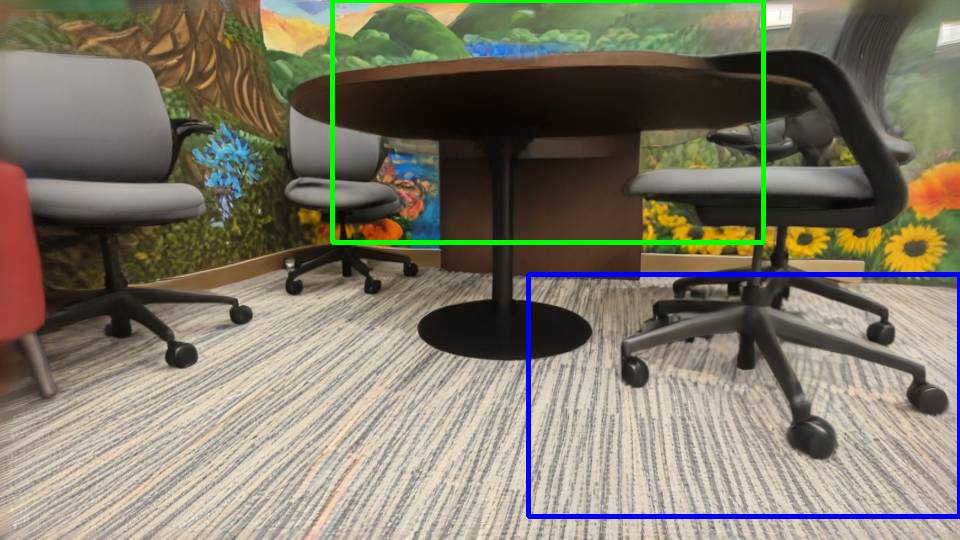} &
    \includegraphics[width=0.235\linewidth]{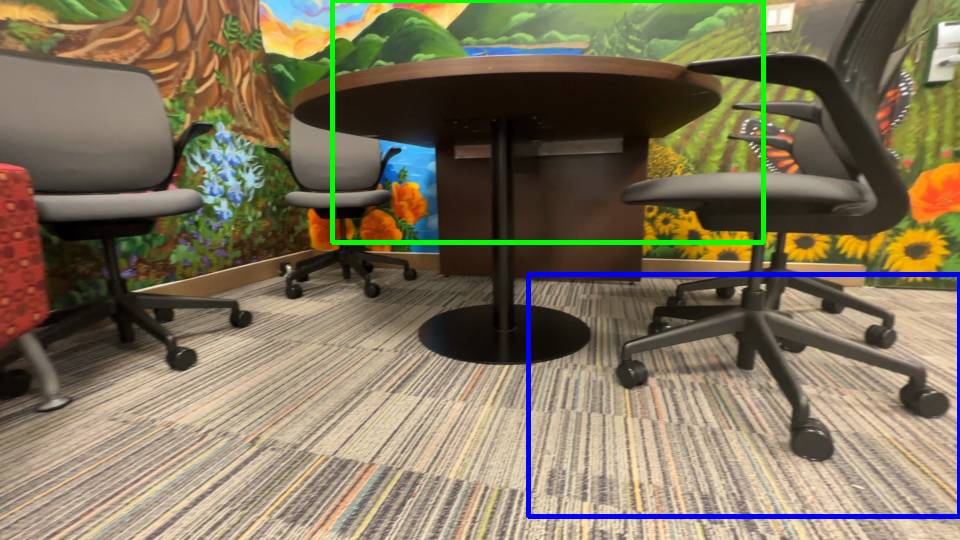} \\

    \viewlabel{\small View 3} &
    \includegraphics[width=0.235\linewidth]{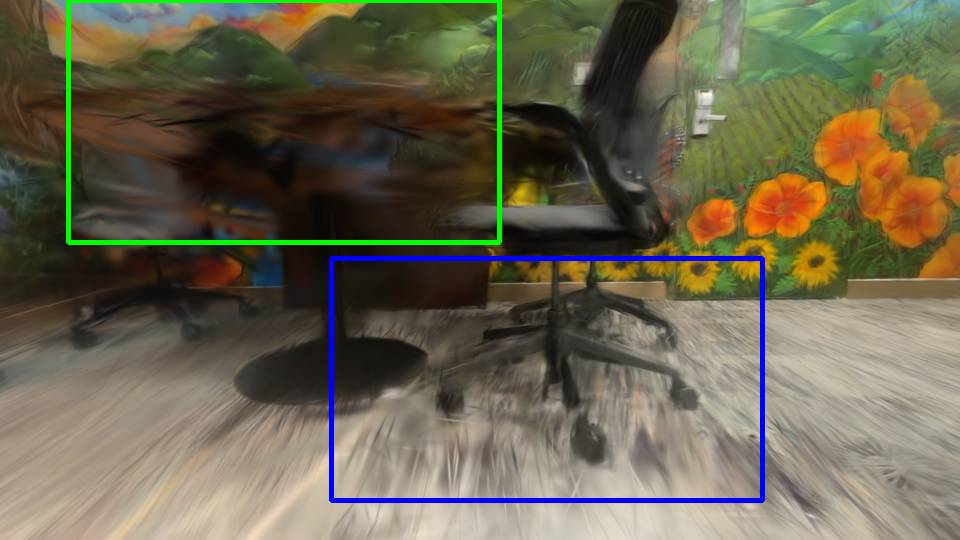} &
    \includegraphics[width=0.235\linewidth]{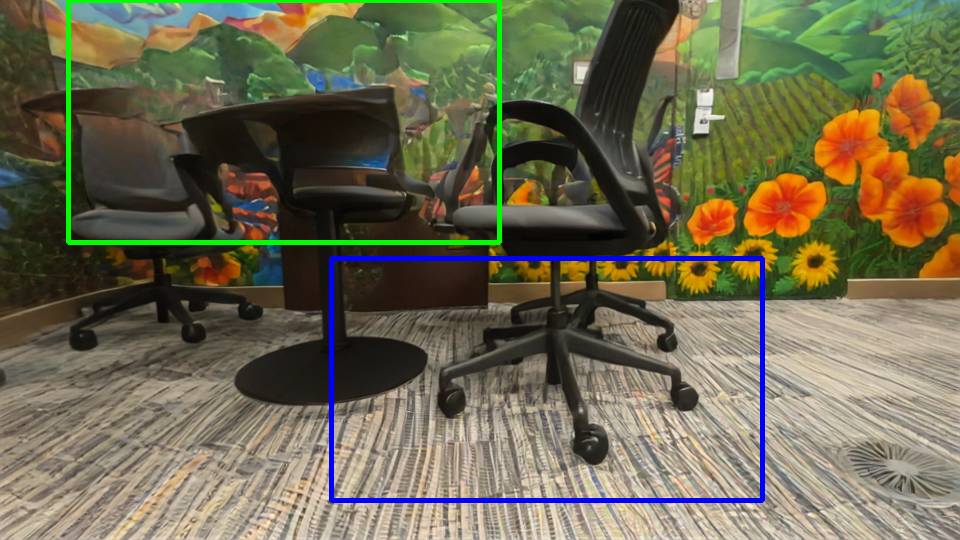} &
    \includegraphics[width=0.235\linewidth]{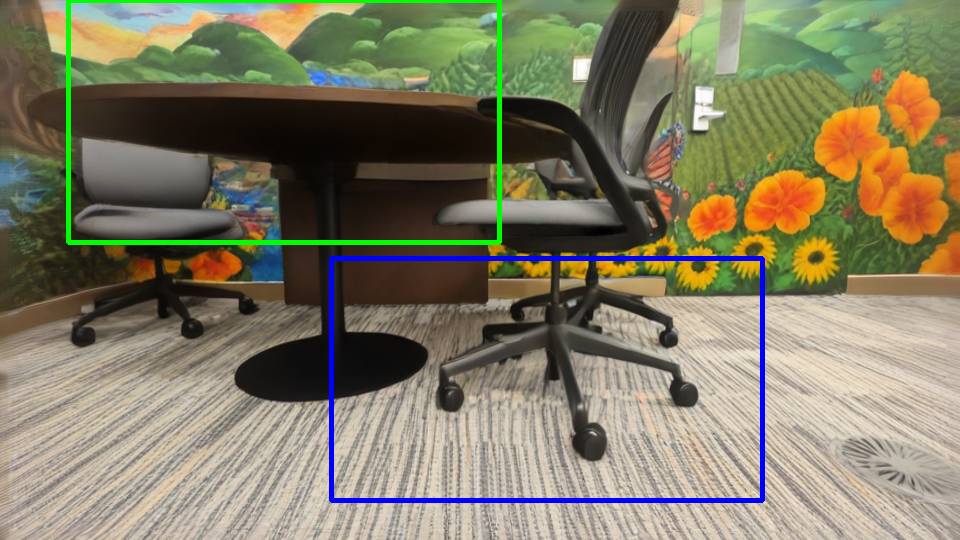} &
    \includegraphics[width=0.235\linewidth]{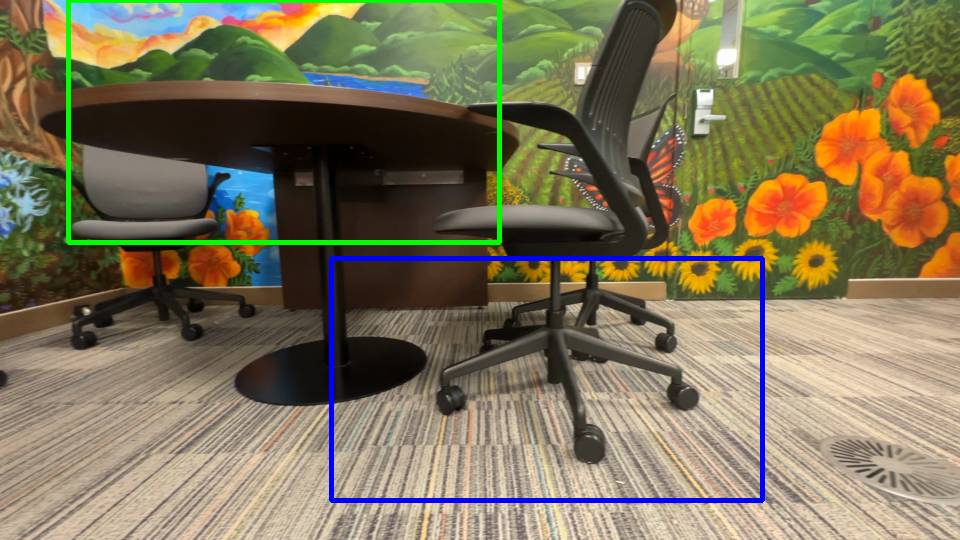} \\
\midrule
\multicolumn{5}{c}{} \\[-10pt]

    \viewlabel{\small View 2} &
    \includegraphics[width=0.235\linewidth]{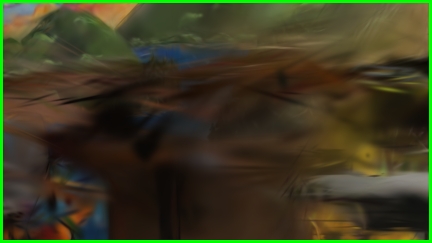} &
    \includegraphics[width=0.235\linewidth]{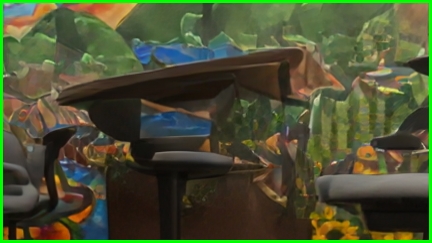} &
    \includegraphics[width=0.235\linewidth]{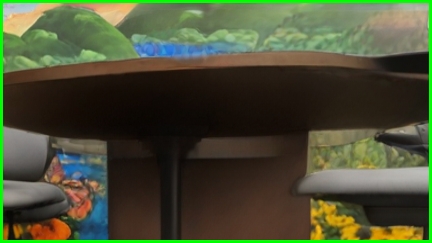} &
    \includegraphics[width=0.235\linewidth]{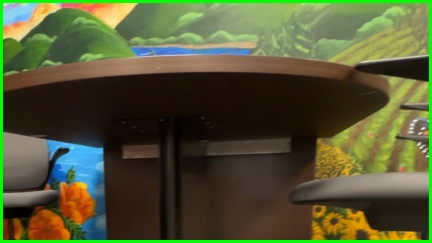} \\

    \viewlabel{\small View 3} &
    \includegraphics[width=0.235\linewidth]{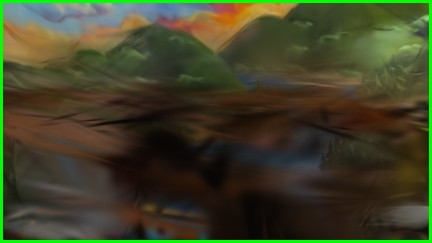} &
    \includegraphics[width=0.235\linewidth]{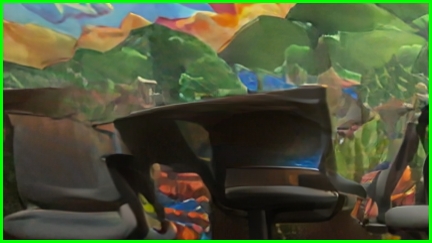} &
    \includegraphics[width=0.235\linewidth]{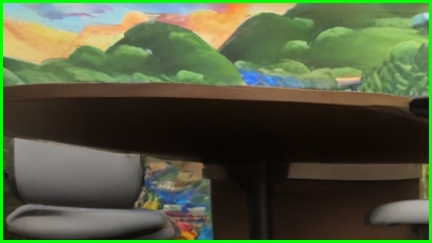} &
    \includegraphics[width=0.235\linewidth]{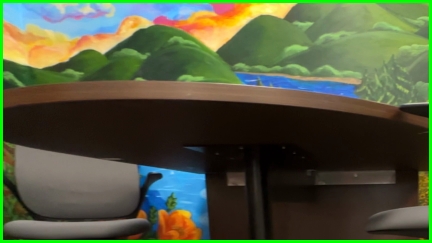} \\

        & {\footnotesize 3DGS} &{\footnotesize Difix3D+} & {\footnotesize Ours} & {\footnotesize GT}
    \end{tabular}
    \vspace{-8pt}
\captionof{figure}{Independent diffusion refinement (DifiX3D+) processes each view separately, leading to inconsistent geometry across views (see close-view of table in the bottom panel and highlighted regions above). SyncFix (Ours) instead refines all views jointly, enforcing cross-view agreement during denoising and producing stable 3D structure.
\vspace{-5pt}
}
\label{fig:teaser_figure}
\end{minipage}
}

\begin{abstract}

We present \textbf{SyncFix}, a framework that enforces cross-view consistency during the diffusion-based refinement of reconstructed scenes. SyncFix formulates refinement as a joint latent bridge matching problem, synchronizing distorted and clean representations across multiple views to fix the semantic and geometric inconsistencies. This means SyncFix learns a joint conditional over multiple views to enforce consistency throughout the denoising trajectory. Our training is done only on image pairs, but it generalizes naturally to an arbitrary number of views during inference. Moreover, reconstruction quality improves with additional views, with diminishing returns at higher view counts. Qualitative and quantitative results demonstrate that SyncFix consistently generates high-quality reconstructions and surpasses current state-of-the-art baselines, even in the absence of clean reference images. SyncFix achieves even higher fidelity when sparse references are available.
 \vspace{-5pt}
\end{abstract}

\section{Introduction}
\label{sec:intro}

Radiance fields \cite{mildenhall2020nerf} and 3D Gaussian splatting \cite{kerbl20233d} under sparse-view or off-trajectory settings often produce scenes plagued by floaters, texture artifacts, and geometric distortions. Recent approaches attempt to mitigate these issues by leveraging the rich priors of 2D generative models \cite{Nerfbusters2023, wu2025difix3d+}. Typically, these methods render novel views, refine each image using a diffusion model conditioned on a reference view, and distill the refined outputs back into the 3D representation.

However, these refinement methods treat each view as an independent 2D generation task. This ignores important cross-view geometric and semantic constraints and that all views are projections of the same underlying 3D scene. Therefore, independent refinement can produce inconsistent results for ambiguous structures or regions with large artifacts across different viewpoints. When these conflicting signals are distilled back into the 3D model, optimization becomes unstable, leading to appearance drift and degraded geometry.

In this paper, we impose multi-view consistency directly during the refinement process rather than correcting it post hoc. To this end, we introduce \textbf{SyncFix}, a latent bridge matching formulation that models refinement as joint conditional generation over multiple views. SyncFix aligns latent representations across views during the denoising process and encourages information exchange via cross-view attention without any explicit camera information. Each view is therefore refined in the context of all others, ensuring consistent semantics and geometry prior to 3D distillation.

During training, we train to refine only two views simultaneously, but its permutation-invariant latent interactions using attention naturally extend seamlessly to arbitrary numbers of views at inference. As more views are provided, the cross-view constraints become stricter and the solution space narrows, which leads to systematically better final reconstructions. We also find that SyncFix produces consistent multi-view outputs even when no clean reference frames are available, and it can further benefit from sparse references when they exist. We substantiate these claims using perceptual metrics and through analyses of geometric alignment and semantic correspondence.

\section{Related Work}
\label{sec:related}

\textbf{Sparse-View 3D Reconstruction.}
NeRF~\cite{mildenhall2020nerf} and 3DGS~\cite{kerbl20233d} produce high-fidelity geometry from dense captures. With sparse inputs the problem becomes ill-posed: reconstructions develop floaters, fragmented surfaces, and blurred textures. Regularization-based approaches constrain frequency content~\cite{niemeyer2021regnerf,yang2023freenerf,wang2023sparsenerf} or inject depth priors~\cite{deng2022depth, xu2025depthsplat}. Feed-forward Gaussian splatting methods~\cite{jiang2025anysplat,ye2026yonosplat} leverage learned multi-view encoders~\cite{wang2025vggt} for direct prediction. These methods reduce artifacts but do not eliminate them; the residual degradation motivates generative refinement.

\vspace{10pt}
\noindent\textbf{Per-View Refinement with 2D Priors.}
The dominant repair paradigm renders novel views, refines each independently with a 2D diffusion model, and distills the results back into the 3D representation. This follows the Score Distillation Sampling framework~\cite{wang2023score,poole2023dreamfusion}, originally developed for text-to-3D generation~\cite{lin2023magic3d,wang2023prolificdreamer} and since adapted for reconstruction repair. Nerfbusters~\cite{Nerfbusters2023} applies diffusion guidance during NeRF optimization. ReconFusion~\cite{wu2024reconfusion} trains a multi-view conditioned prior for few-view regularization. Instruct-NeRF2NeRF~\cite{instructnerf2023} uses SDEdit~\cite{meng2022sdedit} on individual renderings. Difix3D+~\cite{wu2025difix3d+} distills a single-step refinement filter and adds limited cross-view conditioning through reference-view cross-attention. FreeFix~\cite{zhou2026freefix} manipulates latent trajectories without retraining. These methods ultimately optimize per-view reconstruction objectives, even when limited cross-view signals are introduced. The 2D prior never sees multiple views simultaneously, so it may interpret the same artifact differently from each angle. The 3D distillation stage must reconcile the resulting conflicts.

\vspace{10pt}
\noindent\textbf{Coupling Views inside the Generator.}
A natural fix is to condition generation on multiple views at once. Video diffusion models provide implicit coupling through temporal attention: 3DGS-Enhancer~\cite{liu2024dgsenhancer} casts view consistency as temporal coherence and refines 3DGS with a confidence-aware loop; ReconX~\cite{liu2024reconx} conditions video diffusion on a global point cloud for sparse-view reconstruction. Multi-view diffusion architectures couple views more directly. MVDream~\cite{shi2024mvdream}, SyncDreamer~\cite{liu2024syncdreamer}, CAT3D~\cite{gao2024cat3d}, and ViewCrafter~\cite{Yu2024ViewCrafter} synchronize attention across generated views. 3DEnhancer~\cite{luo20253denhancer} applies this idea to 3D enhancement, using multi-view row attention and epipolar aggregation within a latent diffusion model. These methods demonstrate that coupling views improves coherence. However, they remain within the denoising diffusion framework: views interact only through attention over noisy intermediate states, and sampling requires iterative denoising.

\vspace{10pt}
\noindent\textbf{Flow-Based Multi-View Refinement.}
Rectified Flow~\cite{liu2023flow} and Flow Matching~\cite{lipman2023flow} replace iterative denoising with learned ODEs that transport between distributions. Latent Bridge Matching (LBM)~\cite{chadebec2025lbm} extends this to conditional generation via deterministic bridges between paired latents. SyncLight~\cite{serrano2026synclight} adds cross-view attention to LBM for consistent relighting. SyncFix builds on this foundation but targets a different problem: repairing degraded 3D geometry. We learn a joint flow over the product latent space of multiple distorted renderings and their clean counterparts. Where attention-based diffusion models such as 3DGS-Enhancer couple views implicitly through shared noise states, SyncFix couples them with deterministic latent bridge matching, enforcing semantic and geometric agreement before any image is decoded. Thus, SyncFix is a single-pass, intrinsically consistent multi-view refinement without iterative denoising.

\section{Method}
\label{sec:method}

Our goal is to refine 3D reconstructions while preserving multi-view consistency. Existing approaches refine each rendered view independently and rely on the 3D representation to harmonize conflicting corrections. This harmonization is unreliable. For example, one view may introduce high-frequency structures (e.g., a painting on a wall) while another suppresses them, leading to inconsistent geometry and texture when fused back into the 3D representation. We instead model refinement as a \emph{joint conditional generation} problem over multiple views.

Let a distorted 3D reconstruction produce $N$ rendered views $X_D = \{x_D^{(1)}, \dots, \allowbreak x_D^{(N)}\}$, with corresponding clean targets $X_{GT} = \{x_{GT}^{(1)}, \dots, x_{GT}^{(N)}\}$. Marginal methods like Difix3D+ \cite{wu2025difix3d+} learn independent mappings $P(x_{GT}^{(i)} \mid x_D^{(i)})$. SyncFix instead models the joint conditional $P(X_{GT} \mid X_D)$ and enforces cross-view consensus during generation. This shift from marginal to joint modeling is central: corrections for each view are conditioned on all other views simultaneously.

\subsection{Data Generation and Distortion Modeling}

To train SyncFix, we construct paired distorted and clean multi-view data reflecting realistic reconstruction failure modes. 

We use scenes from the DL3DV \cite{ling2024dl3dv} dataset and train 3D Gaussian Splatting (3DGS) \cite{kerbl20233d} models under multiple {sparsity levels} (6, 8, 10, 12, 16, or 32 training views). This induces geometric inconsistencies, floaters, and blurred high-frequency regions at various levels. We further introduce a {cross-split} strategy: the available training views are partitioned into multiple disjoint subsets, and a separate sparse 3DGS is trained for each split, yielding diverse distortions at different viewpoints from the same underlying scene.
Novel views rendered from these degraded 3DGS models form distorted inputs $X_D$, while ground-truth images provide $X_{GT}$.

To encourage multi-view reasoning, rendered views within each batch are sampled with small pose differences to ensure strong visual overlap. At the same time, these poses are far from the sparse 3DGS training views, guaranteeing large artifacts. This construction forces the model to fix shared structural inconsistencies across views rather than performing independent cosmetic corrections.

\begin{figure*}[t!]
\includegraphics[width=\textwidth]{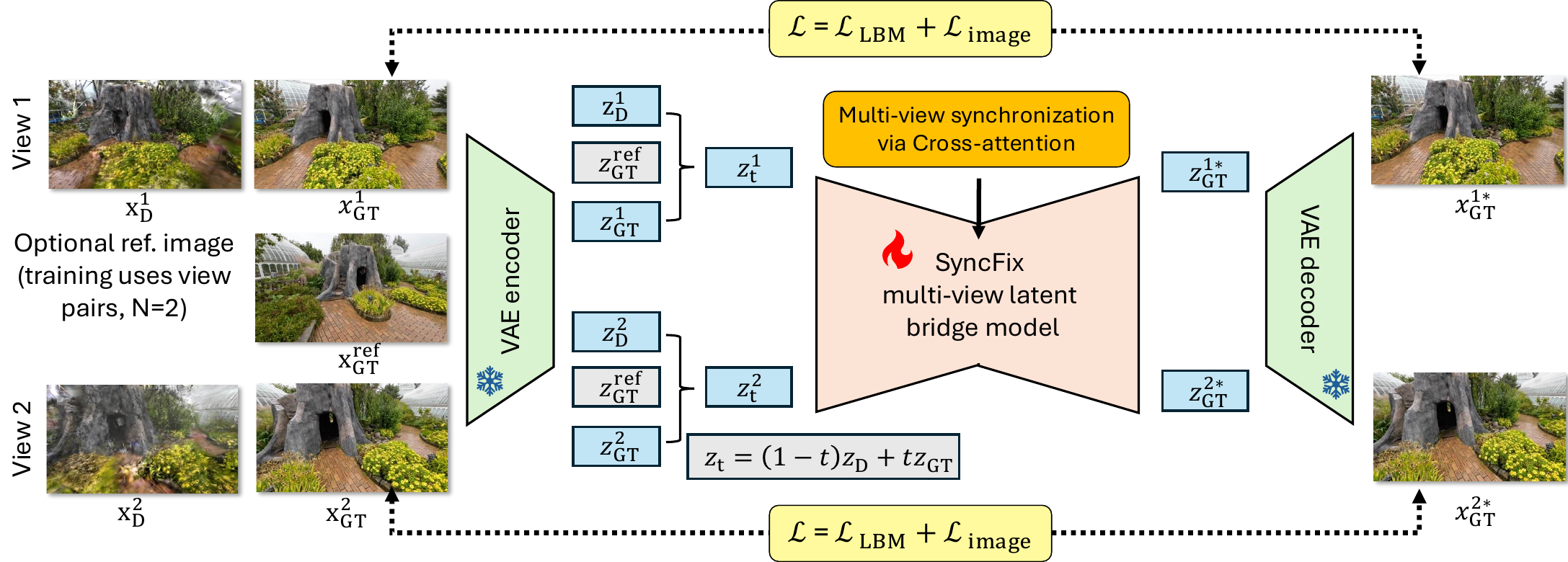}
\caption{\textbf{SyncFix overview.} Distorted renderings from multiple viewpoints $x_D$ are encoded into latent representations and transported toward clean targets $x_{GT}$ using latent bridge matching. 
SyncFix learns a \emph{joint latent bridge} over multiple views, coupling latent trajectories through cross-view attention to enforce multi-view consistency during refinement. 
The model is trained using view pairs ($N=2$) but generalizes to an arbitrary number of views at inference. 
Optional reference images can be provided to guide refinement. 
Here $z_D$ denotes distorted latents, $z_{GT}$ clean target latents, and $z_t = (1-t)z_D + t z_{GT}$ the bridge interpolation between them.}
\label{fig:main_method}
\vspace{-8pt}
\end{figure*}

\subsection{Multi-View Latent Bridge Matching}

Prior diffusion-based refinement methods \cite{wu2025difix3d+} inject artificial noise and perform denoising at a fixed timestep. This assumes artifacts resemble Gaussian corruption. In practice, 3D reconstruction errors are spatially structured and view-dependent; uniform denoising is insufficient.

We adopt a Latent Bridge Matching (LBM) objective to repurpose diffusion model to directly learn the transport from the distorted distribution to clean distribution, yielding a more effective refinement step.

Each image is encoded using a pre-trained autoencoder, yielding latent tensors $z_D$ and $z_{GT}$. For multi-view inputs, we denote the joint latent tensor as:
\begin{equation}
Z_D = \{z_D^{(1)}, \dots, z_D^{(N)}\}, \quad Z_{GT} = \{z_{GT}^{(1)}, \dots, z_{GT}^{(N)}\}.
\end{equation}

Bridge matching defines a continuous path between source and target:
\begin{equation}
Z_t = (1 - t) Z_D + t Z_{GT} + \sigma\sqrt{t(1-t)}\epsilon, \quad t \in [0,1], \quad \epsilon \sim \mathcal{N}(0, I)
\end{equation}
$\sigma$ controls the stochasticity along the path. The corresponding velocity field from 
\begin{equation}
v^*(Z_t, t) = Z_{GT} - Z_D.
\end{equation}

We train a network $v_\theta$ to predict this velocity:
\begin{equation}
\mathcal{L}_{\text{flow}} = \mathbb{E}_{t, Z_D, Z_{GT}} \left[ \| v_\theta(Z_t, t, c) - (Z_{GT} - Z_D) \|_2^2 \right],
\end{equation}
here $c$ is conditioning signals, including distorted renderings and view indices.

Unlike marginal approaches, $v_\theta$ operates on the full multi-view latent tensor by continuously sampling different paired viewpoints, $z_D^{(1)}$ and $z_D^{(2)}$, at each iteration. The predicted vector field is therefore conditioned jointly on all views, defining a probability flow over the product latent space. This naturally induces coupled dynamics: corrections for each view depend on the entire set of views.

\paragraph{Cross-View Attention.}
To implement joint conditioning, we modify the transformer self-attention layers to allow tokens from all views within a batch to attend to one another, learning jointly. Concretely, latent tokens from different views are concatenated along the token dimension, and attention is computed globally while preserving within-view positional encodings. This design is permutation-invariant across views and does not assume fixed view ordering. As a result, the predicted velocity for each view depends explicitly on features from other views, enforcing synchronization during refinement. We also add a reference view, which is the closest training view if available, as in \cite{wu2025difix3d+}. The latent code of reference view $z^{ref}_{GT}$ is concatenated with each of $\{z_D^{(1)}$, $z_D^{(2)}\}$ and $\{z_{GT}^{(1)}$, $z_{GT}^{(2)}\}$ to make $z_D$ and $z_{GT}$ respectively before building the stochastic interpolant $z_t$. In multi-view bridge model, the reference view's tokens are self-attended along with other views' tokens to bridge between contextual information and local details. 

\paragraph{Inference.}
At inference, we initialize $Z_D$, solve the learned differential equation of $Z_t$, and obtain the predicted latent ${Z^*_{GT}}$ in a single step through:
\begin{equation}
\frac{dZ_t}{dt} = v_\theta(Z_t, t, c), \quad Z^*_{GT} = Z_t + v_\theta(Z_t, t, c)
\end{equation}
Decoding $Z^*_{GT}$ yields refined images $\hat{X}$.

Although trained with $N=2$ views for efficiency, the model generalizes to arbitrary $N$ at test time due to the permutation-invariant attention mechanism.

\subsection{Supervision}

To isolate the impact of joint latent modeling, we adopt a similar image-space supervision used in state-of-the-art marginal refinement methods (e.g., Difix3D+). 

Predicted clean images $\hat{X}$ are supervised against $X_{GT}$ using a weighted combination of losses:
\begin{align}
\mathcal{L}_{\text{pixel}} &= \| \hat{X} - X_{GT} \|_1, \\
\mathcal{L}_{\text{LPIPS}} &= \| \phi(\hat{X}) - \phi(X_{GT}) \|_2^2, \\
\mathcal{L}_{\text{Gram}} &= \| G(\phi(\hat{X})) - G(\phi(X_{GT})) \|_F^2.
\end{align}

The total loss to train our model is represented as:
\begin{equation}
    \mathcal{L} = \mathcal{L}_{\text{flow}} + \mathcal{L}_{\text{pixel}} + \lambda_{\text{lpips}}\mathcal{L}_{\text{LPIPS}} + \lambda_{\text{Gram}}\mathcal{L}_{\text{Gram}}
\end{equation}
Unlike marginal baselines, these losses are applied to jointly coupled outputs, ensuring that structural and textural corrections remain consistent across views.

\section{Experiments}
\subsection{Datasets}

We build our training set from 360 scenes of the DL3DV dataset \cite{ling2024dl3dv}, covering diverse indoor and outdoor environments. For each scene, we render paired \emph{corrupted} and \emph{clean} images using 3D Gaussian Splatting (3DGS) \cite{kerbl20233d}, resulting in 480{,}000 paired samples in total. An additional \emph{disjoint} set of 40 scenes is reserved for testing, comprising 41{,}442 paired samples.

We apply a view-splitting protocol designed to prevent trivial overlap between training and rendering views. Specifically, training views are sampled across different splits to encourage broad coverage, while render views are sampled to be mutually closer within each split yet spatially separated from the training views, yielding a challenging novel-view setting. During training, we dynamically assemble multi-view inputs/targets by sampling from a size-bounded cache for efficient view combinations.

To evaluate zero-shot generalization, we report results on Nerfbusters~\cite{warburg2023nerfbusters}, which contains 12 scenes with off-trajectory render views. We adopt their official train/test view split protocol for a fair comparison.

\subsection{Evaluation and Implementation}
We compare SyncFix with Difix3D+ \cite{wu2025difix3d+}, which introduces the iterative refinement by adding refined views to 3D reconstruction and post-cleanup with a diffusion model. We also compare with Fixer in the appendix, which is a follow-up work of Difix3D+ using Cosmos-Predict2 \cite{ali2025world} as its backbone, with the same training strategy but without a reference view. We evaluate image fidelity using pixel-based metrics (PSNR and SSIM) and perceptual metrics (LPIPS, DreamSim~\cite{fu2023dreamsim}, and FID). DreamSim is an ensemble of perceptual embedding models fine-tuned to better align with human judgments of visual similarity, and recent studies~\cite{wickrema2025benchmarking} suggest that it provides a robust signal for assessing image quality in 3D reconstruction and novel view synthesis.

Our method is built on the SDXL backbone \cite{podell2023sdxl}. We fine-tune the modified multi-view diffusion U-Net while keeping the VAE encoder and decoder frozen. We train our model for 100{,}000 optimization steps with a learning rate of $4\times10^{-5}$ and a batch size of 2, using four NVIDIA H100 GPUs. We set $\lambda_{\text{lpips}}=10$ and $\lambda_{\text{Gram}}=0.1$.

\subsection{Cross-View Semantic Consistency Metric}

Maintaining semantic consistency across views is a central objective in 3D reconstruction and multi-view synthesis. Standard reference-based metrics such as PSNR, LPIPS, and DreamSim evaluate image quality at the single-view level and do not capture agreement between views. To quantify multi-view consistency, we introduce a \textit{Cross-View Semantic Consistency} (CVSC) metric that measures semantic feature agreement between matched regions across views. Unlike MEt3R~\cite{asim2025met3r}, which relies on dense correspondences that can be unreliable under occlusions, scene dynamics, or challenging textures, CVSC emphasizes sparse yet high-confidence correspondences obtained via a keypoint matcher and geometric verification.

Formally, given a series of refined views 
$X^*_{GT} = \{x_{GT}^{(1)*}, \dots, x_{GT}^{(N)*}\}$,
we form pairs of adjacent views 
$(x_{GT}^{(i)*}, x_{GT}^{(i+1)*})$, yielding $N-1$ view pairs. 
For each pair, we extract keypoints using RaCo~\cite{shenoi2026raco} and establish correspondences with LightGlue~\cite{lindenberger2023lightglue}. 
This produces a set of matched keypoints
$K_{\text{pair}} = \{(k_A^{(i)}, k_B^{(i)})\}_{i=1}^{M}$.

To suppress mismatches, we perform geometric verification by estimating the fundamental matrix with RANSAC and retain only inlier correspondences 
$\hat{K}_{\text{pair}} = \{(\hat{k}_A^{(i)}, \hat{k}_B^{(i)})\}$.

We then extract dense semantic feature maps $F_A$ and $F_B$ from DINOv3~\cite{simeoni2025dinov3}. 
For each verified correspondence, we compute cosine similarity between features at the matched locations. 
The CVSC score for a pair of views is defined as
\vspace{-5pt}
\begin{equation}
\mathrm{CVSC}
=
\frac{1}{|\hat{K}|}
\sum_{i=1}^{|\hat{K}|}
\frac{\langle F_A(\hat{k}_A^{(i)}), F_B(\hat{k}_B^{(i)}) \rangle}
{\|F_A(\hat{k}_A^{(i)})\|_2 \, \|F_B(\hat{k}_B^{(i)})\|_2}.
\end{equation}

The final CVSC score is obtained by averaging across all view pairs in the sequence. Higher CVSC indicates stronger semantic agreement across viewpoints and thus better multi-view consistency.

\begin{table}[t]
\centering
\setlength{\tabcolsep}{3pt}
\caption{
Quantitative comparison on the DL3DV and NeRFBusters test sets for refining sparse-view 3D Gaussian Splatting reconstruction renderings.
SyncFix improves reconstruction quality and cross-view semantic consistency over prior generative refinement methods. $^{\dag}$ indicates no reference views during training. $\uparrow$ indicates higher-is-better and $\downarrow$ indicates lower-is-better.
\textbf{Bold} denotes best results and \underline{underline} denotes second best.
}
\resizebox{\textwidth}{!}{
\begin{tabular}{lccccc|ccccc}
\toprule
& \multicolumn{5}{c}{DL3DV} & \multicolumn{5}{c}{NeRFBusters} \\
\cmidrule(lr){2-6} \cmidrule(lr){7-11}
Method 
& PSNR$\uparrow$ & LPIPS$\downarrow$ & DSIM$\downarrow$ & FID$\downarrow$ & CVSC$\uparrow$
& PSNR$\uparrow$ & LPIPS$\downarrow$ & DSIM$\downarrow$ & FID$\downarrow$ & CVSC$\uparrow$ \\
\midrule
3DGS \cite{kerbl20233d}
& 15.94 & 0.454 & 0.297 & 80.8 & \underline{0.875}
& 11.85 & 0.575 & 0.336 & 139.4 & \textbf{0.951} \\
\midrule
\rowcolor{lightgrayrow}
Fixer \cite{nvtlabs_fixer_github}
& 15.73 & 0.466 & 0.257 & 45.45 & 0.822
& 12.32 & 0.588 & 0.296 & 126.7 & 0.817 \\
\rowcolor{lightgrayrow}
Difix3D+$^{\dag}$ \cite{wu2025difix3d+}
& 16.05 & 0.368 & 0.171 & 26.2 & 0.819
& 11.94 & 0.528 & 0.202 & 97.4 & 0.925 \\
\rowcolor{darkgrayrow}
Difix3D+ \cite{wu2025difix3d+}
& 16.17 & 0.343 & 0.135 & \textbf{16.7} & 0.850
& 12.05 & \underline{0.496} & \underline{0.174} & \underline{81.1} & 0.931 \\
\midrule
\rowcolor{lightgrayrow}
SyncFix$^{\dag}$
& \underline{16.45} & \underline{0.334} & \underline{0.129} & 22.5 & \underline{0.862}
& \underline{14.07} & 0.517 & 0.183 & 86.2 & \underline{0.937} \\
\rowcolor{darkgrayrow}
SyncFix
& \textbf{16.94} & \textbf{0.305} & \textbf{0.099} & \underline{17.5} & \textbf{0.880}
& \textbf{14.34} & \textbf{0.494} & \textbf{0.173} & \textbf{80.4} & \underline{0.940} \\
\bottomrule
\end{tabular}
}
\label{tab:quantitative}
\vspace{-10pt}
\end{table}

\begin{figure}[t!]
\small
\begin{center}
    \setlength{\tabcolsep}{1pt}      %
    \renewcommand{\arraystretch}{1.0} %
    \small
    
    \settoheight{\imgH}{\includegraphics[width=0.23\linewidth]{figures/teaser_abhay_arxiv_v2/difix_images_pretrained_ref/frame_00337_compare_teaser/frame_00337_compare_teaser_annotated.jpg}}
    
    \newcommand{\viewlabel}[1]{%
      \parbox[b][\imgH][c]{1em}{\centering\rotatebox{90}{#1}}%
    }
    
    \begin{tabular}{c c c c c}
    \viewlabel{View 1} &
    \includegraphics[width=0.235\linewidth]{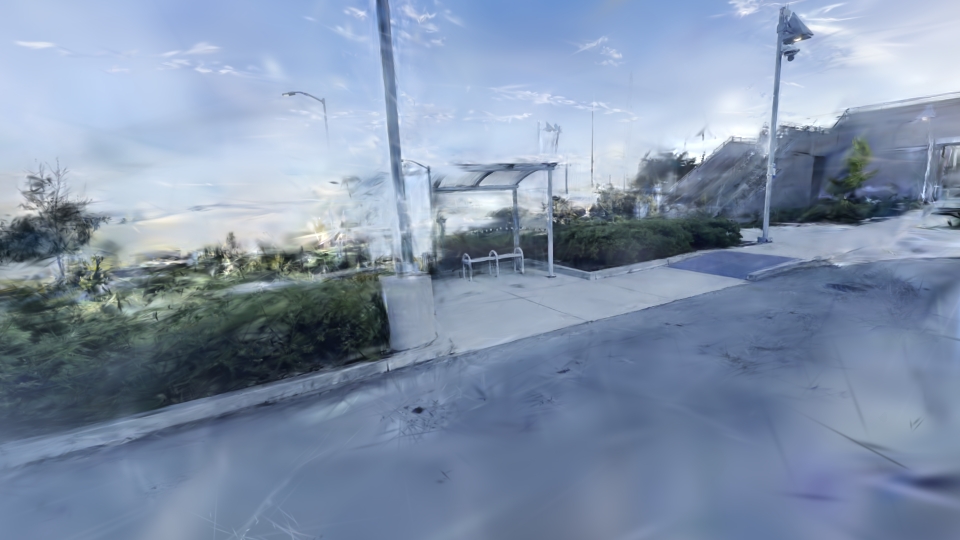} &
    \includegraphics[width=0.235\linewidth]{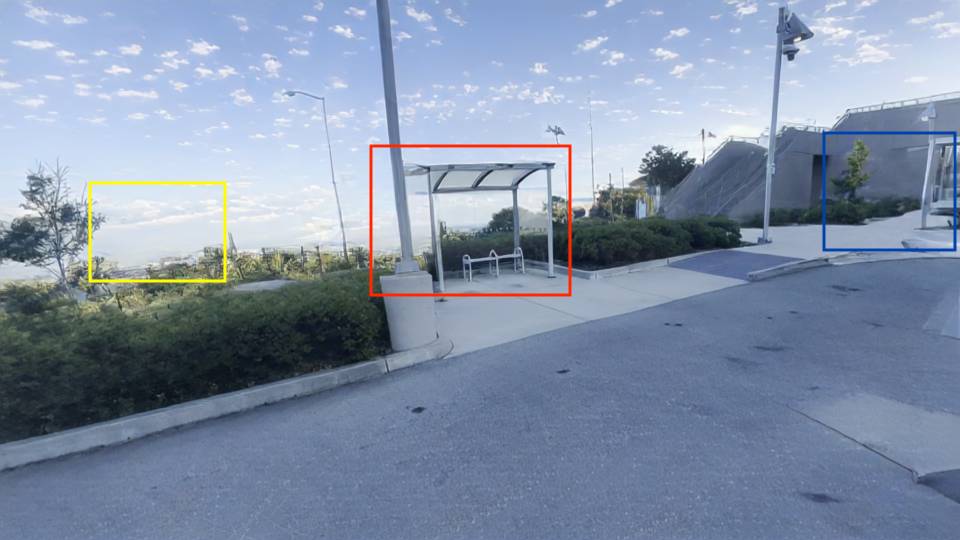} &
    \includegraphics[width=0.235\linewidth]{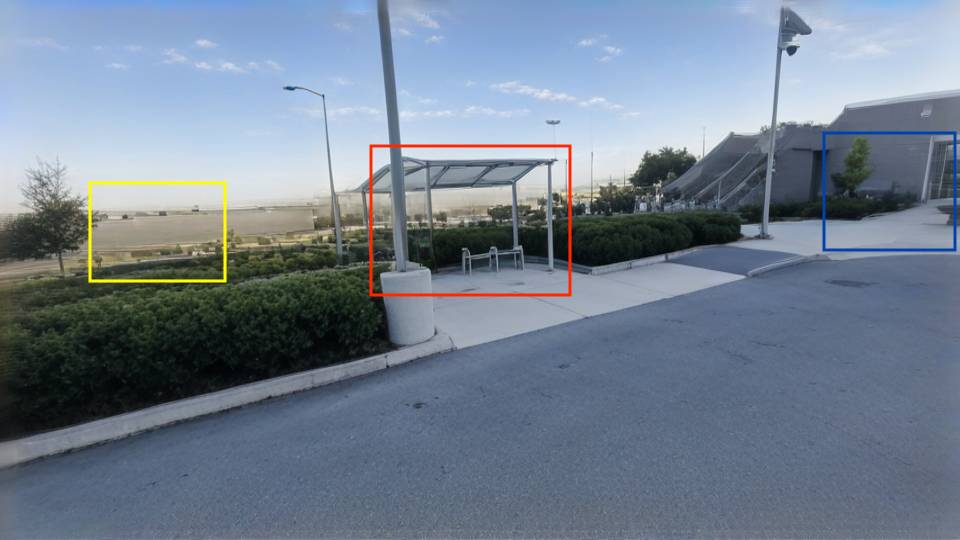} &
    \includegraphics[width=0.235\linewidth]{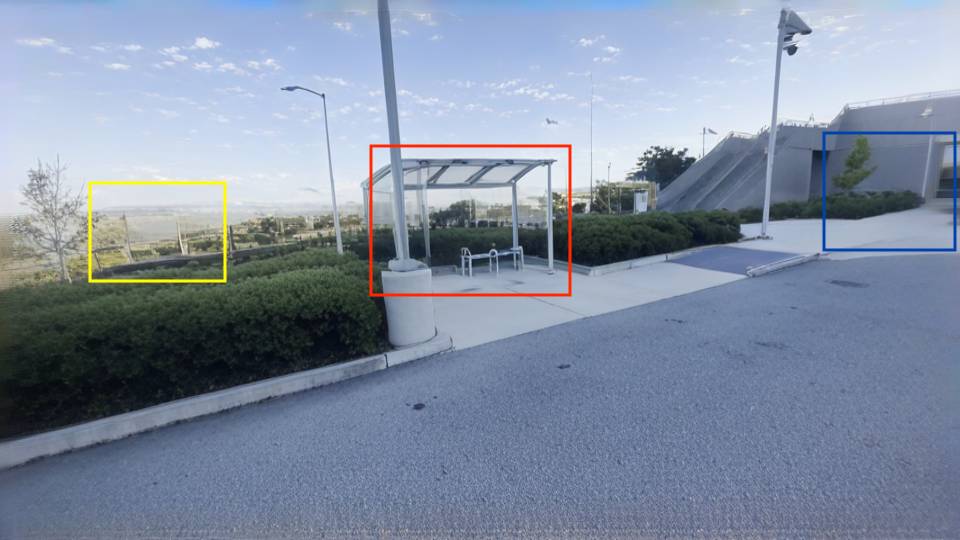} \\
    
    \viewlabel{View 2} &
    \includegraphics[width=0.235\linewidth]{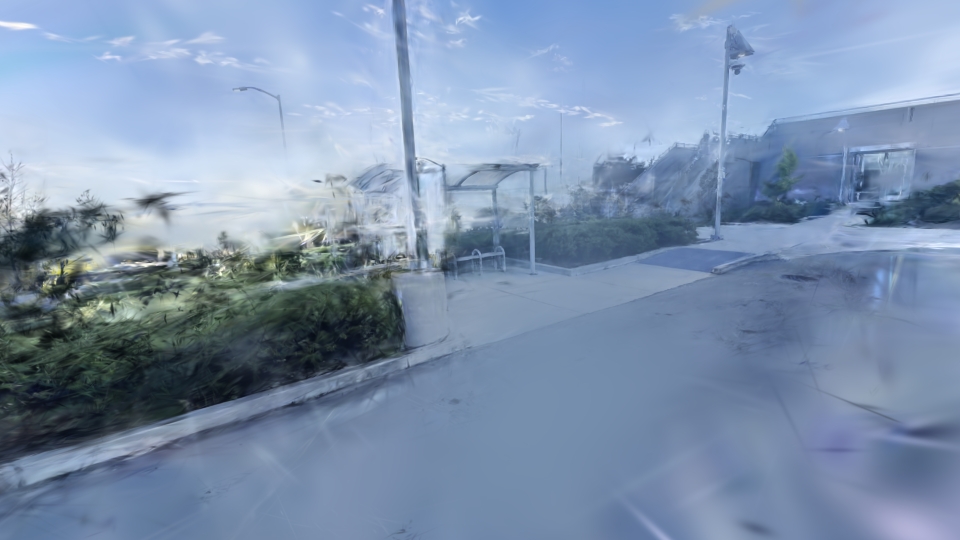} &
    \includegraphics[width=0.235\linewidth]{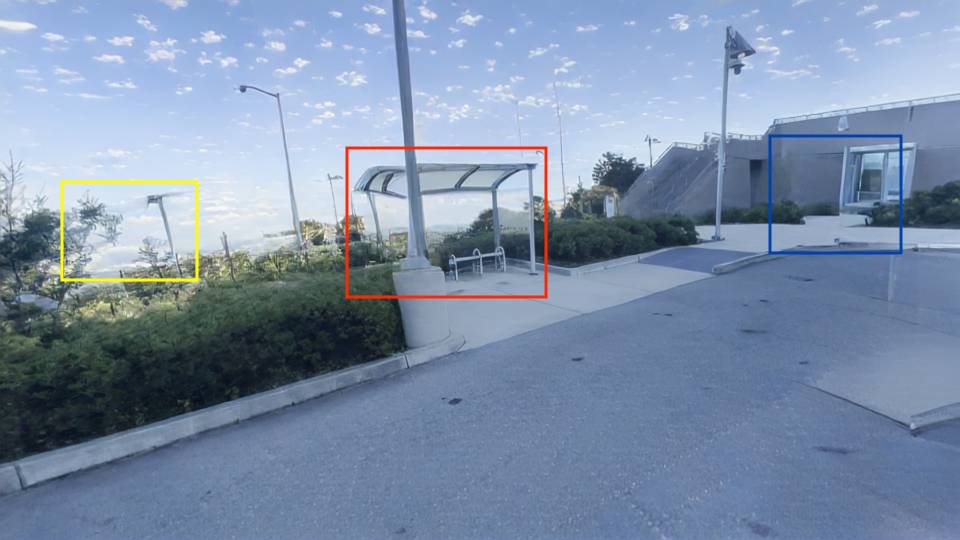} &
    \includegraphics[width=0.235\linewidth]{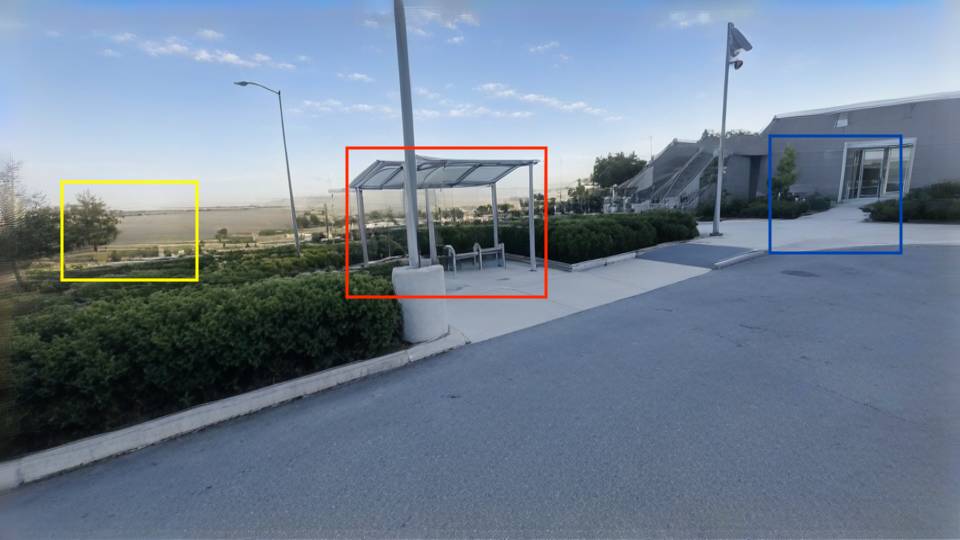} &
    \includegraphics[width=0.235\linewidth]{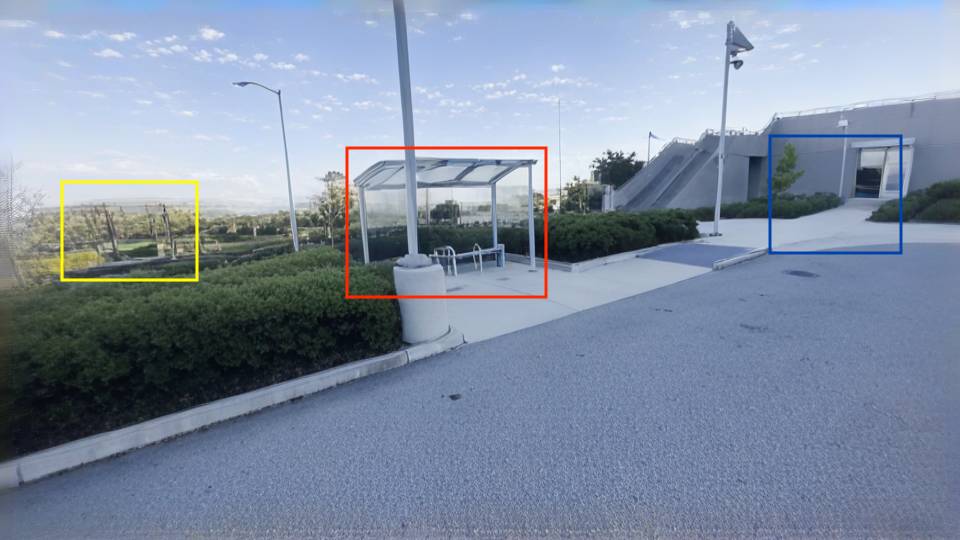} \\

    \viewlabel{View 1} &
    \includegraphics[width=0.235\linewidth]{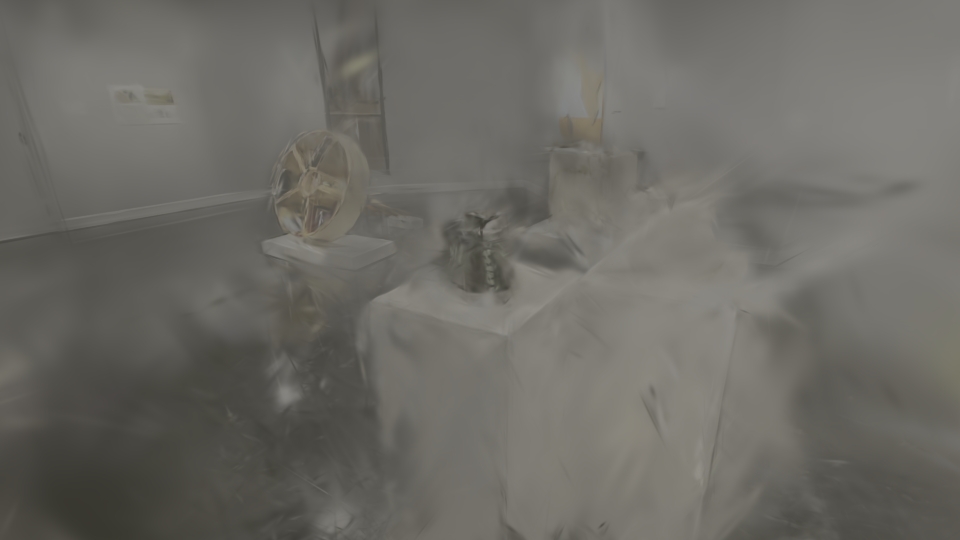} &
    \includegraphics[width=0.235\linewidth]{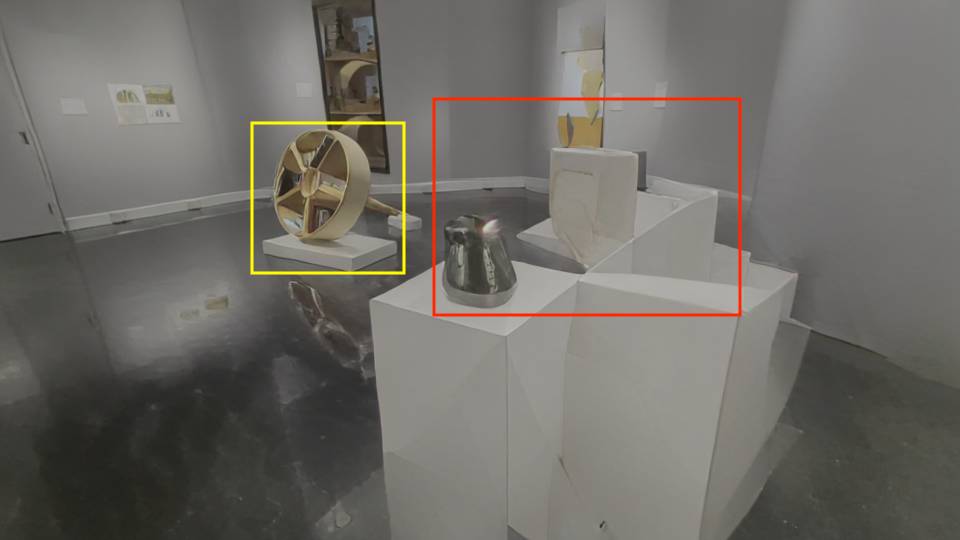} &
    \includegraphics[width=0.235\linewidth]{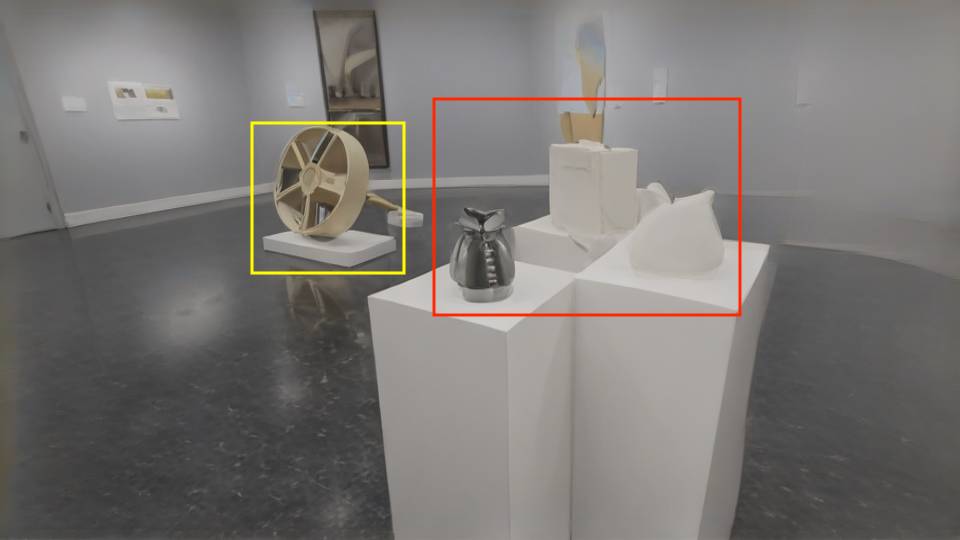} &
    \includegraphics[width=0.235\linewidth]{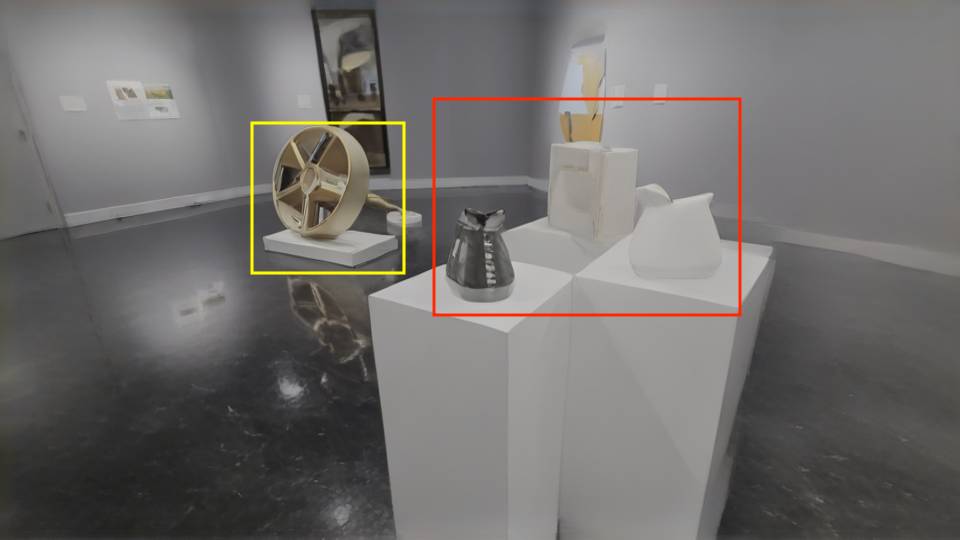} \\
    
    \viewlabel{View 2} &
    \includegraphics[width=0.235\linewidth]{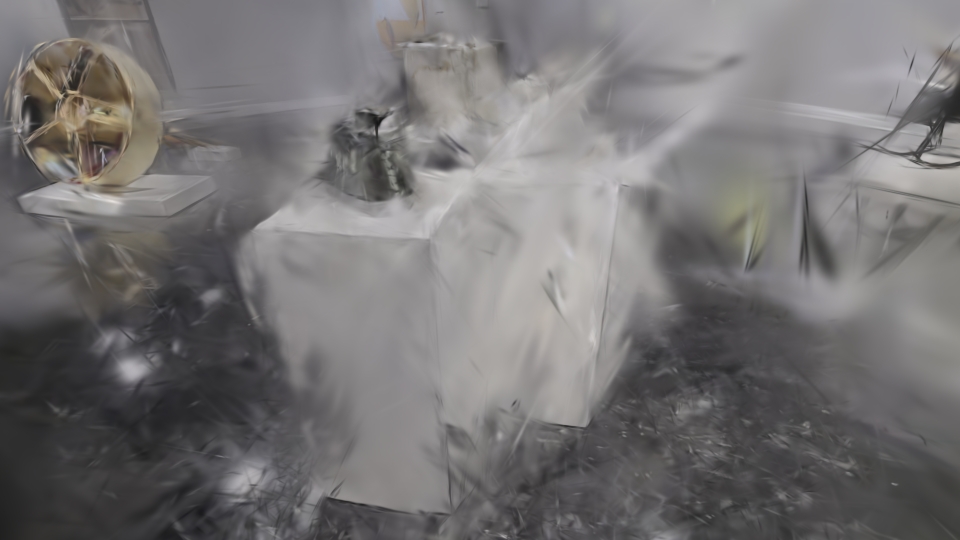} &
    \includegraphics[width=0.235\linewidth]{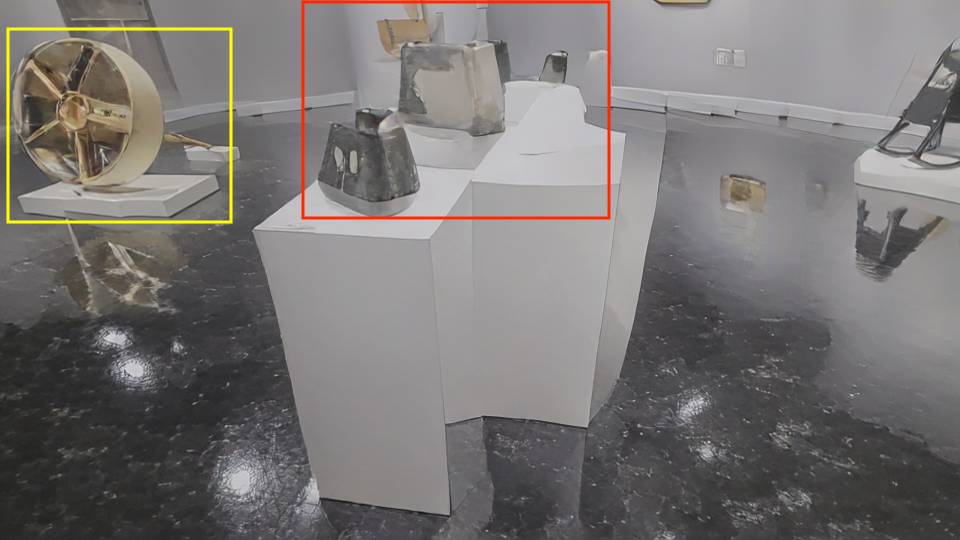} &
    \includegraphics[width=0.235\linewidth]{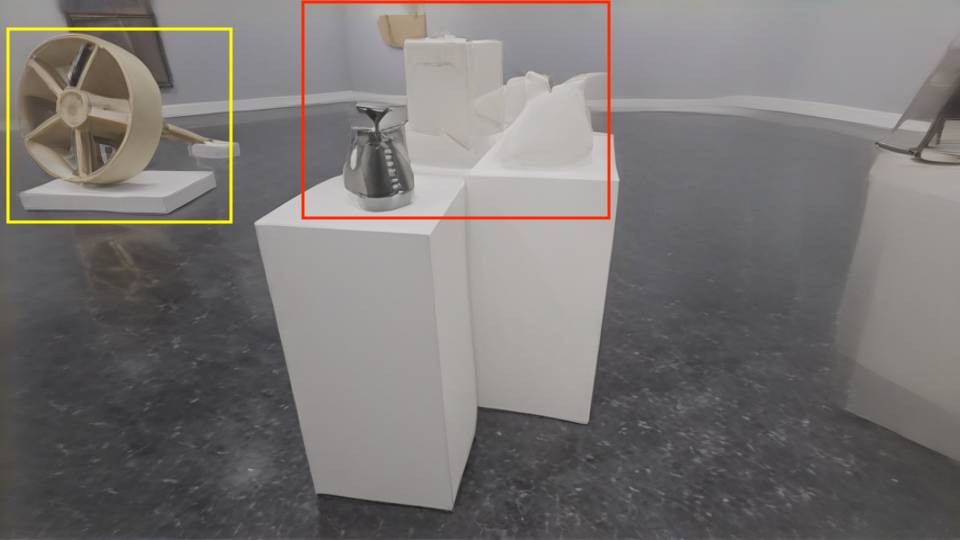} &
    \includegraphics[width=0.235\linewidth]{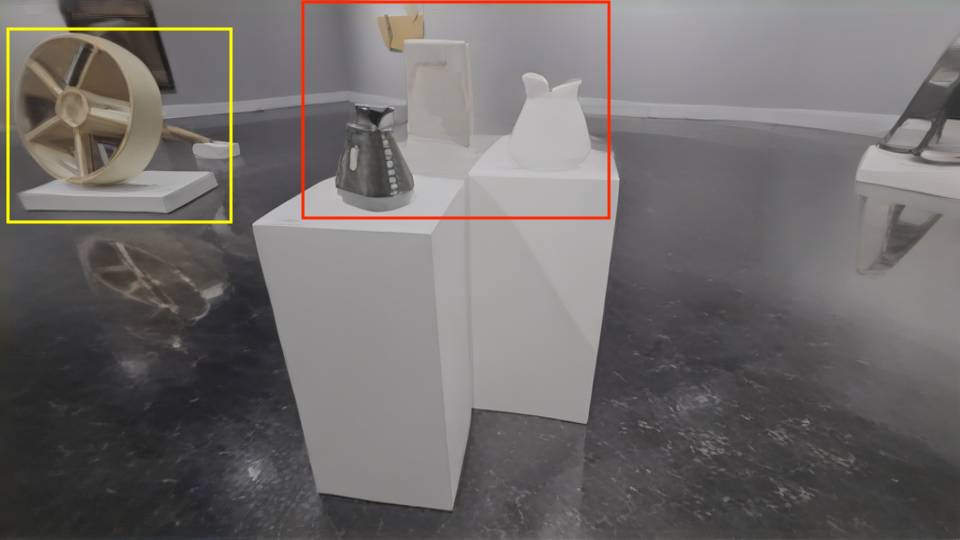} \\

    \viewlabel{View 1} &
    \includegraphics[width=0.235\linewidth]{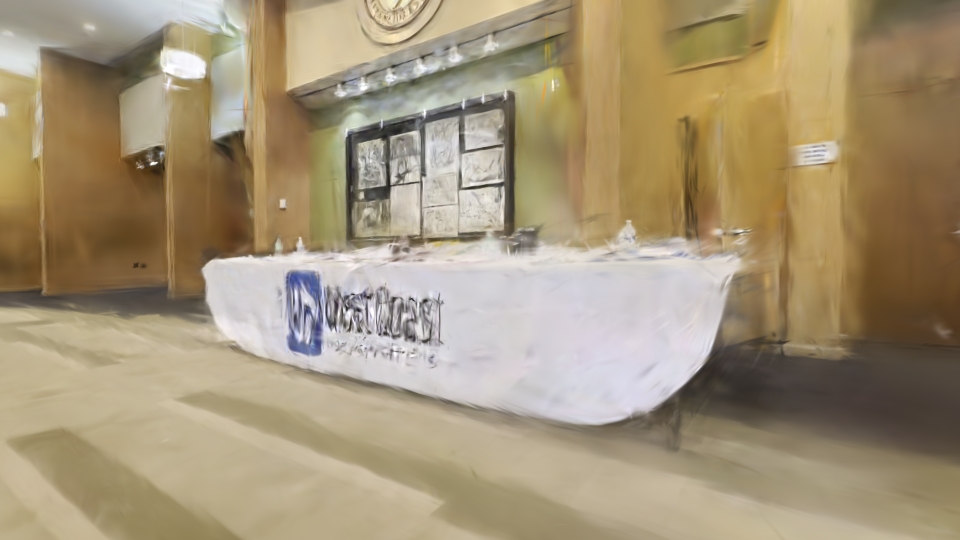} &
    \includegraphics[width=0.235\linewidth]{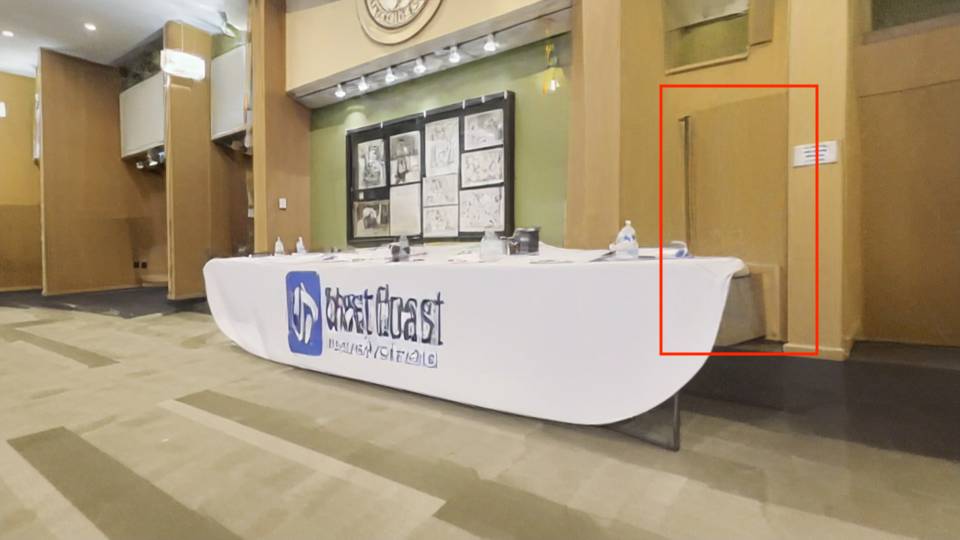} &
    \includegraphics[width=0.235\linewidth]{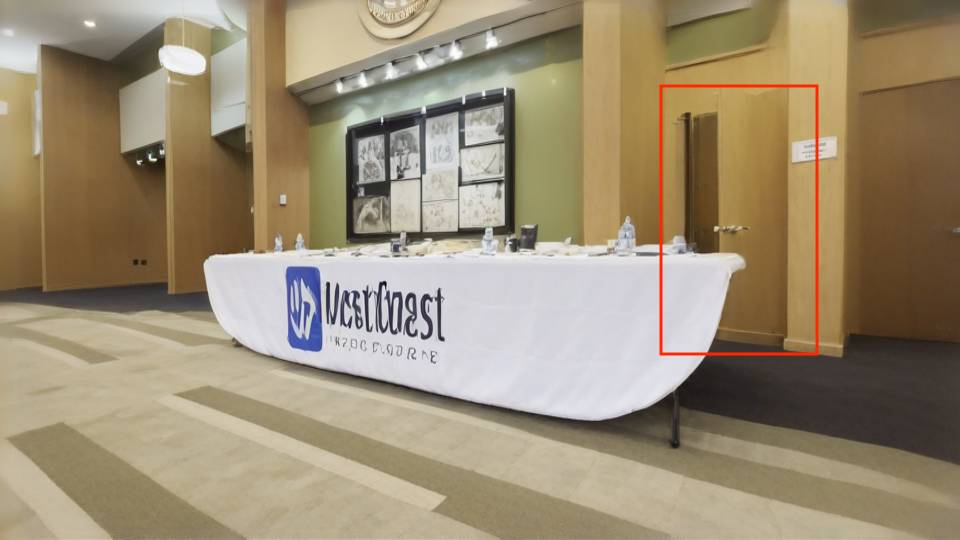} &
    \includegraphics[width=0.235\linewidth]{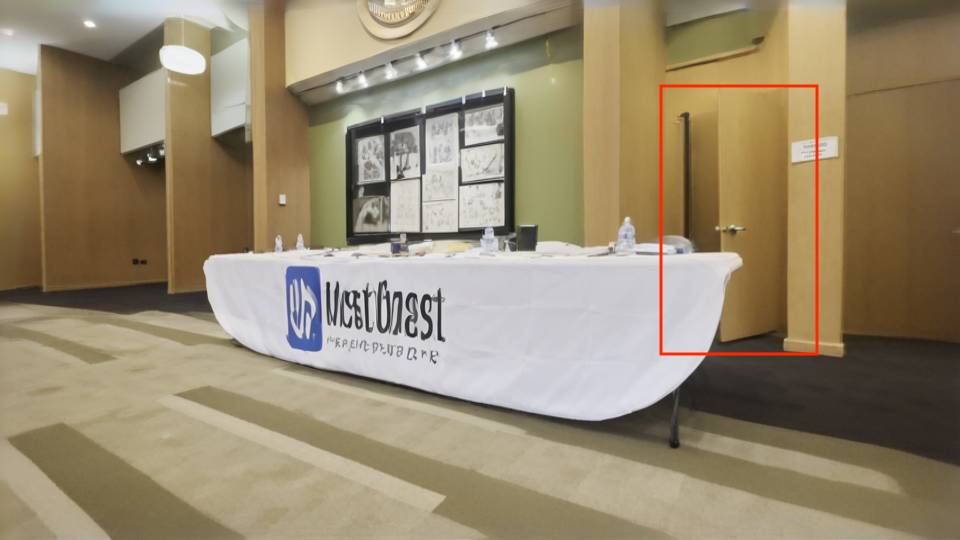} \\
    
    \viewlabel{View 2} &
    \includegraphics[width=0.235\linewidth]{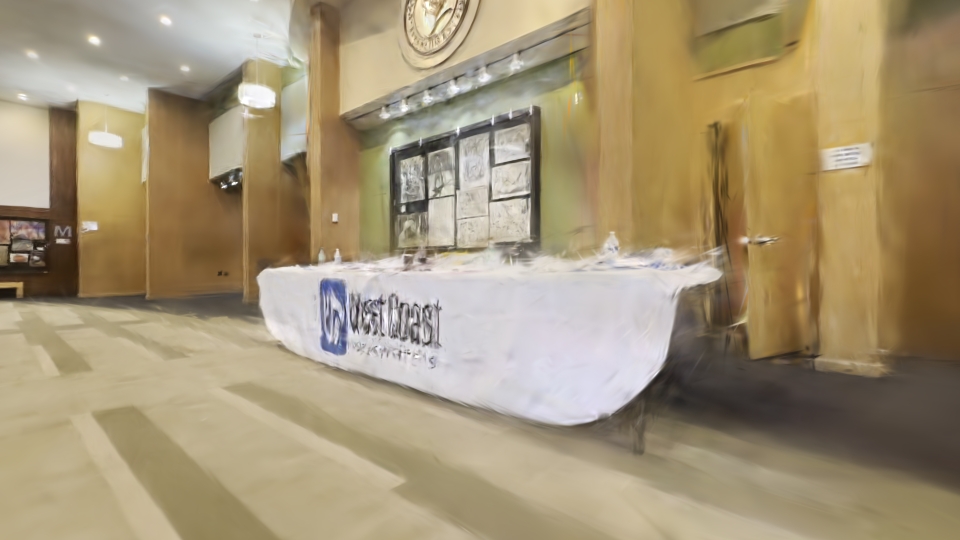} &
    \includegraphics[width=0.235\linewidth]{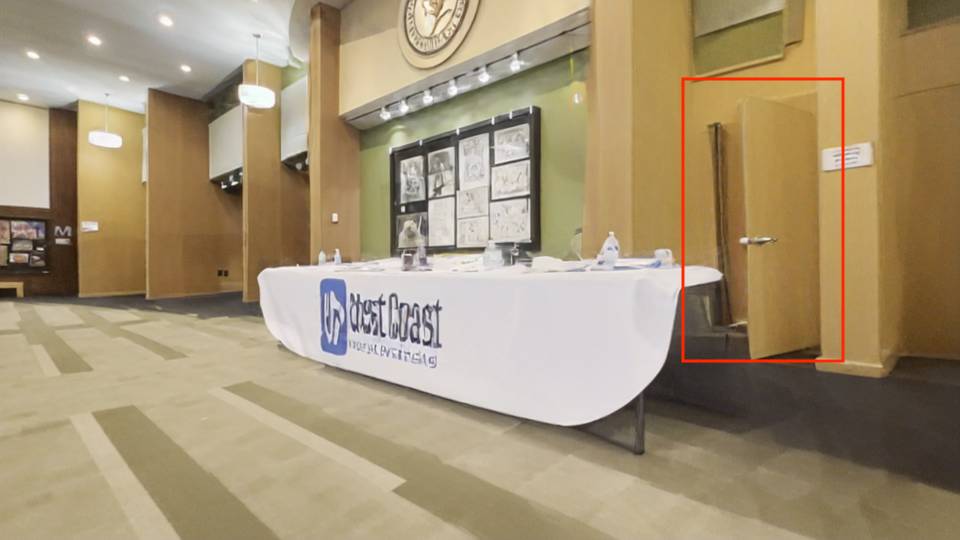} &
    \includegraphics[width=0.235\linewidth]{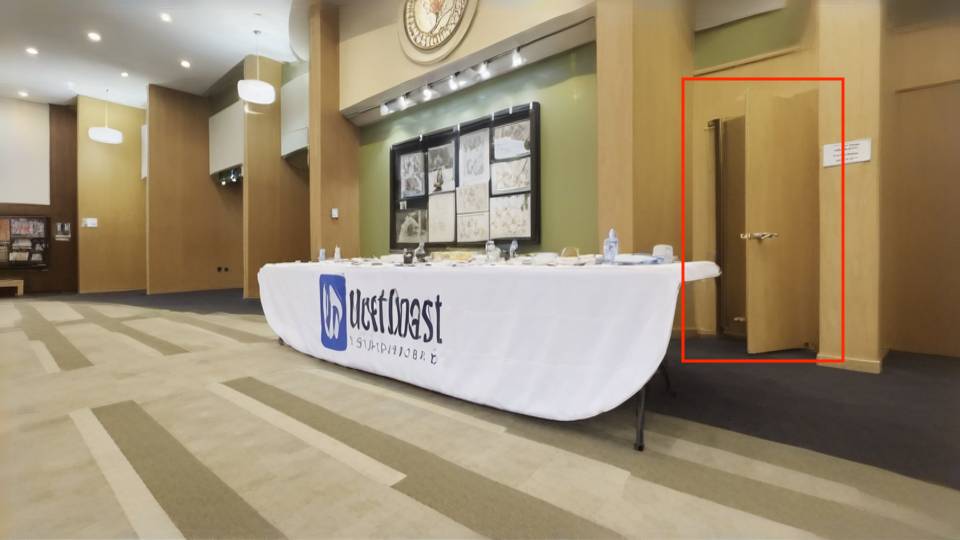} &
    \includegraphics[width=0.235\linewidth]{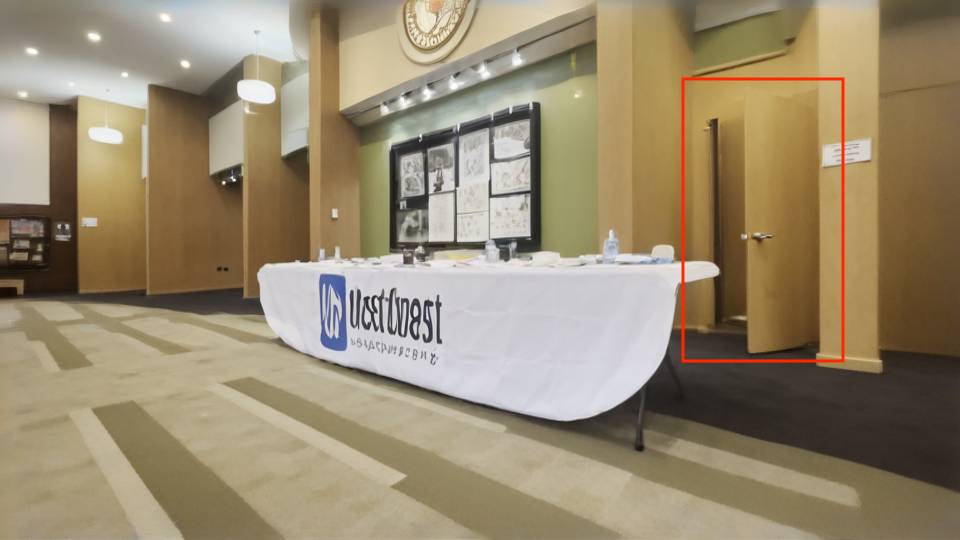} \\
            & 3DGS 
            & Difix3D+ & Ours$^{\dag}$ & Ours 
    \end{tabular}
\captionof{figure}{\textbf{Multi-view consistency under extreme sparse-view degradation}. The baseline 3DGS model (left) displays severe structural artifacts and floaters. While the marginal refinement method Difix3D+ (middle) removes these base artifacts, its independently generated views hallucinate inconsistent geometry across camera poses. As highlighted by the red boxes, Difix3D+ distorts the architectural structure of the outdoor bus shelter in the top scene. In the middle scene, it causes the object on the indoor pedestal to undergo shape changes between View 1 and View 2. Similarly, in the bottom scene, a door visible in one view is absorbed into the wall in another. In contrast, SyncFix (Ours) leverages joint latent synchronization to ensure that hallucinated high-frequency details and structural repairs remain geometrically and semantically consistent across views. Ours$^{\dag}$ is a model trained without clean reference images.}
\label{fig:dl3dv_exp}
\end{center}
\vspace{-10pt}
\end{figure}

\subsection{DL3DV Dataset}
Table \ref{tab:quantitative} (first block) shows the quantitative metrics on DL3DV test set. Although Fixer improves the renderings with a lower DreamSim score and FID compared to 3DGS, it obtains suboptimal results compared with Difix3D+ and SyncFix. We attribute this gap primarily to its single-view nature and the absence of reference images, which limit Fixer’s ability to disambiguate missing content and artifacts for high-fidelity refinements.

SyncFix exceeds Difix3D+ by 0.77 dB on PSNR and reduces LPIPS by 0.038. Notably, SyncFix decreases DreamSim score by 27 percent compared to Difix3D+ and FID by more than four times compared to the 3DGS renderings. Similarly, Syncfix outperforms Difix3D+ in their versions without using reference images. As seen in Figure \ref{fig:dl3dv_exp}, Difix3D+ mitigates the noises and sharpens the renderings. However, Difix3D+ struggles in a multi-view context, i.e., changing geometry of the objects and overfitting to view-dependent artifacts. In comparison, our method reduces artifacts globally through a joint refinement of the renderings and maintains a more 3D-consistent geometry of the scene. This result shows that Difix3D+ is akin to single-view refinement with guidance from reference view, and the multi-view inconsistent artifacts are difficult to remove during 3D reconstruction. Our multi-view training strategy enables consistent generation over a joint distribution of the scene. Interestingly, the raw 3DGS baseline achieves a relatively high CVSC score. This occurs because artifacts produced by sparse-view reconstruction, such as floaters and blurred textures, are often geometrically consistent across views. As a result, although the reconstruction quality is poor, the errors appear similarly from different viewpoints and therefore yield high semantic agreement. In contrast, marginal refinement methods such as Difix3D+ improve perceptual quality but can introduce view-specific hallucinations, reducing cross-view consistency and lowering the CVSC score. SyncFix addresses this issue by jointly refining multiple views, preserving geometric agreement while improving perceptual fidelity, which leads to a better CVSC score overall. More visual comparison can be found in Sec. \ref{sec:quali}.

\begin{figure}[t]
    \centering
    \begin{subfigure}[t]{0.49\linewidth}
        \centering
        \includegraphics[width=\linewidth]{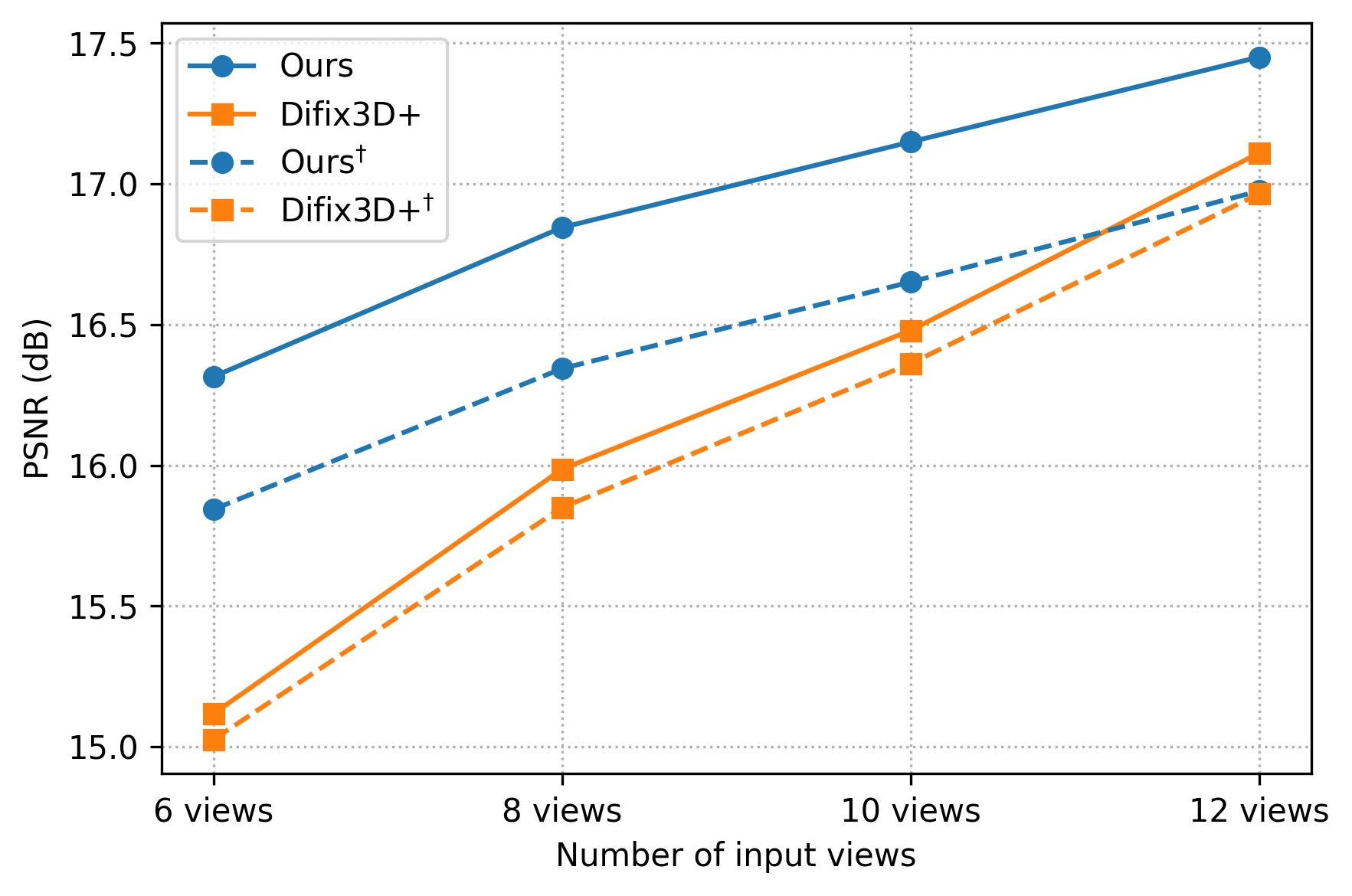}
        \caption{PSNR vs. number of views.}
        \label{fig:views_psnr}
    \end{subfigure}\hfill
    \begin{subfigure}[t]{0.49\linewidth}
        \centering
        \includegraphics[width=\linewidth]{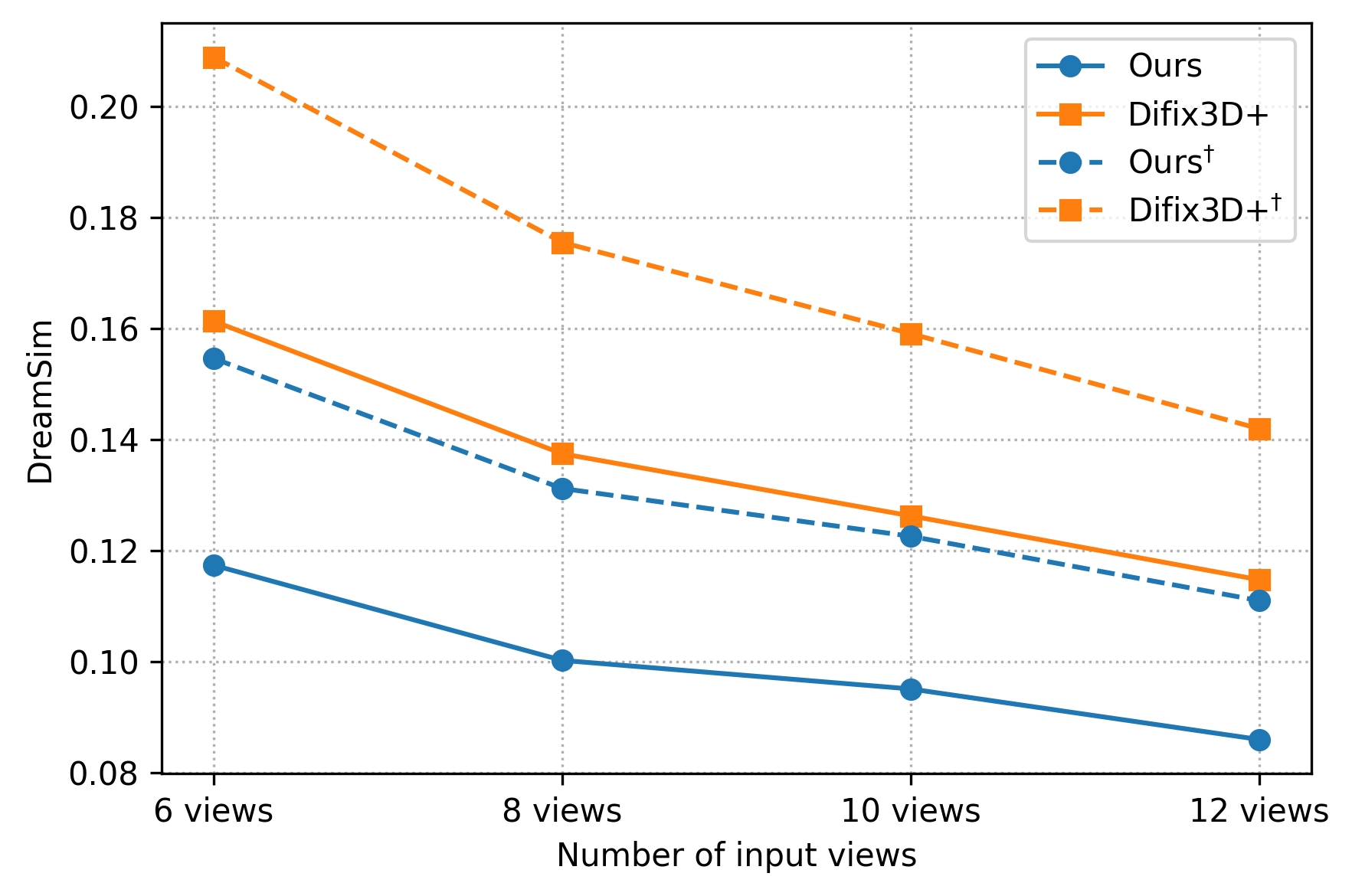}
        \caption{DreamSim vs. number of views.}
        \label{fig:views_dreamsim}
    \end{subfigure}
    \vspace{-5pt}
\caption{
\textbf{Performance at different number of 3DGS training views on the DL3DV dataset.} SyncFix consistently achieves higher PSNR and lower DreamSim than Difix3D+ at different numbers of views, and the benefit of SyncFix is more pronounced at sparser settings.
}
    \label{fig:views_ablation}
        \vspace{-5pt}
\end{figure}

To break down performance under different numbers of 3DGS training views, we report PSNR and DreamSim as a function of the number of views in Fig.~\ref{fig:views_ablation}. SyncFix and its no-reference variant consistently outperform those of Difix3D+ at different view counts. We observe that the performance gap is more prominent in the sparse-view regime: our model provides stronger generative capability to recover missing content under severe corruption, whereas Difix3D+ performs comparatively better when artifacts are milder, and denoising becomes easier.

\begin{figure*}[t!]
\includegraphics[width=0.98\textwidth]{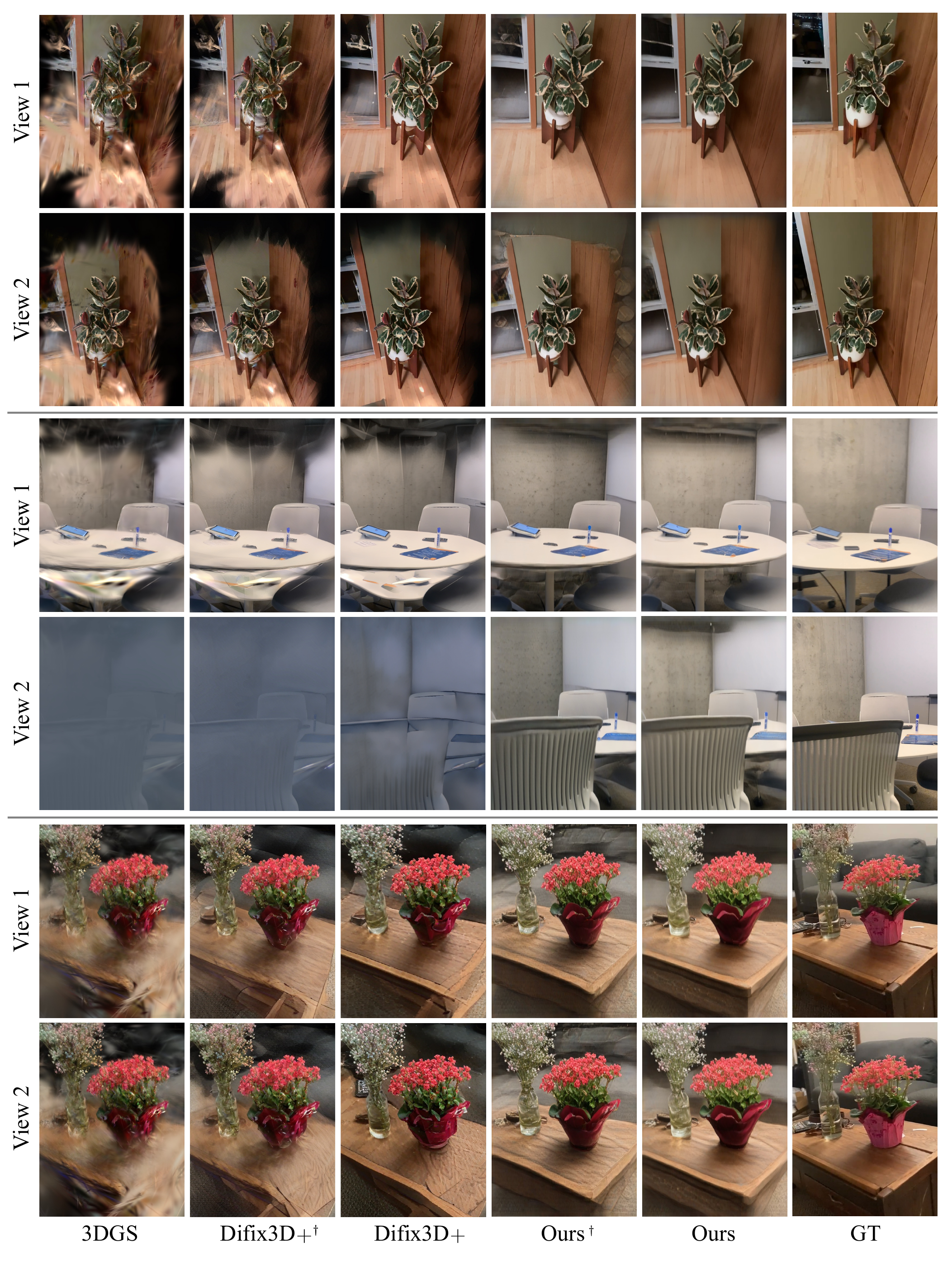}
\vspace{-5pt}
\caption{
\textbf{Comparison on the NeRFBusters evaluation dataset.}
3DGS reconstructions contain severe artifacts and geometric distortions.
Difix3D+ improves the appearance of individual renderings but struggles to extrapolate to unseen regions and often produces unstable structures across views.
In contrast, SyncFix generates plausible reconstructions that remain consistent across viewpoints, recovering coherent object geometry and scene structure. $^{\dag}$ are models trained without clean reference images.
}
\label{fig:nerfbuster}
\end{figure*}

\subsection{Nerfbusters Dataset}

We evaluate the generalization ability of our method on the out-of-distribution NeRFbusters dataset, which provides off-trajectory testing views from dense-view reconstructions and exhibits different camera aspect ratios than our training data. Based on Table \ref{tab:quantitative} (second block on the right), SyncFix achieves notably better PSNR and CVSC than Difix3D+ quantitatively. As shown in Figure \ref{fig:dl3dv_exp}, our method synthesizes a coherent scene from these two views that faithfully fills in the missing door frame and walls, whereas Difix3D+ refined renderings often exhibit inconsistencies across the views. In addition, SyncFix suppresses artifacts globally, yielding clean, refined views. This visualization shows that denoising at a fixed step \cite{wu2025difix3d+} is not sufficient to remove artifacts spanning different severities and provides limited generative capacity. In contrast, SyncFix learns a direct mapping from the corrupted rendering distribution to the clean distribution, producing richer yet faithful generation guided by implicitly learned multi-view consistency.

\subsection{Generalization to Feedforward Method}
\label{sec:feedforward}
3D Gaussian Splatting (3DGS) \cite{kerbl20233d} achieves high-quality novel view synthesis with efficient rendering. However, it requires per-scene optimization as it overfits to each scene. To enable faster inference and leverage prior knowledge learned from large-scale data, recent feedforward splatting methods have gained traction by directly predicting 3D Gaussian parameters and depth from input images. This motivates evaluating whether SyncFix can generalize to refine multi-view renderings produced by feedforward pipelines. To this end, we conduct experiments using AnySplat \cite{jiang2025anysplat} on the Mip-NeRF 360 dataset \cite{barron2021mip}, which includes both indoor and outdoor scenes, to further assess the zero-shot capability of our method.

AnySplat supports feedforward rendering from uncalibrated images by employing separate decoder heads to predict 3D Gaussian parameters, depth, and camera poses. For each scene, we randomly sample 10 input images,  and the target views are selected sequentially while being far from the input views. We follow the evaluation protocol that we use VGGT \cite{wang2025vggt} to calibrate the input and target views in two passes. For pose alignment, we normalize camera poses by setting the first camera to the identity and estimate a relative scale factor to align camera translations across the two sets.

\begin{figure}[t!]
\begin{center}
    \setlength{\tabcolsep}{1pt}      %
    \renewcommand{\arraystretch}{1.0} %
    \small
    
    \settoheight{\imgH}{\includegraphics[width=0.28\linewidth]{figures/teaser_abhay_arxiv_v2/difix_images_pretrained_ref/frame_00337_compare_teaser/frame_00337_compare_teaser_annotated.jpg}}
    
    \newcommand{\viewlabel}[1]{%
      \parbox[b][\imgH][c]{1.0em}{\centering\rotatebox{90}{\text{#1}}}%
    }
    
    \begin{tabular}{c c c c c}

    \viewlabel{View 1} &
    \includegraphics[width=0.235\linewidth]{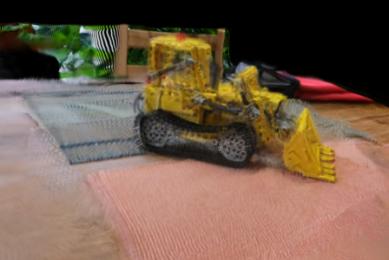} &
    \includegraphics[width=0.235\linewidth]{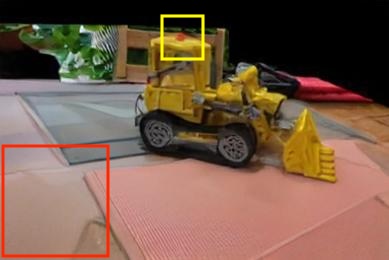} &
    \includegraphics[width=0.235\linewidth]{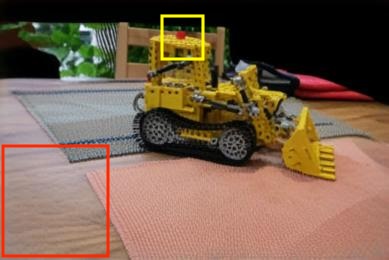} &
    \includegraphics[width=0.235\linewidth]{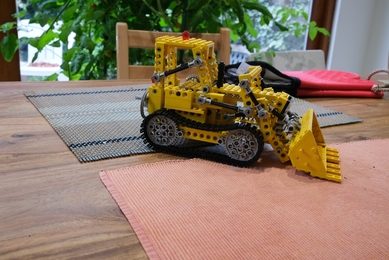} \\
    
    \viewlabel{View 2} &
    \includegraphics[width=0.235\linewidth]{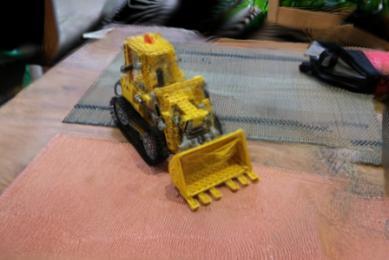} &
    \includegraphics[width=0.235\linewidth]{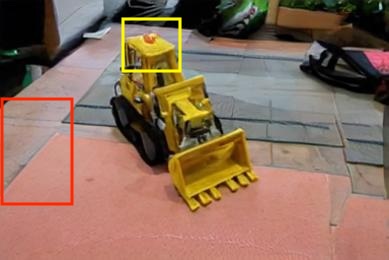} &
    \includegraphics[width=0.235\linewidth]{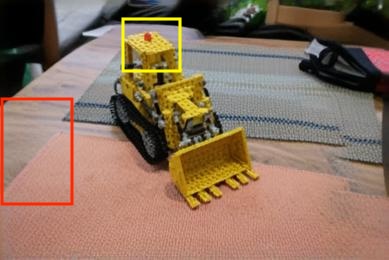} &
    \includegraphics[width=0.235\linewidth]{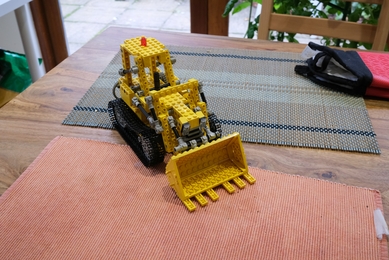} \\
    \midrule
    \viewlabel{View 1} &
    \includegraphics[width=0.235\linewidth]{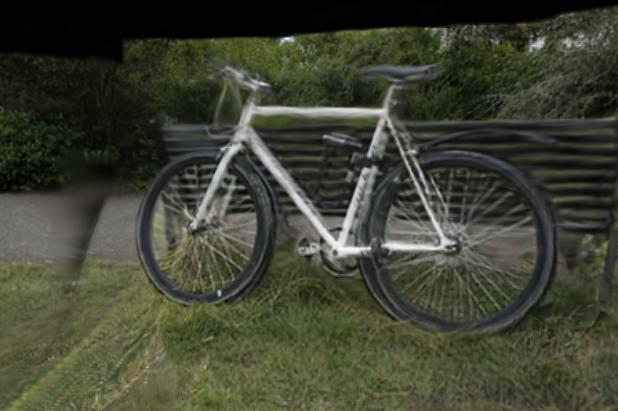} &
    \includegraphics[width=0.235\linewidth]{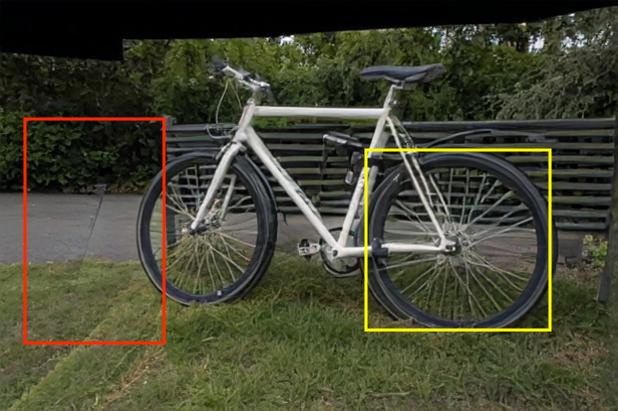} &
    \includegraphics[width=0.235\linewidth]{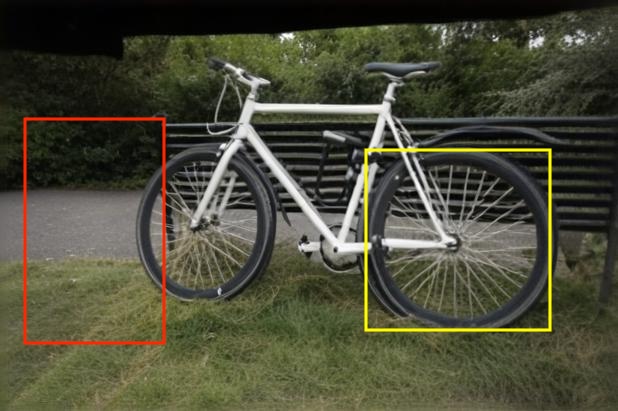} &
    \includegraphics[width=0.235\linewidth]{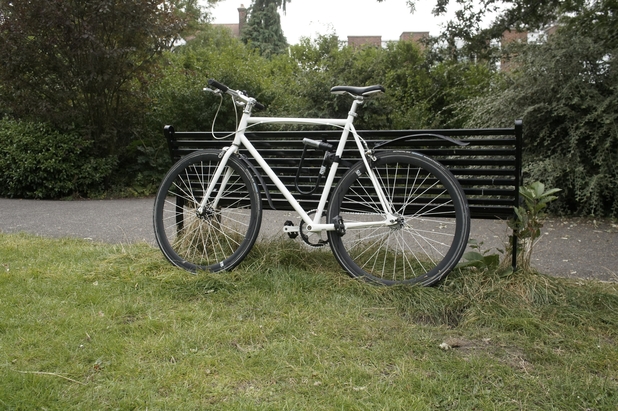} \\
    
    \viewlabel{View 2} &
    \includegraphics[width=0.235\linewidth]{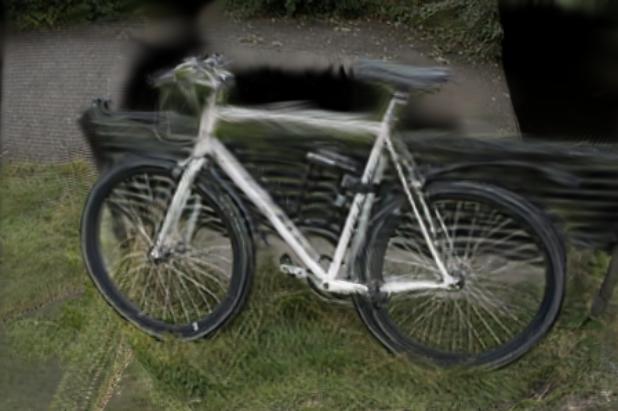} &
    \includegraphics[width=0.235\linewidth]{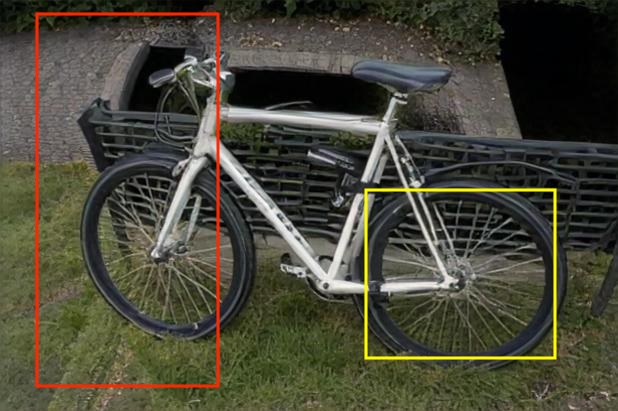} &
    \includegraphics[width=0.235\linewidth]{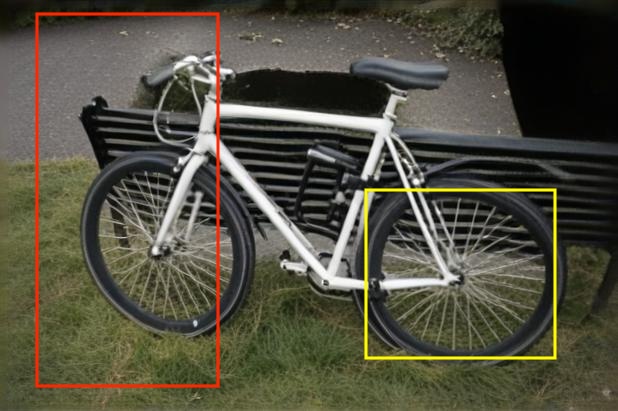} &
    \includegraphics[width=0.235\linewidth]{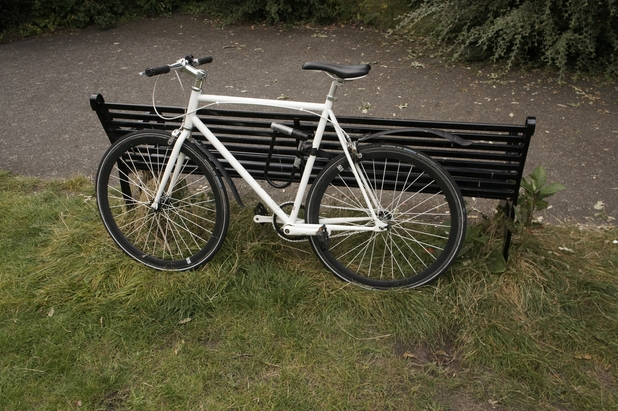} \\
            & \text{AnySplat} & \text{Difix3D+} & \text{Ours} & \text{GT}
    \end{tabular}
    \vspace{-10pt}
\captionof{figure}{\textbf{Comparison of multi-view refinement with feedforward renderings}. SyncFix provides coherent multi-view refinement that reduces the artifacts and maintains consistency. Notice the region highlighted in the boxes.}
\label{fig:mipnerf_exp}
\end{center}
\vspace{-10pt}
\end{figure}

Table~\ref{tab:quantitative_feedforward} reports quantitative results on AnySplat renderings, with Difix3D+, and with SyncFix. We gray out PSNR because the predicted camera poses contain residual errors; the resulting pixel shifts (see Fig.~\ref{fig:mipnerf_exp}) make pixel-aligned metrics unreliable. Overall, SyncFix consistently improves perceptual quality over Difix3D+, achieving lower LPIPS and DSIM. More importantly, SyncFix achieves a significantly higher CVSC score than both Difix3D+ and the raw AnySplat renderings, indicating stronger cross-view consistency.

The qualitative comparisons in Fig.~\ref{fig:mipnerf_exp} support these findings. Difix3D+ shows view-dependent inconsistencies, e.g., around the table mat and the red cap in the Lego scene. A similar pattern is observed in the outdoor bicycle scene: Difix3D+ introduces non-uniform residual artifact patterns around the road and grass regions. In contrast, SyncFix has consistent refinements across views.

\begin{table}[htpb!]
\centering
\setlength{\tabcolsep}{3pt}
\caption{
Quantitative comparison on the MipNeRF 360 dataset for refining feedforward Gaussian Splatting renderings. SyncFix improves perceptual quality and cross-view semantic consistency over prior generative refinement methods. We gray out the PSNR as pixel shifts make the pixel-aligned metric unreliable.
$\uparrow$ indicates higher-is-better and $\downarrow$ indicates lower-is-better.
\textbf{Bold} denotes best results and \underline{underline} denotes second best.
}
\vspace{-10pt}
\begin{tabular}{l >{\columncolor{psnrgray}\color{psnrtext}}c c c c}
\toprule
& \multicolumn{4}{c}{MipNeRF 360} \\
\cmidrule(lr){2-5}
Method 
& PSNR$\uparrow$ & LPIPS$\downarrow$ & DSIM$\downarrow$ & CVSC$\uparrow$ \\
\midrule
AnySplat \cite{jiang2025anysplat}
& 13.58 & 0.454 & 0.232 & \underline{0.649} \\
\midrule
Difix3D+ \cite{wu2025difix3d+}
& 13.41 & \underline{0.448} & \underline{0.196} & 0.589 \\
SyncFix (ours)
& 13.51 & \textbf{0.430} & \textbf{0.189} & \textbf{0.686} \\
\bottomrule
\end{tabular}
\label{tab:quantitative_feedforward}
\end{table}

\subsection{Ablation Study}

We analyze the contribution of components through a series of ablations on the DL3DV test set (Table~\ref{tab:ablation}). While the train-in-loop strategy in Difix3D brings moderate improvement, the CVSC score is still suboptimal to our multi-view synchronization and it requires access of testing-view poses during training. In addition, we augment Difix3D+ with multiple input views during inference by providing the same number of views as SyncFix. This allows us to evaluate whether supplying additional views is sufficient to improve consistency. The last two rows of Difix3D+ in Table~\ref{tab:ablation} show that performance drops, indicating that the model remains fundamentally limited by its single-view training objective and struggles in fusing multi-view information.

\begin{figure}[t!]
\begin{center}
\begin{minipage}{\linewidth}
    \centering
    \setlength{\tabcolsep}{1pt}      %
    \renewcommand{\arraystretch}{1.0} %
    \small
    \settoheight{\imgH}{\includegraphics[width=0.28\linewidth]{figures/teaser_abhay_arxiv_v2/difix_images_pretrained_ref/frame_00337_compare_teaser/frame_00337_compare_teaser_annotated.jpg}}
    
    \newcommand{\viewlabel}[1]{%
      \parbox[b][\imgH][c]{2.0em}{\centering\rotatebox{90}{#1}}%
    }
    \begin{tabular}{c c c c}
    \viewlabel{View 1} &
    \includegraphics[width=0.31\linewidth]{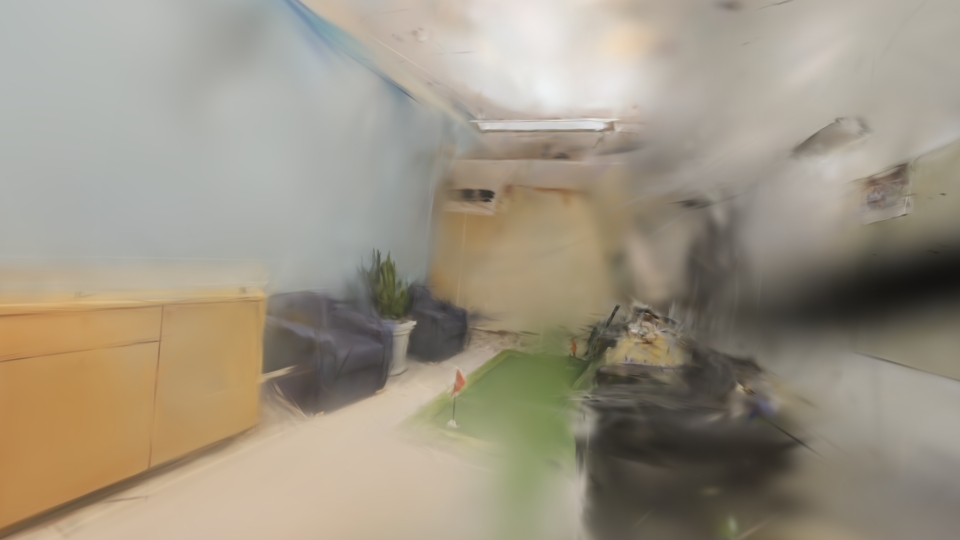} &
    \includegraphics[width=0.31\linewidth]{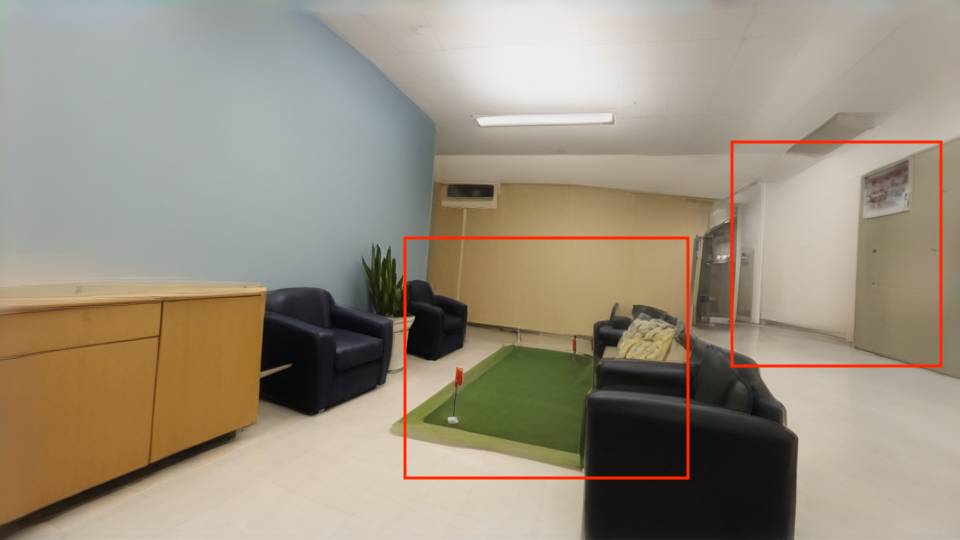} &
    \includegraphics[width=0.31\linewidth]{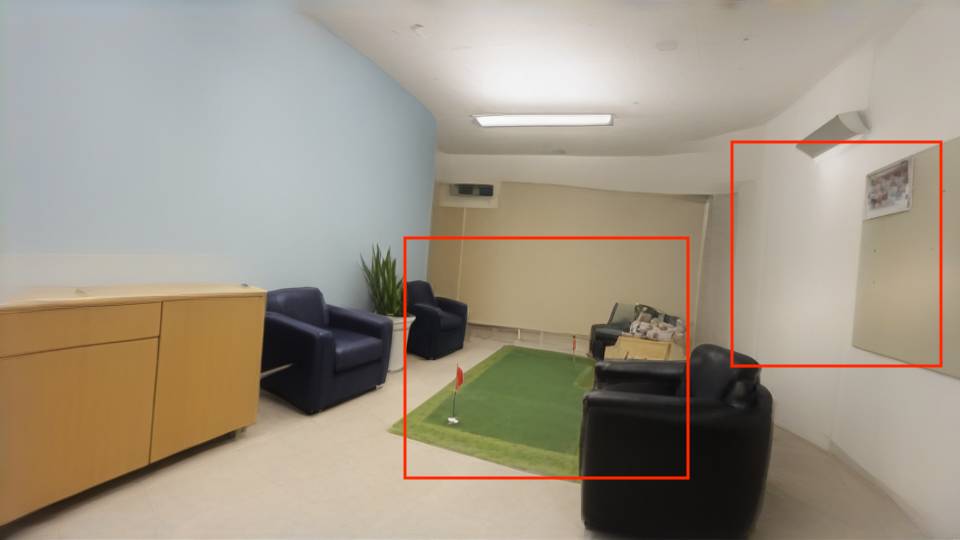} \\
    
    \viewlabel{View 2} &
    \includegraphics[width=0.31\linewidth]{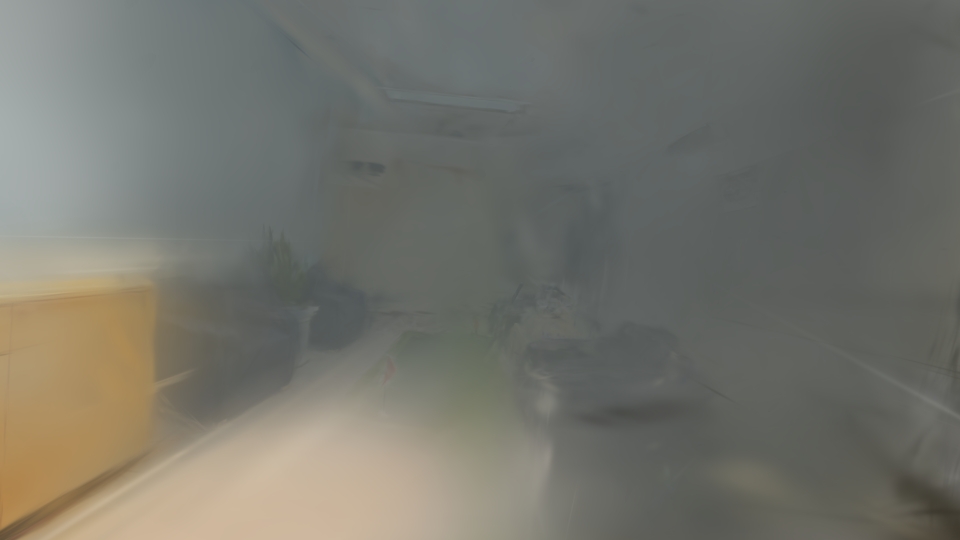} &
    \includegraphics[width=0.31\linewidth]{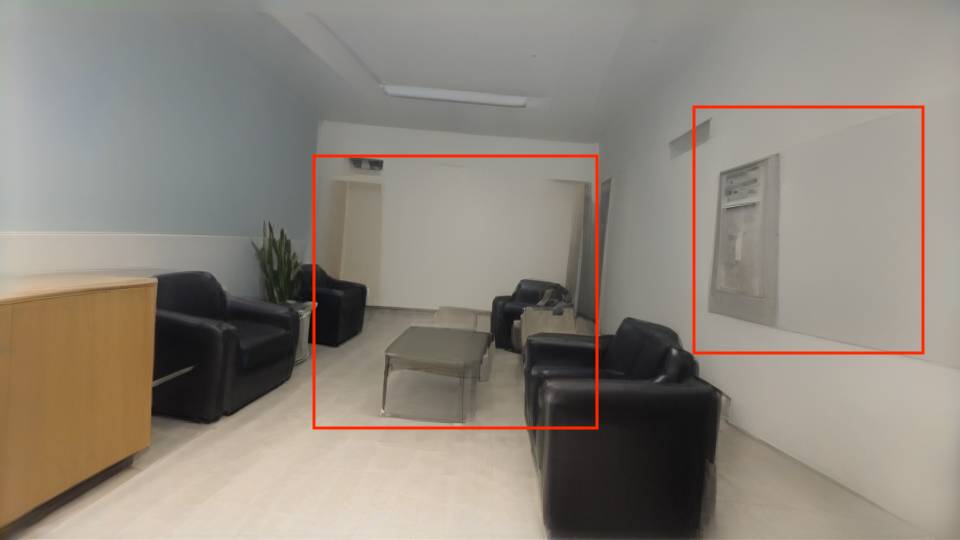} &
    \includegraphics[width=0.31\linewidth]{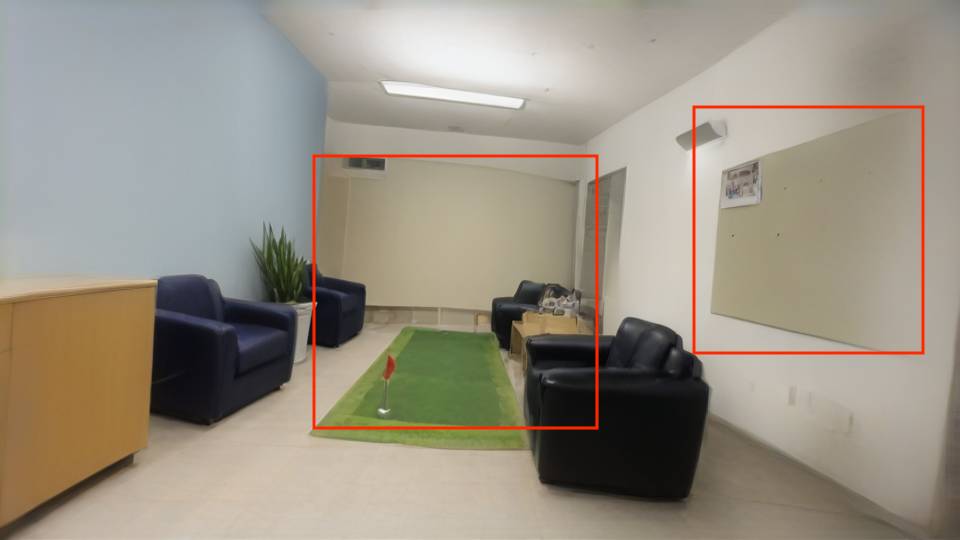} \\
    
         & 3DGS & Ours(single-view) & Ours(multi-view) \\
         
    \end{tabular}
\captionof{figure}{
\textbf{Ablation on multi-view synchronization.}
We compare SyncFix with and without joint multi-view refinement using the same latent-bridge matching training protocol and without any reference images.
Processing each view independently (Ours, single-view) improves individual renderings but produces inconsistent scene structure across viewpoints.
In particular, the green carpet on the floor and the drawing board on the right wall disappear or are rendered inconsistently across the two views (highlighted in red).
Because SyncFix (multi-view) jointly refines the views, it produces coherent geometry and consistent scene layout across viewpoints.
}
\label{fig:ablation}
\end{minipage}
\end{center}
\vspace{-10pt}
\end{figure}

We then evaluate variants of SyncFix. A single-view version of our model already improves image quality compared to Difix3D+, demonstrating the benefit of the latent bridge formulation. Introducing joint multi-view refinement further improves performance across all metrics. In particular, cross-view semantic consistency (CVSC) increases from 0.821 to 0.862, confirming that joint training encourages the model to agree on scene structure across viewpoints. Qualitative examples in Fig.~\ref{fig:ablation} illustrate this effect: the single-view model produces inconsistent hallucinations between views, whereas the multi-view model maintains spatial and semantic coherence. We also find that by supplementing reference views, the textures and details are more aligned with ground truth in the generated content by SyncFix. We note that the multi-view inference setting is five degraded views plus five closest training images as reference views for Difix3D+ and SyncFix. In our supplementary materials, we show that the multi-view consistency improves when adding more views during the refinement.

\begin{table}[t!]
\centering
\setlength{\tabcolsep}{3pt}
\caption{
Ablation study on the DL3DV test set evaluating the effect of joint multi-view refinement. We also ablate with and without reference views.
Simply applying a multi-view input to Difix3D+ provides limited benefit, while our joint training substantially improves both image quality and cross-view consistency. 
$\uparrow$ indicates higher-is-better and $\downarrow$ indicates lower-is-better.
}
\vspace{-10pt}
\resizebox{0.85\linewidth}{!}{
\begin{tabular}{lccccc}
\toprule
Method & PSNR$\uparrow$ & LPIPS$\downarrow$ & DSSIM$\downarrow$ & FID$\downarrow$ & CVSC$\uparrow$ \\
\midrule
3DGS \cite{kerbl20233d}
& 15.94 & 0.454 & 0.297 & 80.8 & 0.875 \\
\midrule
Difix3D \cite{wu2025difix3d+}
& 16.14 & 0.445 & 0.285 & 76.6 & 0.853 \\
Difix3D+ \cite{wu2025difix3d+} & 16.17 & 0.343 & 0.135 & 16.7 & 0.850 \\
Difix3D+ (multi-view inference) & 16.08 & 0.370 & 0.163 & 29.28 & 0.827 \\
\midrule
Ours (single-view, w/o reference) & 16.26 & 0.347 & 0.142 & 22.2 & 0.821 \\
Ours (multi-view, w/o reference) & 16.45 & 0.334 & 0.129 & 22.5 & 0.862 \\
\textbf{Ours (multi-view, with reference)} & \textbf{16.94} & \textbf{0.305} & \textbf{0.099} & \textbf{17.5} & \textbf{0.880} \\
\bottomrule
\end{tabular}
}
\label{tab:ablation}
\vspace{-10pt}
\end{table}

\begin{figure}[t!]
    \centering
    \begin{subfigure}[t]{0.49\linewidth}
        \centering
        \includegraphics[width=\linewidth]{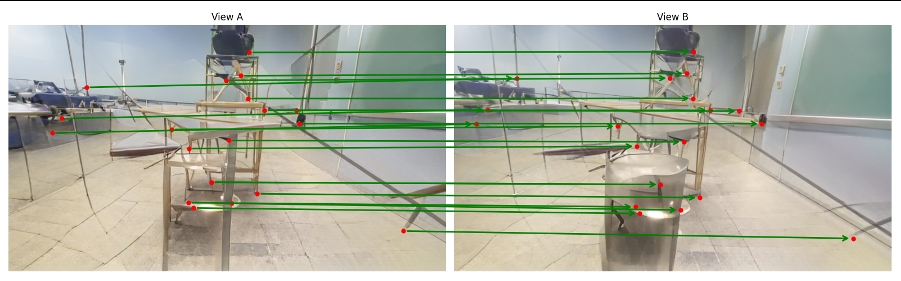}
        \caption{Matches of Difix3D+ refined renderings}
        \label{fig:difix_match}
    \end{subfigure}\hfill
    \begin{subfigure}[t]{0.49\linewidth}
        \centering
        \includegraphics[width=\linewidth]{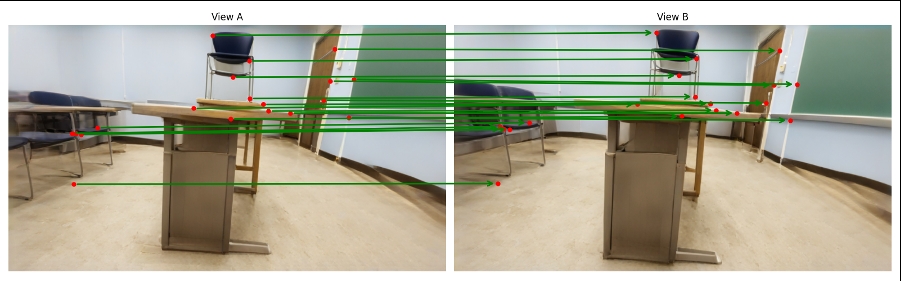}
        \caption{Matches of SyncFix refined renderings}
        \label{fig:syncfix_match}
    \end{subfigure}
    \vspace{-5pt}
    \caption{\textbf{Cross-view geometric consistency.}
Keypoint correspondences between two refined views. 
Difix3D+ produces inconsistent refinements, resulting in irregular correspondences across views. 
SyncFix preserves the underlying scene geometry, producing coherent cross-view matches. 
For visualization clarity, we display 20 sampled correspondences. Matches computed using RaCo + LightGlue.}
    \label{fig:matched_keys}
    \vspace{-10pt}
    
\end{figure}

\section{Discussion}
Our experiments suggest that refining views independently with a 2D generative prior is fundamentally misaligned with the multi-view nature of 3D reconstruction. Marginal refinement can improve individual images, but the resulting hallucinations often differ across viewpoints. The reconstruction process must then harmonize these incompatible predictions, which frequently leads to drift and unstable geometry. SyncFix addresses this by coupling views during generation, forcing the model to agree on structure before the images are produced.

An interesting observation is that the multi-view consistency achieved by SyncFix arises purely from cross-view attention. The model does not receive explicit geometric supervision such as camera poses, epipolar constraints, or reprojection losses. Instead, synchronization emerges from learned correlations between views within a shared latent space. This suggests that generative models contain sufficient structure to align views through attention alone, although incorporating explicit geometric constraints may further improve robustness.

The results also highlight limitations of current evaluation protocols. Standard image metrics such as PSNR, LPIPS, and DSSIM assume that a single ground-truth image represents the only correct solution. However, sparse-view 3D reconstruction is fundamentally under-constrained: large portions of a scene may be poorly observed or entirely unseen in the training views. In such cases, generative refinement methods must infer plausible geometry and textures consistent with the visible context. SyncFix can produce coherent and consistent structures even when the predicted appearance deviates from the ground-truth rendering. Pixel-wise metrics penalize these plausible reconstructions because they measure strict image correspondence rather than semantic or geometric consistency. As Figure \ref{fig:matched_keys} shows, consistent refined renderings receive higher patch feature correlation at matched keypoints. Developing evaluation protocols that better capture multi-view agreement and perceptual plausibility remains an important direction for future work.

Finally, although SyncFix is trained using view pairs, the architecture naturally generalizes to larger sets of views at inference. Additional viewpoints provide stronger constraints and typically improve stability. This suggests that pairwise training is sufficient to learn consistent multi-view refinement dynamics.

\begin{figure}[t!]
    \centering
    \setlength{\tabcolsep}{1pt}      %
    \renewcommand{\arraystretch}{1.0} %
    \small
    
    \settoheight{\imgH}{\includegraphics[width=0.28\linewidth]{figures/teaser_abhay_arxiv_v2/difix_images_pretrained_ref/frame_00337_compare_teaser/frame_00337_compare_teaser_annotated.jpg}}
    
    \newcommand{\viewlabel}[1]{%
      \parbox[b][\imgH][c]{2.0em}{\centering\rotatebox{90}{\text{#1}}}%
    }
    
    \begin{tabular}{c c c c}
    \viewlabel{View 1} &
    \includegraphics[width=0.30\linewidth]{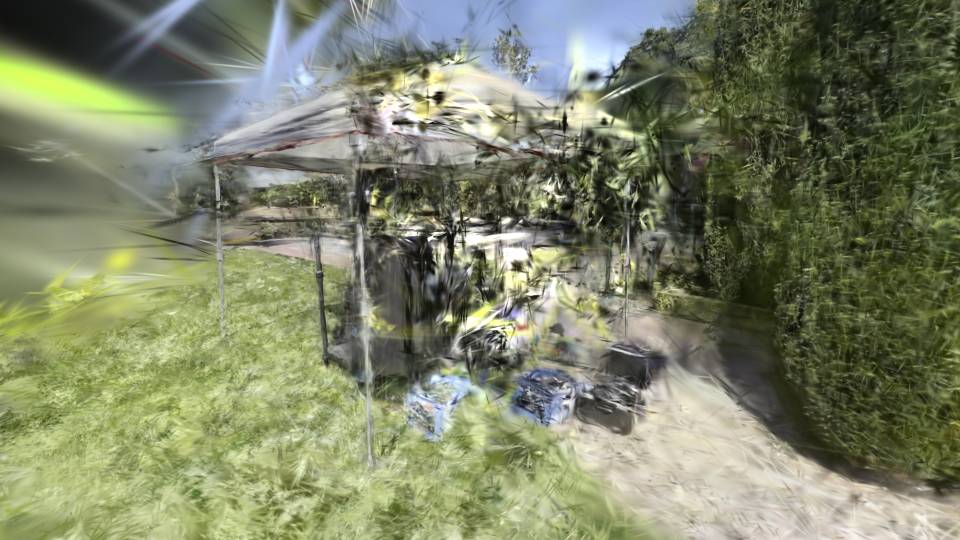} &
    \includegraphics[width=0.30\linewidth]{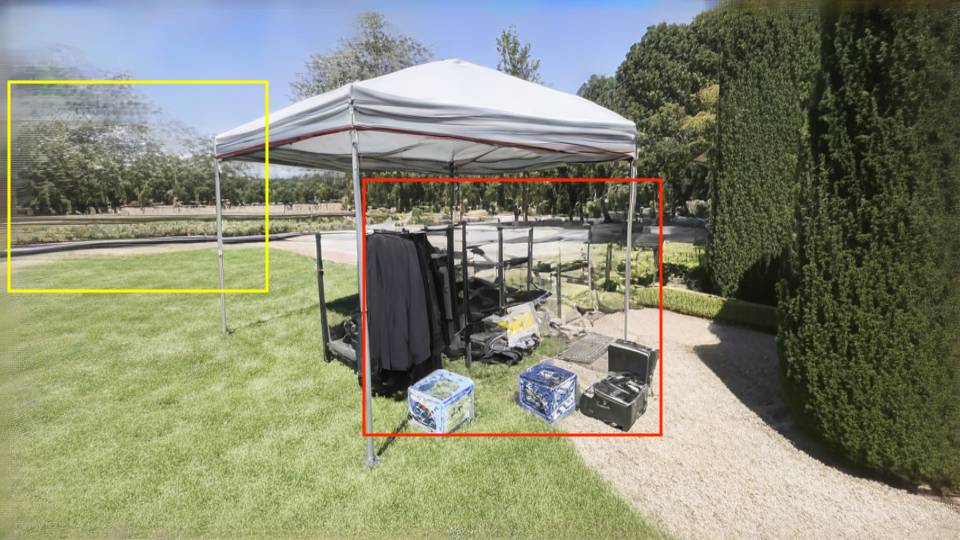} &
    \includegraphics[width=0.30\linewidth]{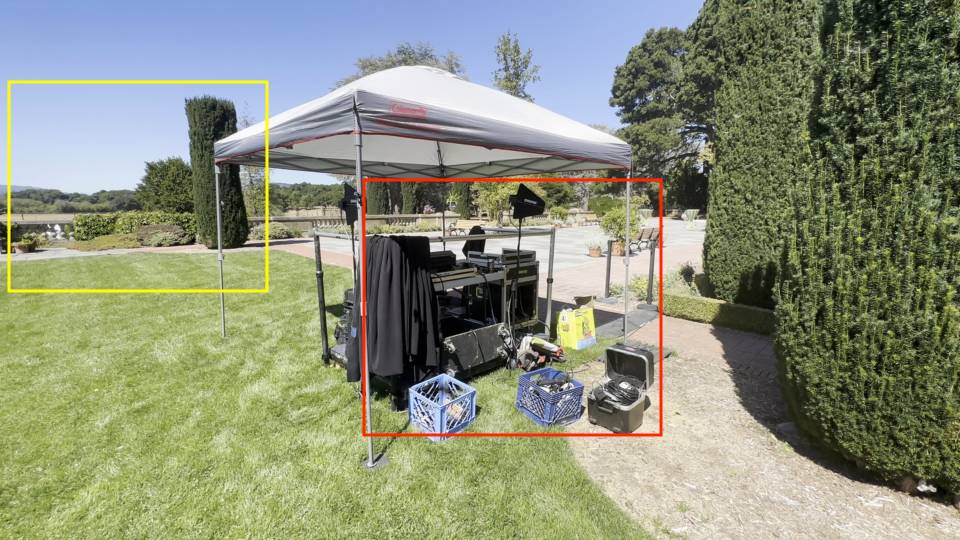} \\
    
    \viewlabel{View 2} &
    \includegraphics[width=0.30\linewidth]{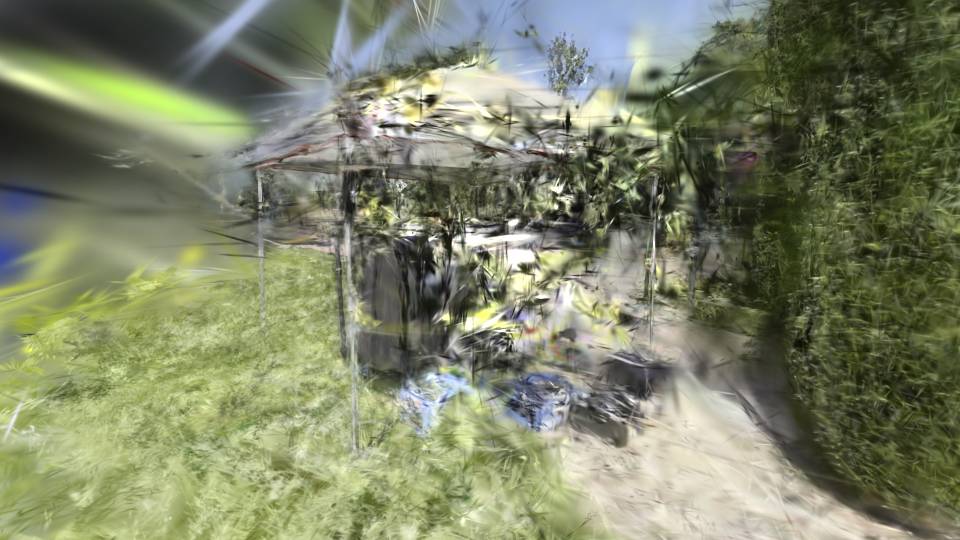} &
    \includegraphics[width=0.30\linewidth]{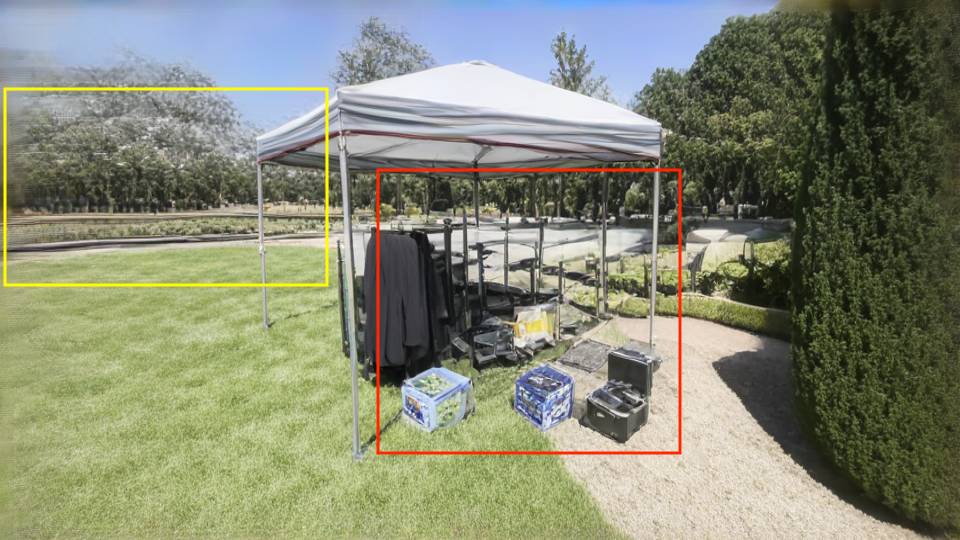} &
    \includegraphics[width=0.30\linewidth]{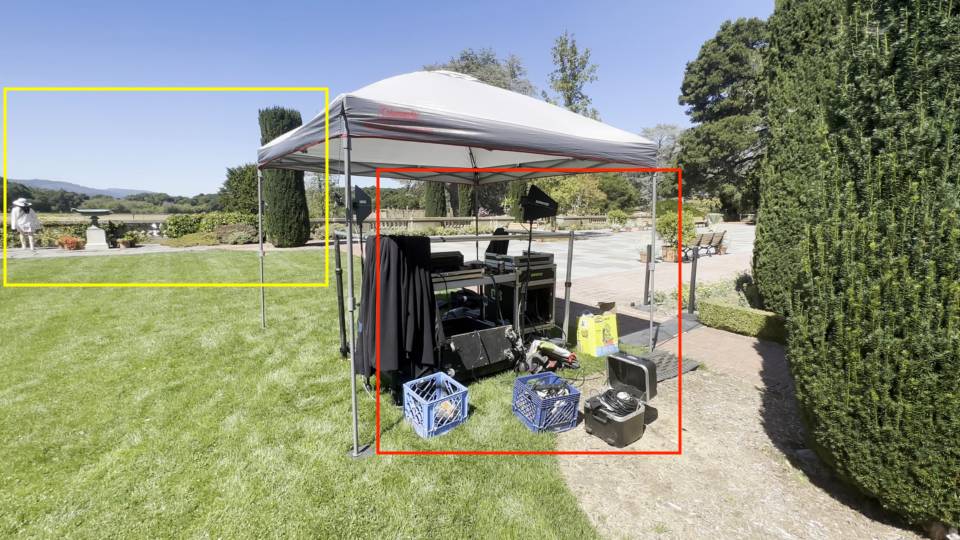} \\
    \midrule
    \viewlabel{View 1} &
    \includegraphics[width=0.30\linewidth]{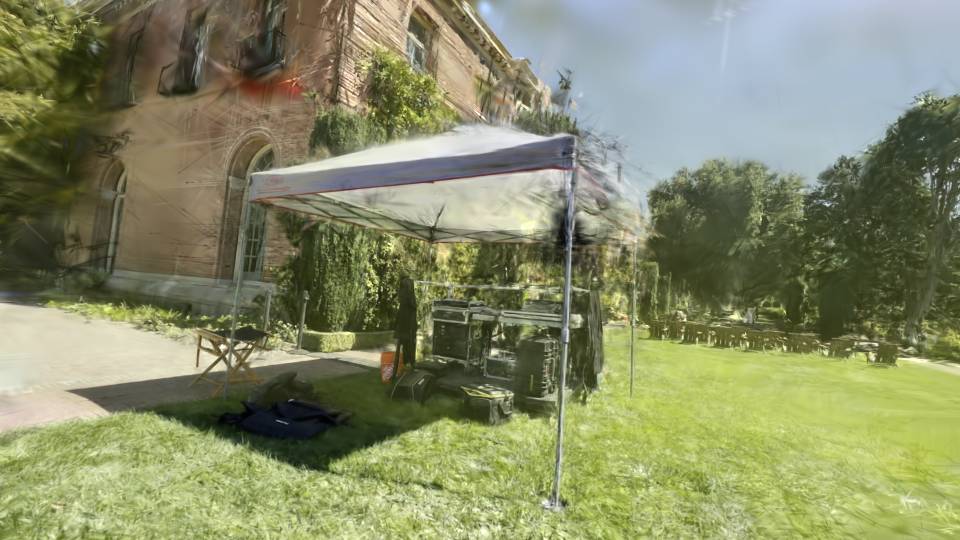} &
    \includegraphics[width=0.30\linewidth]{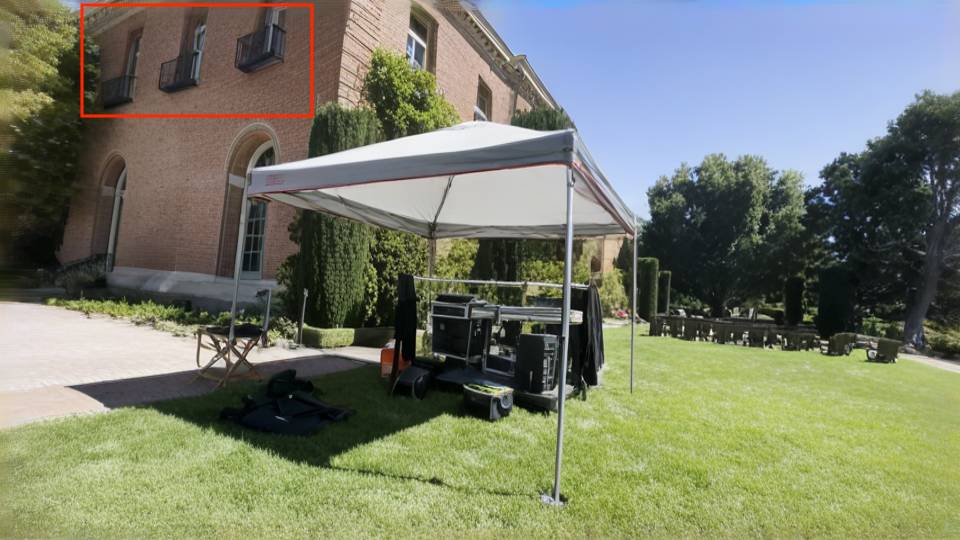} &
    \includegraphics[width=0.30\linewidth]{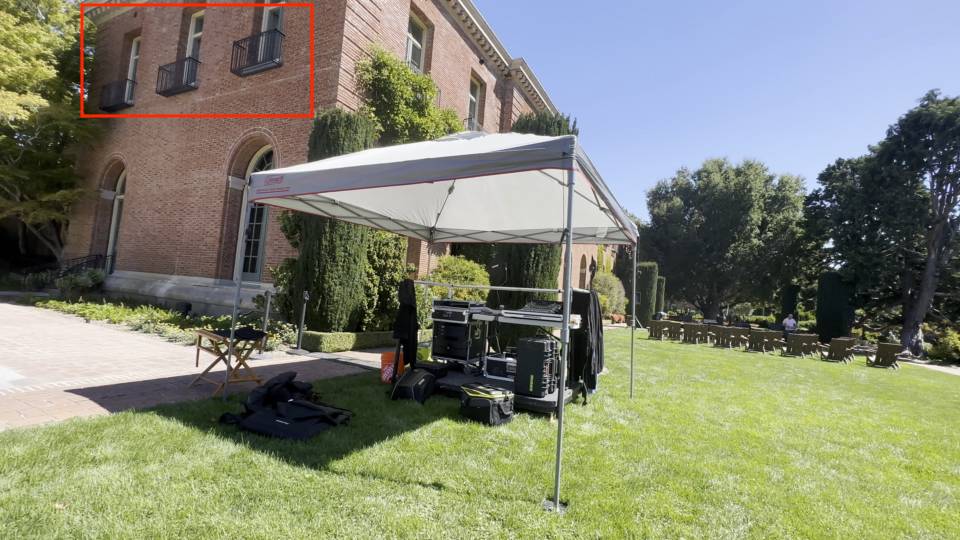} \\
    
    \viewlabel{View 2} &
    \includegraphics[width=0.30\linewidth]{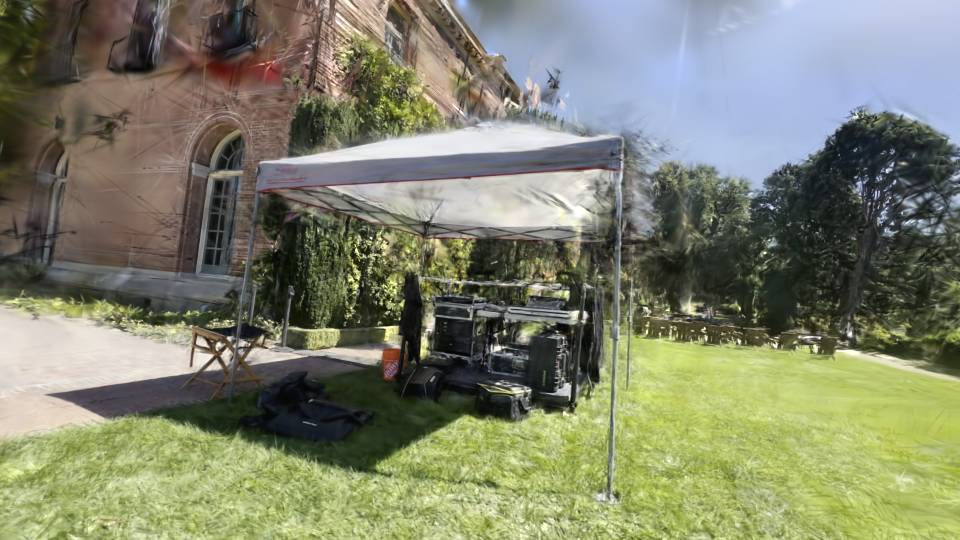} &
    \includegraphics[width=0.30\linewidth]{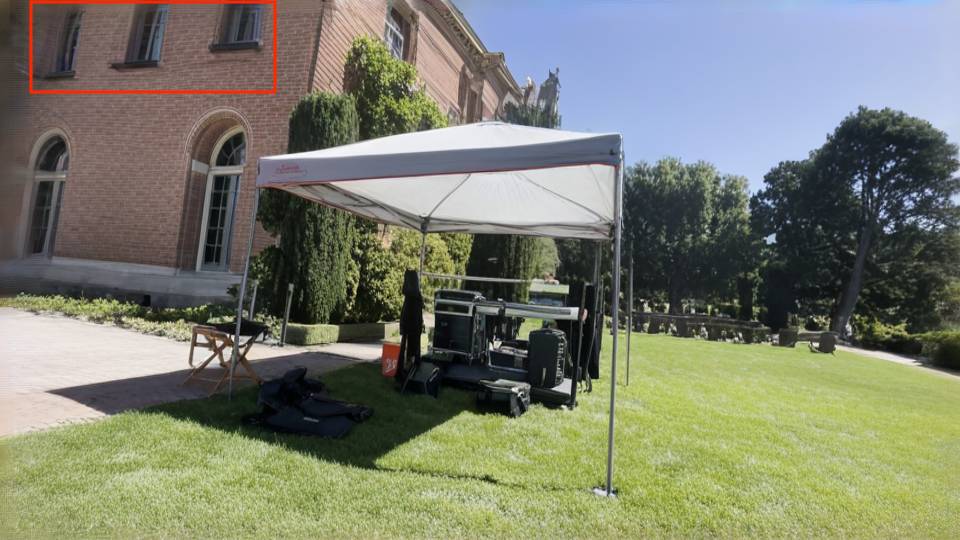} &
    \includegraphics[width=0.30\linewidth]{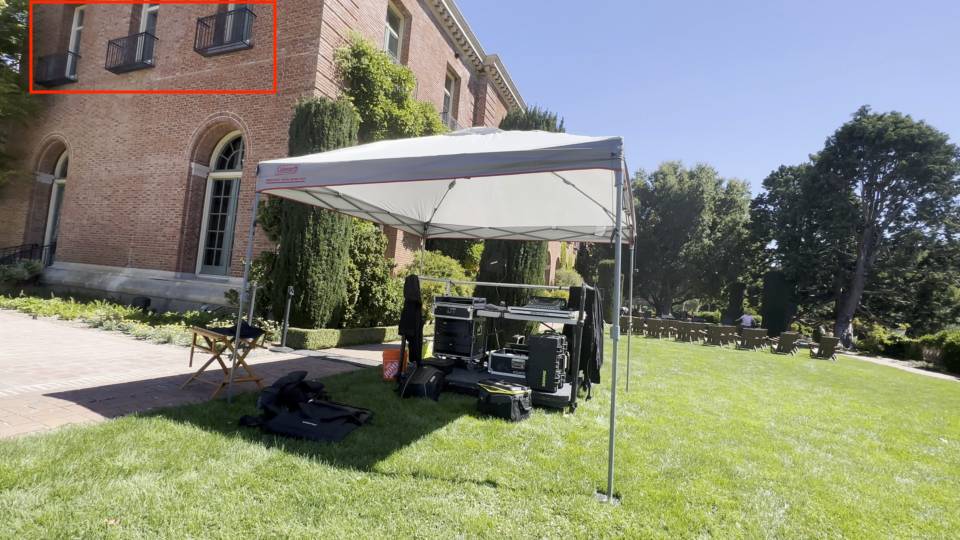} \\
            & \text{3DGS} & \text{SnycFix} & \text{GT}
    \end{tabular}
\captionof{figure}{\textbf{Limitation}. SyncFix may synthesize content that mismatches the ground-truth information. The resulting pixel-wise metrics are worse for this scene despite being multi-view consistent. Multi-view consistency can break down when views are refined in isolation. As seen in our SyncFix results, the windowsills in the bottom views become inconsistent because the independent network passes fail to synchronize geometric details across different perspectives. Note that the level of corruption for the top and bottom panels is different in the 3DGS views.}
\label{fig:limitation}
\end{figure}

\paragraph{Limitations.}
SyncFix assumes that the input views share sufficient overlap to establish meaningful correspondences during joint refinement. When viewpoints are extremely sparse or observe largely disjoint regions of the scene, the synchronization signal becomes weak and the model may revert to view-dependent hallucinations. The approach also relies on the quality of the underlying reconstruction; if the geometry is severely corrupted or missing large structures, the generative prior may produce plausible but incorrect completions. Finally, the generative prior may introduce biases toward visually plausible textures and structures that are consistent with the training distribution but not necessarily faithful to the true scene. Developing generative refinement models that integrate explicit geometric reasoning and more principled evaluation protocols remains an important direction for future work.

\section{Conclusion}
We presented SyncFix, a framework for refining sparse-view 3D reconstructions using a joint multi-view generative prior. The method formulates refinement as latent bridge matching between distorted and clean renderings and synchronizes the latent trajectories of multiple views through cross-view attention. This design allows the model to enforce multi-view agreement during generation rather than relying on post-hoc reconciliation by the 3D reconstruction.
Experiments demonstrate improved perceptual quality and stronger cross-view consistency across several reconstruction settings. More broadly, the results suggest that generative priors for visual scenes should operate on joint multi-view distributions rather than independent view marginals. 
Finally, visual scenes are composed of multi-view objects; generative models that operate on them should reflect this structure.

\vspace{-10pt}
\section*{Acknowledgement}
\vspace{-5pt}
This research is based upon work supported by the Office of the Director of National Intelligence (ODNI), Intelligence Advanced Research Projects Activity (IARPA), via IARPA R\&D Contract No. 140D0423C0076. The views and conclusions contained herein are those of the authors and should not be interpreted as necessarily representing the official policies or endorsements, either expressed or implied, of the ODNI, IARPA, or the U.S. Government. The U.S. Government is authorized to reproduce and distribute reprints for Governmental purposes notwithstanding any copyright annotation thereon.

\bibliographystyle{splncs04}
\bibliography{main}

\newpage
\appendix
\section{Technical Appendices and Supplementary Material}
\subsection{}section{Number of Views Analysis}
\label{sec:view_analysis}
 SyncFix is trained on view pairs, but our permutation-invariant latent coupling enables it to generalize to and naturally support inference with an arbitrary number of input views, including degraded views and reference views. We therefore perform a systematic study to quantify how performance varies with the number of views provided at test time.

Concretely, we vary the number of degraded views from 1 to 5 and the number of reference views from 1 to 5 during inference, yielding 25 configurations in total. We report the results as 5×5 matrices, which facilitate direct comparison of performance trends across different view compositions for different metrics.

\begin{figure}[htbp]
\centering
\includegraphics[width=0.8\textwidth]{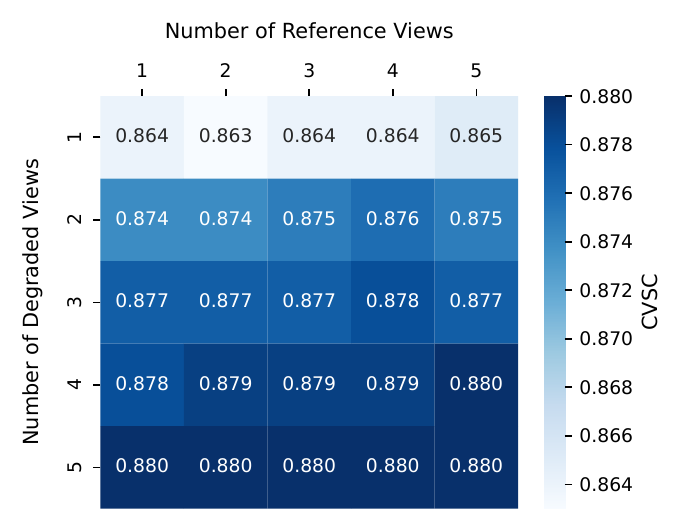}
\caption{
\textbf{CVSC scores as the number of degraded views and reference views changes}. Joint refinement of more views leads to better multi-view consistency.
}
\label{fig:cvsc_matrice}
\end{figure}

\begin{figure}[htbp]
\centering
\includegraphics[width=0.8\textwidth]{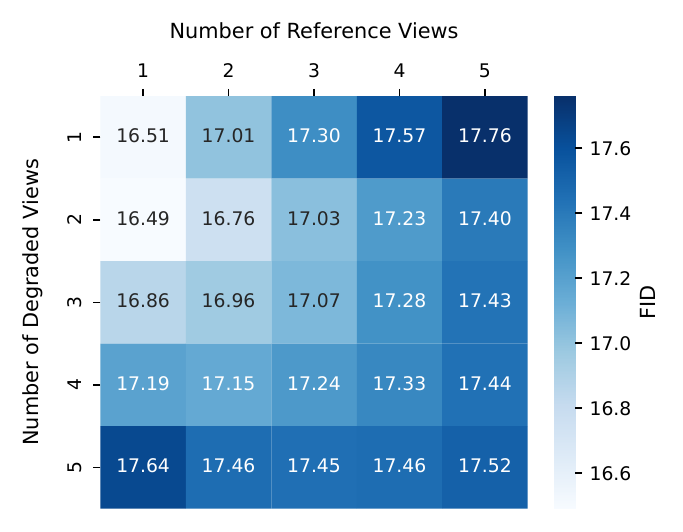}
\caption{
\textbf{FID as the number of degraded views and reference views changes}. Increasing the total number of input images increases the FID as attention is distributed between views.
}
\label{fig:fid_matrice}
\end{figure}

\begin{figure}[htbp]
\centering
\includegraphics[width=0.8\textwidth]{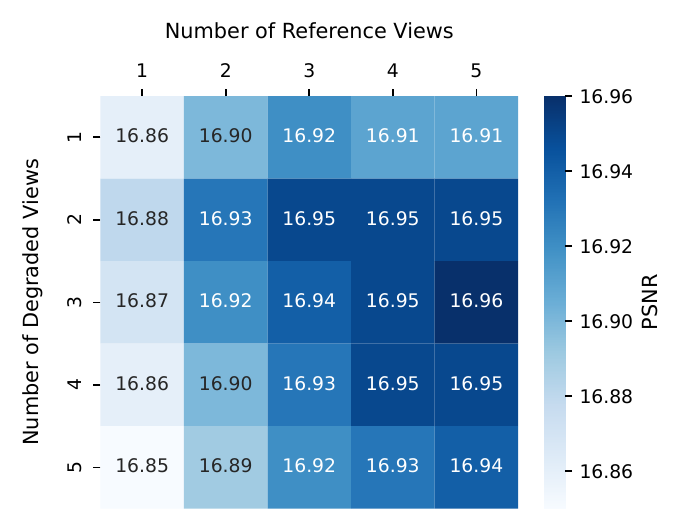}
\caption{
\textbf{PSNR as the number of degraded views and reference views changes}. Adding more reference views helps with the local details, thus higher PSNR, and plateaus with more views.
}
\label{fig:psnr_matrice}
\end{figure}

We draw three insights:
\begin{enumerate}
    \item Fig. \ref{fig:cvsc_matrice} reveals a clear trend that the CVSC score improves with more degraded views, regardless of the number of reference views. The biggest gain occurs when moving from single-view to two-view inference. This finding supports our claim that jointly refining more views yields better cross-view consistency.
    \item Fig. \ref{fig:fid_matrice} shows that FID tends to increase as either the number of degraded views or the number of reference views grows. We attribute this behavior to the attention being distributed across a larger set of inputs during joint conditioning, which can introduce a moderate averaging/smoothing effect.
    \item Fig. \ref{fig:psnr_matrice} indicates that increasing the number of input views, particularly the reference images, leads to a higher PSNR and plateaus with more views. We observe that the local details are better recovered in the regions that are visible in the reference views.
\end{enumerate}

We report the results with 5 degraded views and 5 reference views in our main text.

\subsection{}section{More Qualitative Comparison}
\label{sec:quali}

%%%%%%new code to render crops and images for many scenes and views

% =====================================================
% Image helper
% =====================================================
\newcommand{\lbmimg}[1]{\includegraphics[width=0.19\textwidth]{#1}}

% =====================================================
% Vertical row label
% =====================================================
\newcommand{\lbmlabel}[1]{\rotatebox{90}{\text{#1}}}

% =====================================================
% Full render path
% =====================================================
% \newcommand{\lbmfull}[4]{%
% figures/for_lbm_supplimentary/#1/K_12/run_002/it_03000/#2/#3/#4.png
% }
\newcommand{\lbmfull}[4]{%
figures/for_lbm_supplimentary/#1/K_12/run_002/it_03000/#2/#3/#4_compare_v8/#4_compare_v8_annotated.jpg
}

% =====================================================
% Boxed crop path
% =====================================================
\newcommand{\lbmcrop}[5]{%
figures/for_lbm_supplimentary/#1/K_12/run_002/it_03000/#2/#3/#4_compare_v8/#4_compare_v8_#5_boxed.jpg
}

% =====================================================
% Row macro
% =====================================================
\newcommand{\lbmrow}[6]{%
\lbmlabel{#1} &
\lbmimg{#2} &
\lbmimg{#3} &
\lbmimg{#4} &
\lbmimg{#5} &
\lbmimg{#6} \\
}

% =====================================================
% Pair figure macro
% =====================================================
\newcommand{\lbmpairfigure}[6]{

\begin{figure}[htbp]
\centering

\setlength{\tabcolsep}{1pt}
\renewcommand{\arraystretch}{1.0}
\small

\begin{tabular}{c c c c c c}

& \text{3DGS} & \text{Fixer} & \text{Difix3D+} & \text{Ours} & \text{GT} \\

\lbmrow{View 1}
{\lbmfull{#1}{images}{#2}{#3}}
{\lbmfull{#1}{fixer}{#2}{#3}}
{\lbmfull{#1}{difix_images_pretrained_ref}{#2}{#3}}
{\lbmfull{#1}{lbm_images_ref_new_final_finetune3}{#2}{#3}}
{\lbmfull{#1}{gt}{#2}{#3}}

\lbmrow{View 2}
{\lbmfull{#1}{images}{#2}{#4}}
{\lbmfull{#1}{fixer}{#2}{#4}}
{\lbmfull{#1}{difix_images_pretrained_ref}{#2}{#4}}
{\lbmfull{#1}{lbm_images_ref_new_final_finetune3}{#2}{#4}}
{\lbmfull{#1}{gt}{#2}{#4}}

\lbmrow{View 1}
{\lbmcrop{#1}{images}{#2}{#3}{crop1}}
{\lbmcrop{#1}{fixer}{#2}{#3}{crop1}}
{\lbmcrop{#1}{difix_images_pretrained_ref}{#2}{#3}{crop1}}
{\lbmcrop{#1}{lbm_images_ref_new_final_finetune3}{#2}{#3}{crop1}}
{\lbmcrop{#1}{gt}{#2}{#3}{crop1}}

\lbmrow{View 2}
{\lbmcrop{#1}{images}{#2}{#4}{crop1}}
{\lbmcrop{#1}{fixer}{#2}{#4}{crop1}}
{\lbmcrop{#1}{difix_images_pretrained_ref}{#2}{#4}{crop1}}
{\lbmcrop{#1}{lbm_images_ref_new_final_finetune3}{#2}{#4}{crop1}}
{\lbmcrop{#1}{gt}{#2}{#4}{crop1}}

\end{tabular}

\vspace{-5pt}

\caption{#5}

\label{#6}

\vspace{-10pt}

\end{figure}
}

% =====================================================
% Scene f102
% =====================================================

\lbmpairfigure
{f102f93ee1b7304a9a99330b98a49229cf6111da9e764620ae96fae0622040fa}
{pair1}
{frame_00096}
{frame_00101}
{\textbf{Visual comparison on scene \texttt{f102f93e...}, Pair 1.}
The baseline 3DGS renderings exhibit strong blur and missing structure in the chair and surrounding objects. 
Fixer improves stability but suppresses fine details, while Difix3D+ produces sharper results with inconsistent textures across views. 
Our method reconstructs clearer chair geometry and preserves the object boundaries more faithfully, while maintaining consistent appearance across viewpoints.}
{fig:lbm_f102_pair1}

\lbmpairfigure
{f102f93ee1b7304a9a99330b98a49229cf6111da9e764620ae96fae0622040fa}
{pair2}
{frame_00135}
{frame_00140}
{\textbf{Visual comparison on scene \texttt{f102f93e...}, Pair 2.}
The cropped regions emphasize the chair structure, where baseline methods either blur the geometry or introduce inconsistent textures. 
Our method restores sharper edges and produces a more coherent object appearance across the two views. Note that the corruption rates are different from Fig 6.}
{fig:lbm_f102_pair2}

% =====================================================
% Scene e0a547
% =====================================================

\lbmpairfigure
{e0a5470d1f203b42e58ef6513deefbc52f9720680080c448f6d29e162c9b7378}
{pair1}
{frame_00036}
{frame_00060}
{\textbf{Visual comparison on scene \texttt{e0a5470d...}, Pair 1.}
In the doorway region, the baseline renderings show severe blur and inconsistent geometry. 
Fixer produces smoother results but removes structural details, while Difix3D+ partially restores textures with view-dependent artifacts. 
Our method recovers cleaner structural edges and maintains a consistent appearance across both viewpoints.}
{fig:lbm_e0a5_pair1}

\lbmpairfigure
{e0a5470d1f203b42e58ef6513deefbc52f9720680080c448f6d29e162c9b7378}
{pair2}
{frame_00096}
{frame_00101}
{\textbf{Visual comparison on scene \texttt{e0a5470d...}, Pair 2.}
The boxed crops highlight brick and railing structures that are difficult to reconstruct from sparse observations. 
Our method restores sharper geometric detail and more consistent textures compared with Fixer and Difix3D+. Note that the level of corruption is different in the 3DGS views compared to Fig 9.}
{fig:lbm_e0a5_pair2}

% =====================================================
% Scene eba3
% =====================================================

\lbmpairfigure
{eba3bdcb9304819a1cb05d385378a5a3703caceaf5fd6b4e28401659c761db61}
{pair1}
{frame_00166}
{frame_00171}
{\textbf{Visual comparison on scene \texttt{eba3bdcb...}, Pair 1.}
The baseline 3DGS renderings contain strong blur and missing geometry around the wall fixtures. 
Fixer produces smoother results but loses structural detail, whereas Difix3D+ introduces artifacts around object boundaries. 
Our method reconstructs clearer object shapes and preserves the layout of switches and outlets more faithfully.}
{fig:lbm_eba3_pair1}

\lbmpairfigure
{eba3bdcb9304819a1cb05d385378a5a3703caceaf5fd6b4e28401659c761db61}
{pair2}
{frame_00097}
{frame_00102}
{\textbf{Visual comparison on scene \texttt{eba3bdcb...}, Pair 2.}
The selected crops illustrate improvements in structural clarity and artifact removal. 
Our method produces sharper boundaries and a more accurate layout of the wall fixtures while maintaining a consistent appearance across views. Note that the level of corruption is different in the 3DGS views compared to Fig 10.}
{fig:lbm_eba3_pair2}

\end{document}

\title{SyncFix Supplemental Material} 

\titlerunning{SyncFix}

\maketitle

\section{Number of Views Analysis}
\label{sec:view_analysis}
 SyncFix is trained on view pairs, but our permutation-invariant latent coupling enables it to generalize to and naturally support inference with an arbitrary number of input views, including degraded views and reference views. We therefore perform a systematic study to quantify how performance varies with the number of views provided at test time.

Concretely, we vary the number of degraded views from 1 to 5 and the number of reference views from 1 to 5 during inference, yielding 25 configurations in total. We report the results as 5×5 matrices, which facilitate direct comparison of performance trends across different view compositions for different metrics.

\begin{figure}[htbp]
\centering
\includegraphics[width=0.8\textwidth]{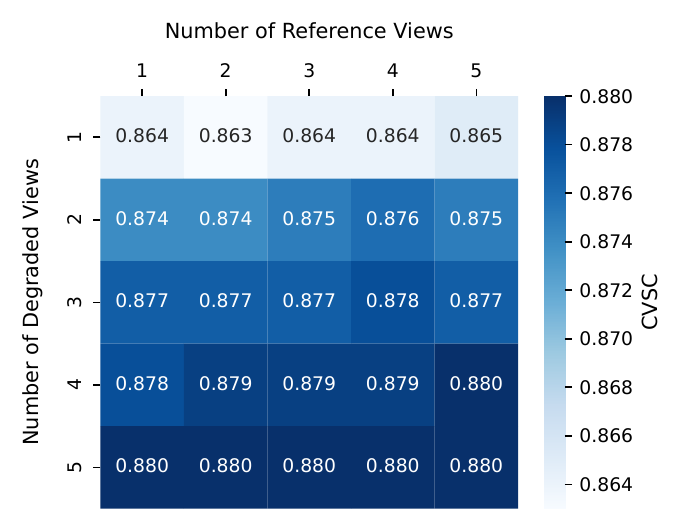}
\caption{
\textbf{CVSC scores as the number of degraded views and reference views changes}. Joint refinement of more views leads to better multi-view consistency.
}
\label{fig:cvsc_matrice}
\end{figure}

\begin{figure}[htbp]
\centering
\includegraphics[width=0.8\textwidth]{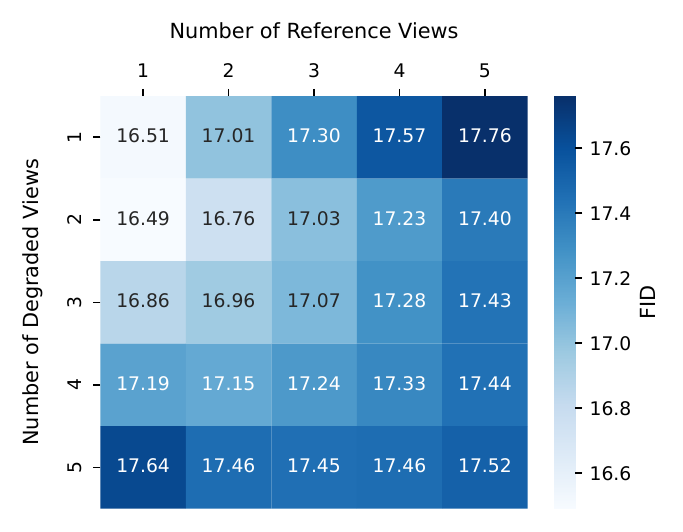}
\caption{
\textbf{FID as the number of degraded views and reference views changes}. Increasing the total number of input images increases the FID as attention is distributed between views.
}
\label{fig:fid_matrice}
\end{figure}

\begin{figure}[htbp]
\centering
\includegraphics[width=0.8\textwidth]{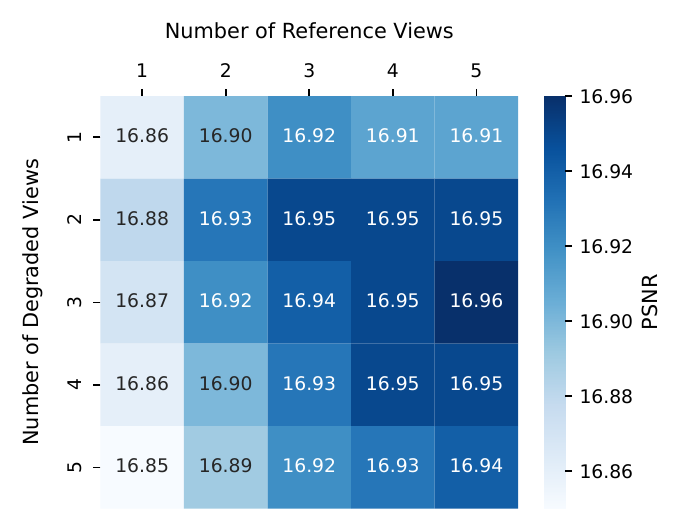}
\caption{
\textbf{PSNR as the number of degraded views and reference views changes}. Adding more reference views helps with the local details, thus higher PSNR, and plateaus with more views.
}
\label{fig:psnr_matrice}
\end{figure}

We draw three insights:
\begin{enumerate}
    \item Fig. \ref{fig:cvsc_matrice} reveals a clear trend that the CVSC score improves with more degraded views, regardless of the number of reference views. The biggest gain occurs when moving from single-view to two-view inference. This finding supports our claim that jointly refining more views yields better cross view consistency.
    \item Fig. \ref{fig:fid_matrice} shows that FID tends to increase as either the number of degraded views or the number of reference views grows. We attribute this behavior to the attention being distributed across a larger set of inputs during joint conditioning, which can introduce a moderate averaging/smoothing effect.
    \item Fig. \ref{fig:psnr_matrice} indicates that increasing the number of input views, particularly the reference images, leads to a higher PSNR and plateaus with more views. We observe that the local details are better recovered in the regions that are visible in the reference views.
\end{enumerate}

We report the results with 5 degraded views and 5 reference views in our main text.

\section{Generalization to Feedforward Method}
\label{sec:feedforward}
3D Gaussian Splatting (3DGS) \cite{kerbl20233d} achieves high-quality novel view synthesis with efficient rendering. However, it requires per-scene optimization as it overfits to each scene. To enable faster inference and leverage prior learned from large-scale data, recent feedforward splatting methods have gained traction by directly predicting 3D Gaussian parameters and depth from input images. This motivates evaluating whether SyncFix can generalize to refine multi-view renderings produced by feedforward pipelines. To this end, we conduct experiments using AnySplat \cite{jiang2025anysplat} on the Mip-NeRF 360 dataset \cite{barron2021mip}, which includes both indoor and outdoor scenes, to further assess the zero-shot capability of our method.

AnySplat supports feedforward rendering from uncalibrated images by employing separate decoder heads to predict 3D Gaussian parameters, depth, and camera poses. For each scene, we randomly sample 10 input images,  and the target views are selected sequentially while being far from the input views. We follow the evaluation protocol that we use VGGT \cite{wang2025vggt} to calibrate the input and target views in two passes. For pose alignment, we normalize camera poses by setting the first camera to the identity and estimate a relative scale factor to align camera translations across the two sets.

Table~\ref{tab:quantitative_feedforward} reports quantitative results on AnySplat renderings, with Difix3D+, and with SyncFix. We gray out PSNR because the predicted camera poses that contain residual errors; the resulting pixel shifts (see Fig.~\ref{fig:mipnerf_exp}) make pixel-aligned metrics unreliable. Overall, SyncFix consistently improves perceptual quality over Difix3D+, achieving lower LPIPS and DSIM. More importantly, SyncFix achieves a significantly higher CVSC score than both Difix3D+ and the raw AnySplat renderings, indicating stronger cross-view consistency.

The qualitative comparisons in Fig.~\ref{fig:mipnerf_exp} support these findings. Difix3D+ exhibits view-dependent inconsistencies, e.g., around the table mat and the red cap in the Lego scene. In contrast, SyncFix produces coherent refinements across viewpoints. A similar pattern is observed in the outdoor bicycle scene: Difix3D+ introduces non-uniform residual artifact patterns around the road and grass regions, whereas SyncFix delivers cleaner and more consistent refinements across views.

\begin{table}[t]
\centering
\setlength{\tabcolsep}{3pt}
\caption{
Quantitative comparison on the MipNeRF 360 dataset for refining feedforward Gaussian Splatting renderings.
SyncFix improves perceptual quality and cross-view semantic consistency over prior generative refinement methods.
$\uparrow$ indicates higher-is-better and $\downarrow$ indicates lower-is-better.
\textbf{Bold} denotes best results and \underline{underline} denotes second best.
}
\begin{tabular}{l >{\columncolor{psnrgray}\color{psnrtext}}c c c c}
\toprule
& \multicolumn{4}{c}{MipNeRF 360} \\
\cmidrule(lr){2-5}
Method 
& PSNR$\uparrow$ & LPIPS$\downarrow$ & DSIM$\downarrow$ & CVSC$\uparrow$ \\
\midrule
AnySplat \cite{jiang2025anysplat}
& 13.58 & 0.454 & 0.232 & \underline{0.649} \\
\midrule
Difix3D+ \cite{wu2025difix3d+}
& 13.41 & \underline{0.448} & \underline{0.196} & 0.589 \\
SyncFix (ours)
& 13.51 & \textbf{0.430} & \textbf{0.189} & \textbf{0.686} \\
\bottomrule
\end{tabular}
\label{tab:quantitative_feedforward}
\end{table}

\begin{figure}[t!]
\begin{center}
    \setlength{\tabcolsep}{1pt}      %
    \renewcommand{\arraystretch}{1.0} %
    \small
    
    \newlength{\imgH}
    \settoheight{\imgH}{\includegraphics[width=0.28\linewidth]{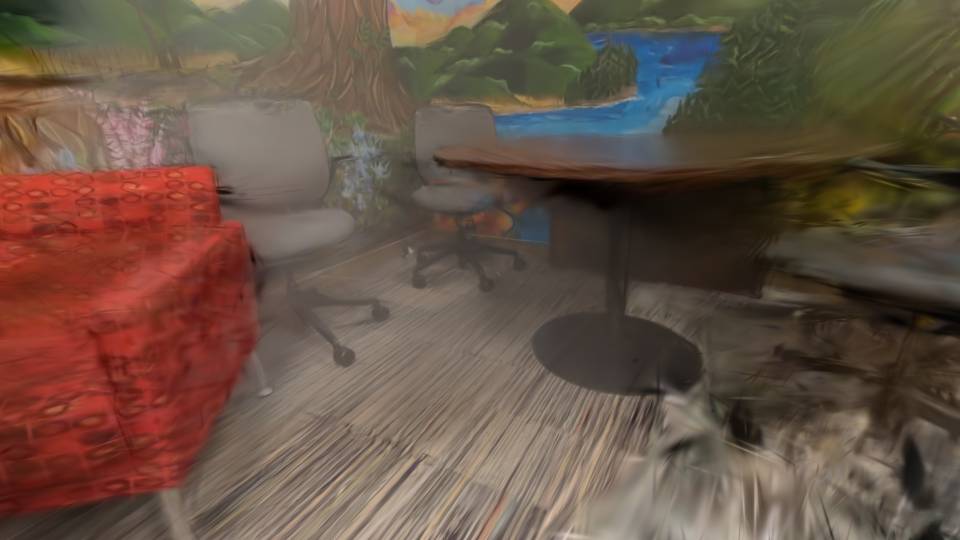}}
    
    \newcommand{\viewlabel}[1]{%
      \parbox[b][\imgH][c]{2.0em}{\centering\rotatebox{90}{\text{#1}}}%
    }
    
    \begin{tabular}{c c c c c}
        & \text{AnySplat} & \text{Difix3D+} & \text{Ours} & \text{GT} \\

    \viewlabel{View 1} &
    \includegraphics[width=0.23\linewidth]{figures/supplementary/mipnerf/bicycle/render/DSCF0656.JPG} &
    \includegraphics[width=0.23\linewidth]{figures/supplementary/mipnerf/bicycle/difix3d/DSCF0656.JPG} &
    \includegraphics[width=0.23\linewidth]{figures/supplementary/mipnerf/bicycle/ours/DSCF0656.JPG} &
    \includegraphics[width=0.23\linewidth]{figures/supplementary/mipnerf/bicycle/gt/DSCF0656.JPG} \\
    
    \viewlabel{View 2} &
    \includegraphics[width=0.23\linewidth]{figures/supplementary/mipnerf/bicycle/render/DSCF0680.JPG} &
    \includegraphics[width=0.23\linewidth]{figures/supplementary/mipnerf/bicycle/difix3d/DSCF0680.JPG} &
    \includegraphics[width=0.23\linewidth]{figures/supplementary/mipnerf/bicycle/ours/DSCF0680.JPG} &
    \includegraphics[width=0.23\linewidth]{figures/supplementary/mipnerf/bicycle/gt/DSCF0680.JPG} \\
    \midrule
    \viewlabel{View 1} &
    \includegraphics[width=0.23\linewidth]{figures/supplementary/mipnerf/kitchen/render/_DSC8679.JPG} &
    \includegraphics[width=0.23\linewidth]{figures/supplementary/mipnerf/kitchen/difix3d/_DSC8679.JPG} &
    \includegraphics[width=0.23\linewidth]{figures/supplementary/mipnerf/kitchen/ours/_DSC8679.JPG} &
    \includegraphics[width=0.23\linewidth]{figures/supplementary/mipnerf/kitchen/gt/_DSC8679.JPG} \\
    
    \viewlabel{View 2} &
    \includegraphics[width=0.23\linewidth]{figures/supplementary/mipnerf/kitchen/render/_DSC8816.JPG} &
    \includegraphics[width=0.23\linewidth]{figures/supplementary/mipnerf/kitchen/difix3d/_DSC8816.JPG} &
    \includegraphics[width=0.23\linewidth]{figures/supplementary/mipnerf/kitchen/ours/_DSC8816.JPG} &
    \includegraphics[width=0.23\linewidth]{figures/supplementary/mipnerf/kitchen/gt/_DSC8816.JPG} \\

    \end{tabular}
    \vspace{-5pt}
\captionof{figure}{\textbf{Visual comparison of multi-view refinement with feedforward renderings}. SyncFix provides coherent multi-view refinement that reduces the artifacts and maintains consistency.}
\label{fig:mipnerf_exp}
\end{center}
\vspace{-10pt}
\end{figure}

\section{Limitation and Failure Cases}

\begin{figure}[t!]
\begin{center}
    \setlength{\tabcolsep}{1pt}      %
    \renewcommand{\arraystretch}{1.0} %
    \small
    
    \settoheight{\imgH}{\includegraphics[width=0.28\linewidth]{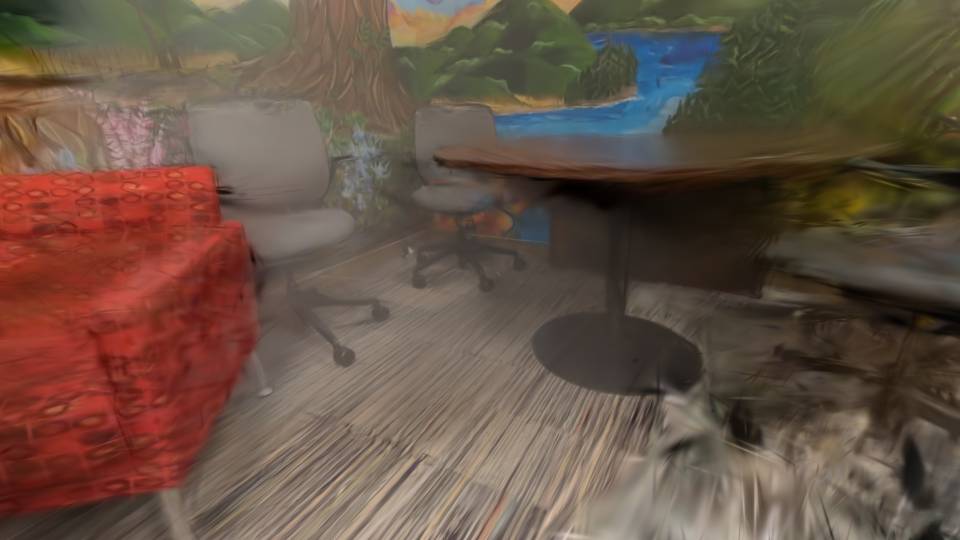}}
    
    \newcommand{\viewlabel}[1]{%
      \parbox[b][\imgH][c]{2.0em}{\centering\rotatebox{90}{\text{#1}}}%
    }
    
    \begin{tabular}{c c c c}
        & \text{3DGS} & \text{SnycFix} & \text{GT} \\

    \viewlabel{View 1} &
    \includegraphics[width=0.30\linewidth]{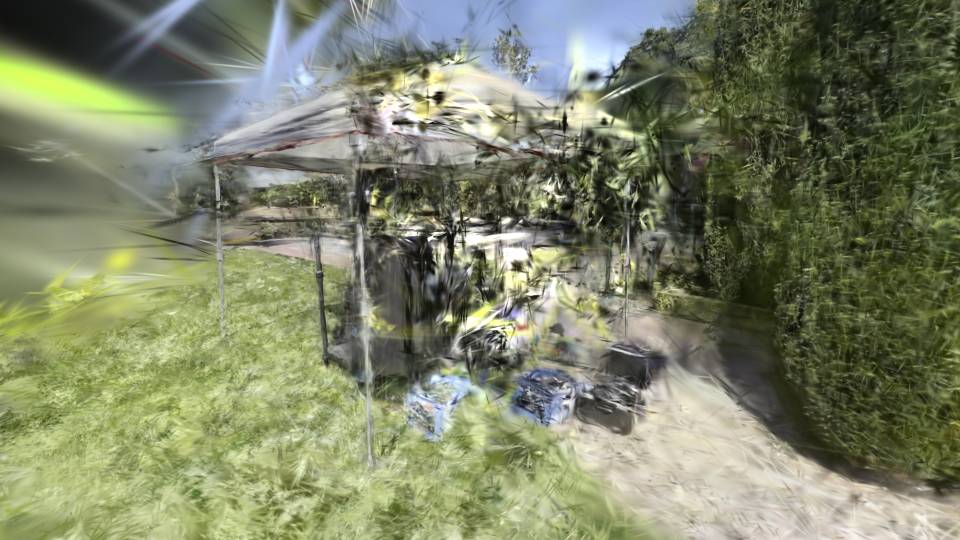} &
    \includegraphics[width=0.30\linewidth]{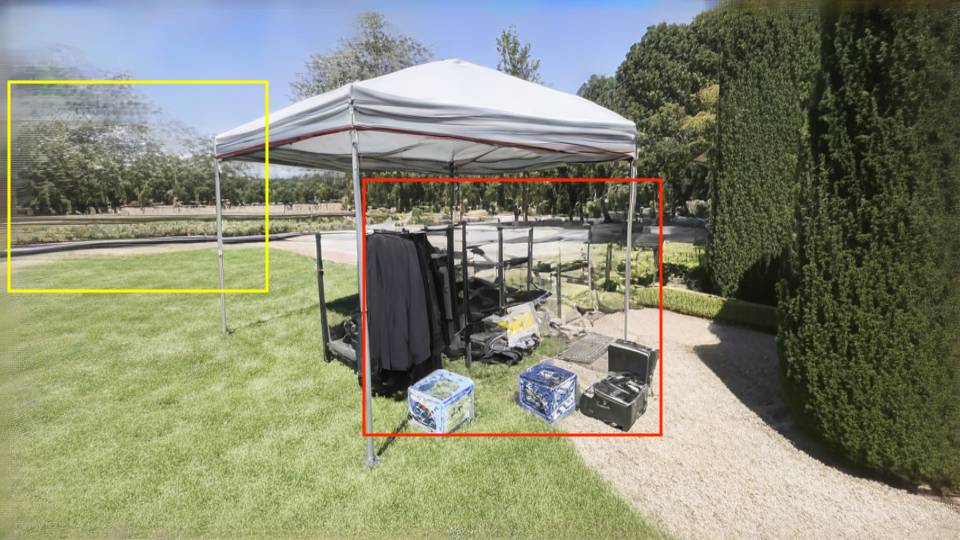} &
    \includegraphics[width=0.30\linewidth]{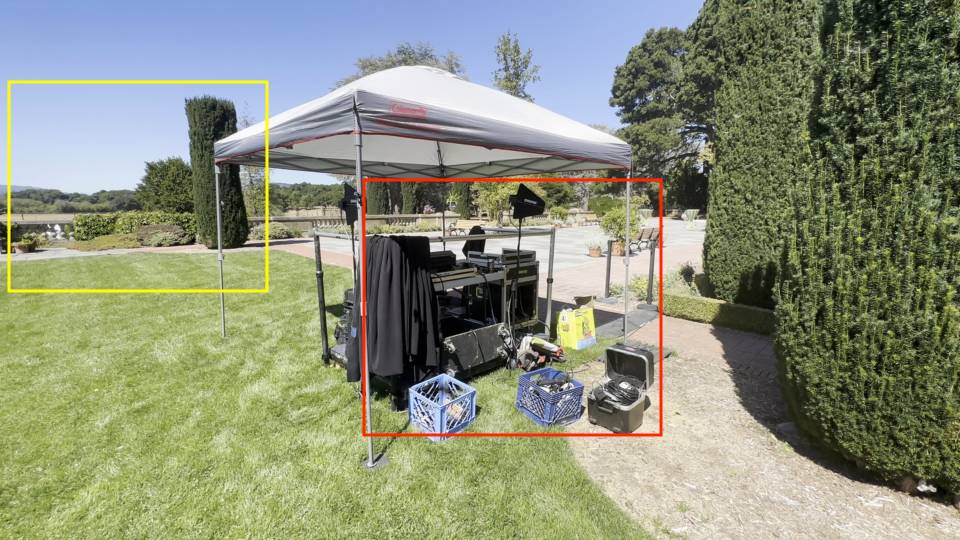} \\
    
    \viewlabel{View 2} &
    \includegraphics[width=0.30\linewidth]{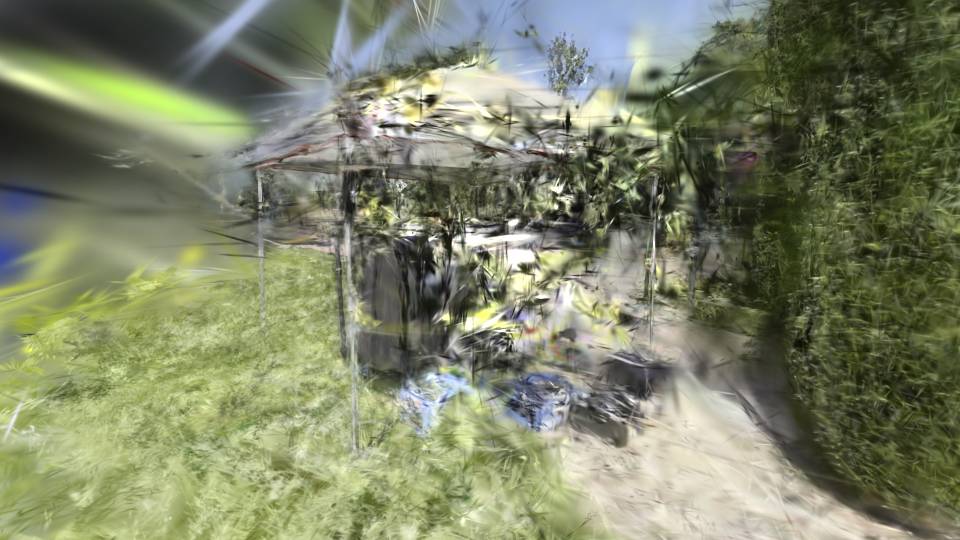} &
    \includegraphics[width=0.30\linewidth]{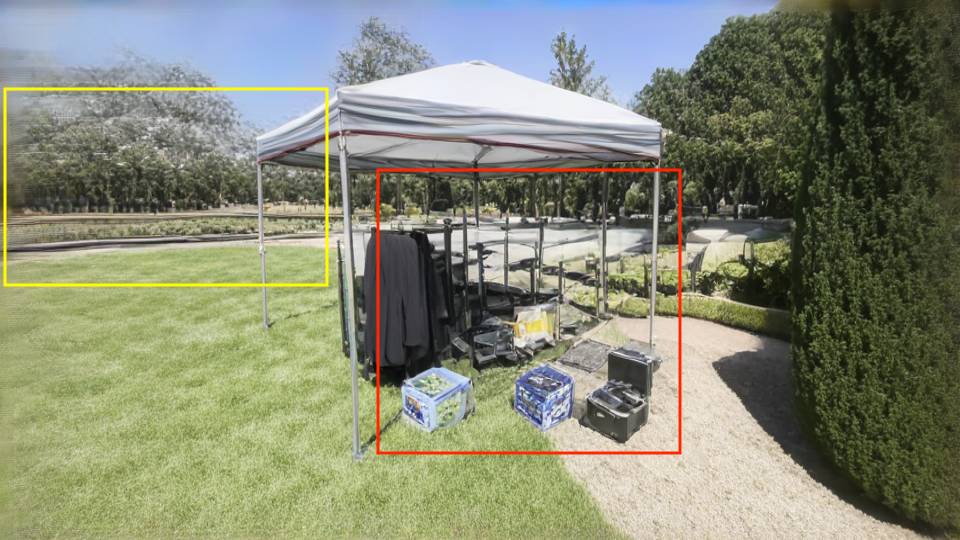} &
    \includegraphics[width=0.30\linewidth]{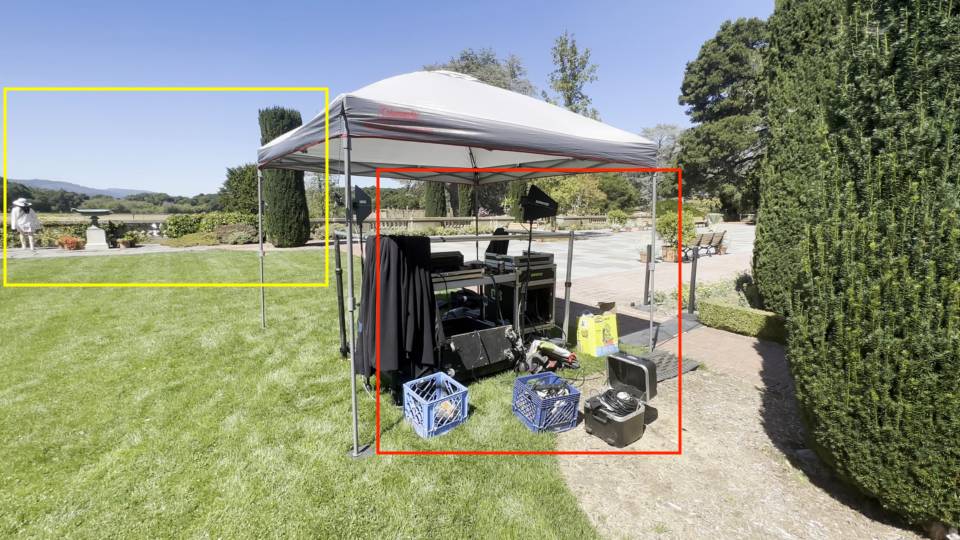} \\
    \midrule
    \viewlabel{View 1} &
    \includegraphics[width=0.30\linewidth]{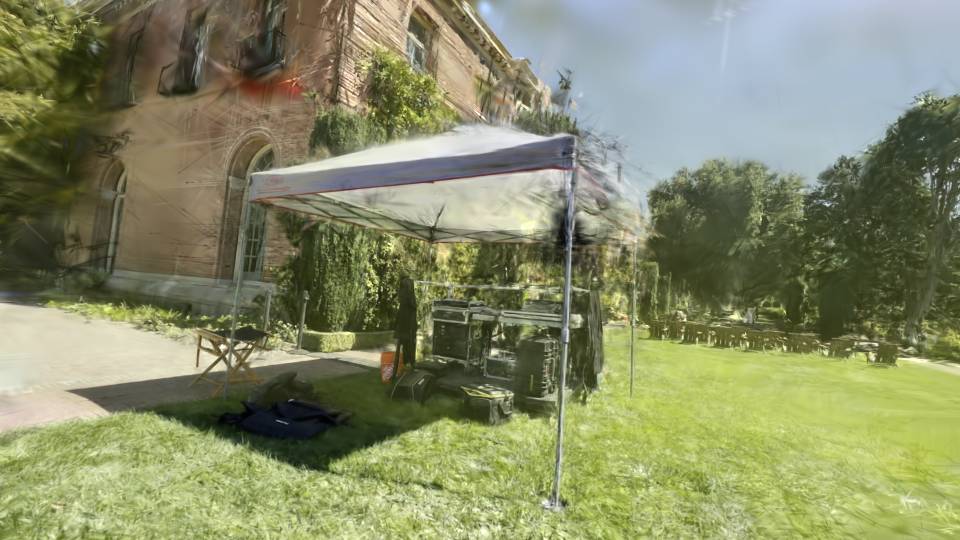} &
    \includegraphics[width=0.30\linewidth]{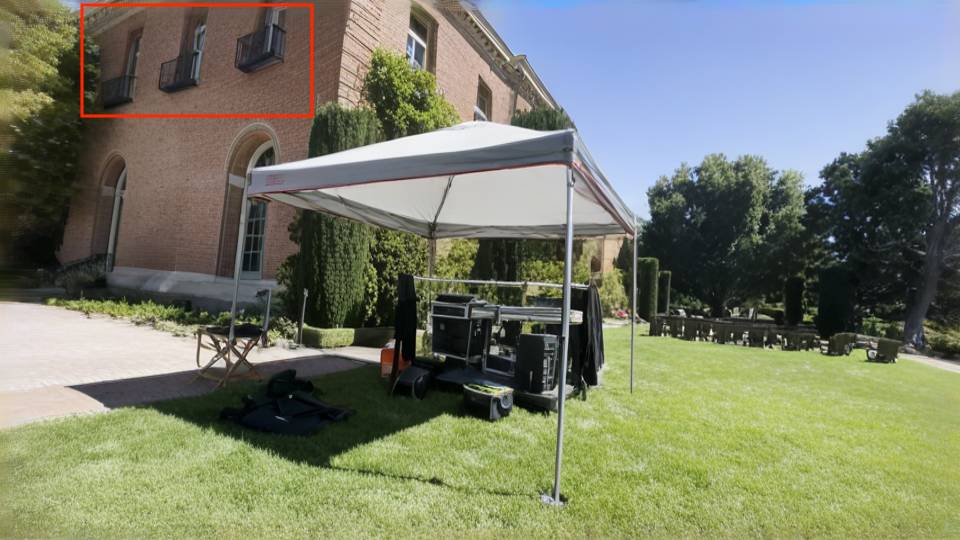} &
    \includegraphics[width=0.30\linewidth]{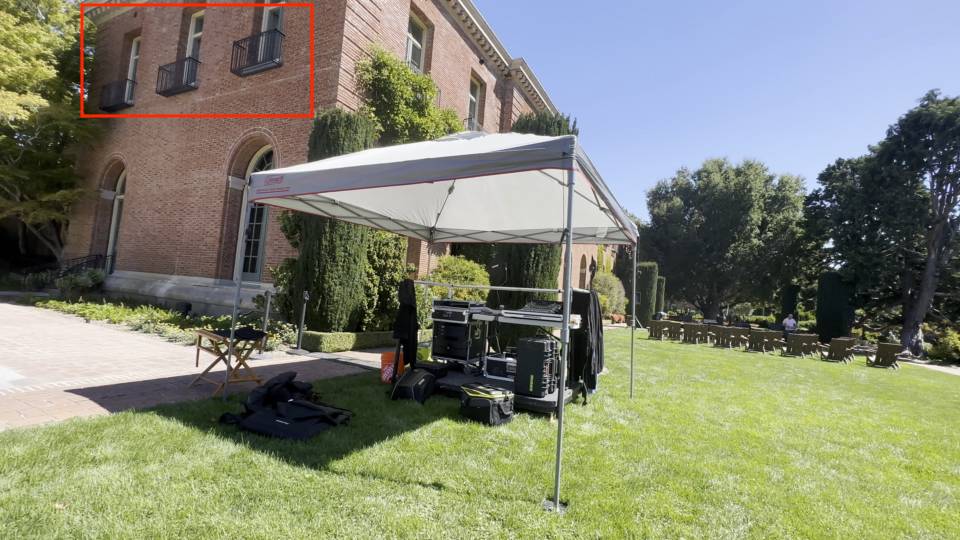} \\
    
    \viewlabel{View 2} &
    \includegraphics[width=0.30\linewidth]{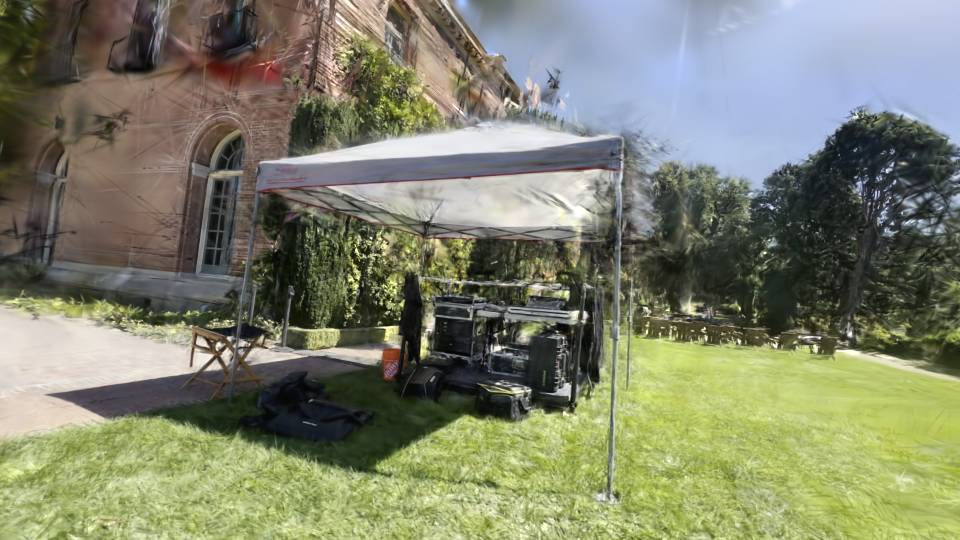} &
    \includegraphics[width=0.30\linewidth]{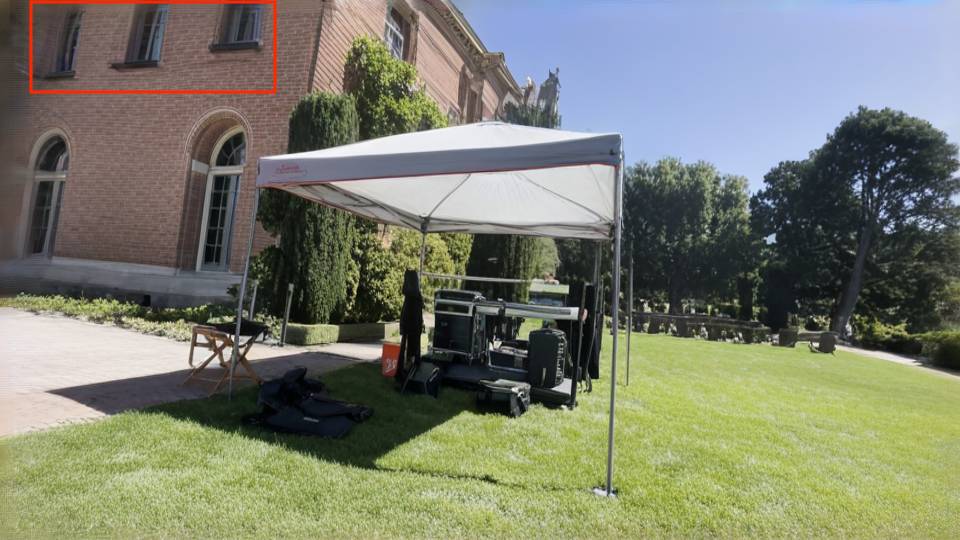} &
    \includegraphics[width=0.30\linewidth]{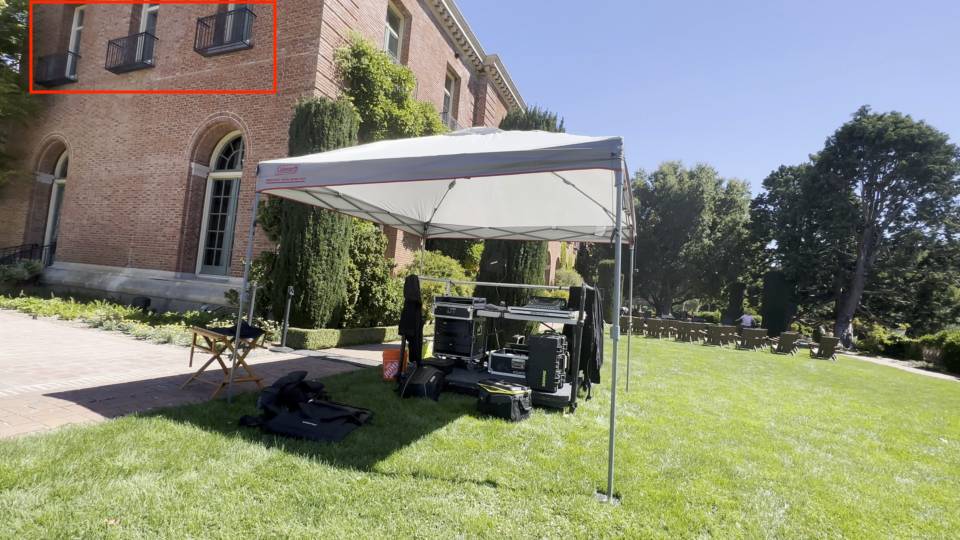} \\
    \end{tabular}
    \vspace{-5pt}
\captionof{figure}{\textbf{Limitation}. SyncFix may synthesize content that mismatches the ground-truth information. The resulting pixel-wise metrics are worse for this scene despite being multi-view consistent. Multi-view consistency can break down when views are refined in isolation. As seen in our SyncFix results, the windowsills in the bottom views become inconsistent because the independent network passes fail to synchronize geometric details across different perspectives. Note that the level of corruption for the top and bottom panels is different in the 3DGS views.}
\label{fig:limitation}
\end{center}
\vspace{-10pt}
\end{figure}

Regions in sparse 3DGS reconstructions may be severely corrupted or unobserved from the available training viewpoints. In such cases, SyncFix tends to synthesize plausible structures and textures (Fig.~\ref{fig:limitation}); however, these generated details may not match the ground-truth scene content. To better characterize cross-view behavior under such ambiguity, we introduce CVSC to quantify multi-view consistency beyond pixel-level fidelity. Developing more comprehensive evaluation protocols for this setting remains an important direction for future work. Moreover, while SyncFix encourages content that is consistent across views, it does not explicitly guarantee semantically meaningful or physically valid structures. Incorporating additional training constraints (e.g., semantic or geometric priors) is another promising avenue for future improvement. Another limitation is that SyncFix aims to maintain multi-view consistency within each single forward pass over the provided set of input images. Refined views from different passes may result in multi-view inconsistency. As illustrated in Fig.~\ref{fig:limitation} (e.g., the windowsill region), refinements produced across separate passes can diverge in appearance or geometry. To mitigate this issue, incorporating more views per pass or enforcing explicit cross-pass consistency constraints are promising directions.

\section{Comparison with Fixer}
\label{sec:fixer}

Fixer \cite{nvtlabs_fixer_github} is a follow-up work of Difix3D+ \cite{wu2025difix3d+} based on Cosmos-Predict \cite{ali2025world} Diffusion Transformer architecture. Although Fixer improves the renderings with lower DreamSim score and FID, it obtains suboptimal results compared with Difix3D+ and SyncFix. We attribute this gap primarily to the absence of reference images, which limits Fixer’s ability to disambiguate missing content and artifacts for high-fidelity refinements. More visual comparison can be found in Sec. \ref{sec:quali}

\begin{table}[htbp]
\centering
\setlength{\tabcolsep}{3pt}
\caption{
Quantitative comparison on the DL3DV and NeRFBusters test sets for refining sparse-view 3D Gaussian Splatting reconstructions.
SyncFix improves reconstruction quality and cross-view semantic consistency over prior generative refinement methods.
$\uparrow$ indicates higher-is-better and $\downarrow$ indicates lower-is-better.
\textbf{Bold} denotes best results and \underline{underline} denotes second best.
}
\resizebox{\textwidth}{!}{
\begin{tabular}{lccccc|ccccc}
\toprule
& \multicolumn{5}{c}{DL3DV} & \multicolumn{5}{c}{NeRFBusters} \\
\cmidrule(lr){2-6} \cmidrule(lr){7-11}
Method 
& PSNR$\uparrow$ & LPIPS$\downarrow$ & DSIM$\downarrow$ & FID$\downarrow$ & CVSC$\uparrow$
& PSNR$\uparrow$ & LPIPS$\downarrow$ & DSIM$\downarrow$ & FID$\downarrow$ & CVSC$\uparrow$ \\
\midrule
3DGS \cite{kerbl20233d}
& 15.94 & 0.454 & 0.297 & 80.8 & \underline{0.875}
& 11.85 & 0.575 & 0.336 & 139.4 & \textbf{0.951} \\
\midrule
Difix3D+ \cite{wu2025difix3d+}
& \underline{16.05} & \underline{0.368} & \underline{0.171} & \underline{26.2} & 0.819
& 11.94 & \underline{0.528} & \underline{0.202} & \underline{97.4} & 0.925 \\
Fixer \cite{nvtlabs_fixer_github}
& 15.73 & 0.466 & 0.257 & 45.45 & 0.822
& \underline{12.32} & 0.588 & 0.296 & 126.7 & 0.817 \\
\rowcolor{lavenderrow}
SyncFix (ours)
& \textbf{16.94} & \textbf{0.305} & \textbf{0.099} & \textbf{17.5} & \textbf{0.880}
& \textbf{13.83} & \textbf{0.501} & \textbf{0.163} & \textbf{79.5} & \underline{0.940} \\
\bottomrule
\end{tabular}
}
\label{tab:quantitative}
\vspace{+10pt}
\end{table}

\section{More Qualitative Comparison}
\label{sec:quali}

\newcommand{\lbmimg}[1]{\includegraphics[width=0.19\textwidth]{#1}}

\newcommand{\lbmlabel}[1]{\rotatebox{90}{\text{#1}}}

\newcommand{\lbmfull}[4]{%
figures/for_lbm_supplimentary/#1/K_12/run_002/it_03000/#2/#3/#4_compare_v8/#4_compare_v8_annotated.png
}

\newcommand{\lbmcrop}[5]{%
figures/for_lbm_supplimentary/#1/K_12/run_002/it_03000/#2/#3/#4_compare_v8/#4_compare_v8_#5_boxed.png
}

\newcommand{\lbmrow}[6]{%
\lbmlabel{#1} &
\lbmimg{#2} &
\lbmimg{#3} &
\lbmimg{#4} &
\lbmimg{#5} &
\lbmimg{#6} \\
}

\newcommand{\lbmpairfigure}[6]{

\begin{figure}[htbp]
\centering

\setlength{\tabcolsep}{1pt}
\renewcommand{\arraystretch}{1.0}
\small

\begin{tabular}{c c c c c c}

& \text{3DGS} & \text{Fixer} & \text{Difix3D+} & \text{Ours} & \text{GT} \\

\lbmrow{View 1}
{\lbmfull{#1}{images}{#2}{#3}}
{\lbmfull{#1}{fixer}{#2}{#3}}
{\lbmfull{#1}{difix_images_pretrained_ref}{#2}{#3}}
{\lbmfull{#1}{lbm_images_ref_new_final_finetune3}{#2}{#3}}
{\lbmfull{#1}{gt}{#2}{#3}}

\lbmrow{View 2}
{\lbmfull{#1}{images}{#2}{#4}}
{\lbmfull{#1}{fixer}{#2}{#4}}
{\lbmfull{#1}{difix_images_pretrained_ref}{#2}{#4}}
{\lbmfull{#1}{lbm_images_ref_new_final_finetune3}{#2}{#4}}
{\lbmfull{#1}{gt}{#2}{#4}}

\lbmrow{View 1}
{\lbmcrop{#1}{images}{#2}{#3}{crop1}}
{\lbmcrop{#1}{fixer}{#2}{#3}{crop1}}
{\lbmcrop{#1}{difix_images_pretrained_ref}{#2}{#3}{crop1}}
{\lbmcrop{#1}{lbm_images_ref_new_final_finetune3}{#2}{#3}{crop1}}
{\lbmcrop{#1}{gt}{#2}{#3}{crop1}}

\lbmrow{View 2}
{\lbmcrop{#1}{images}{#2}{#4}{crop1}}
{\lbmcrop{#1}{fixer}{#2}{#4}{crop1}}
{\lbmcrop{#1}{difix_images_pretrained_ref}{#2}{#4}{crop1}}
{\lbmcrop{#1}{lbm_images_ref_new_final_finetune3}{#2}{#4}{crop1}}
{\lbmcrop{#1}{gt}{#2}{#4}{crop1}}

\end{tabular}

\vspace{-5pt}

\caption{#5}

\label{#6}

\vspace{-10pt}

\end{figure}
}

\includegraphics{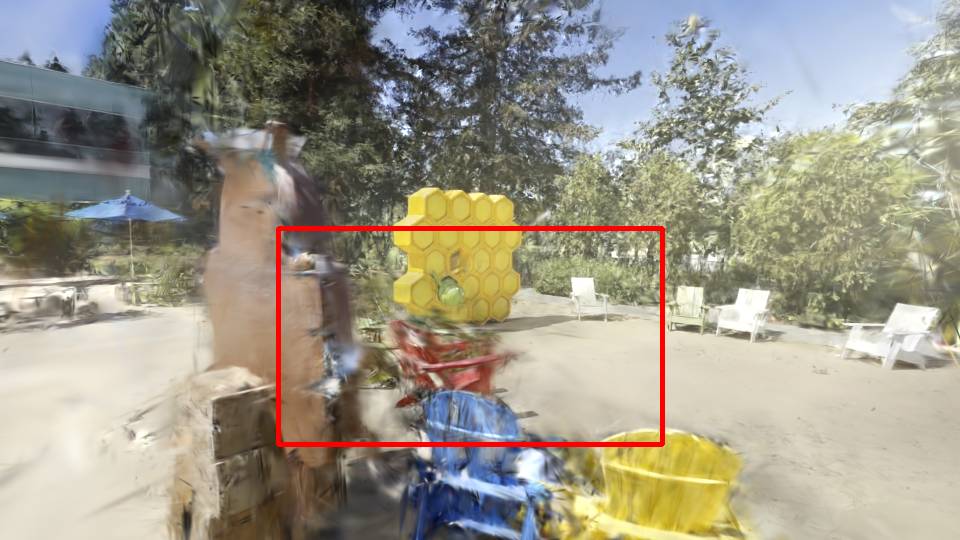}\includegraphics{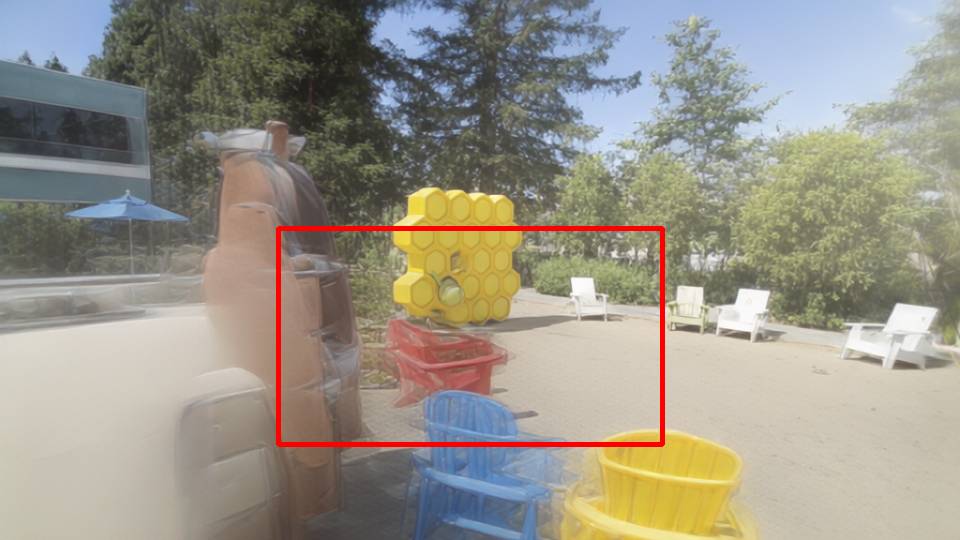}\includegraphics{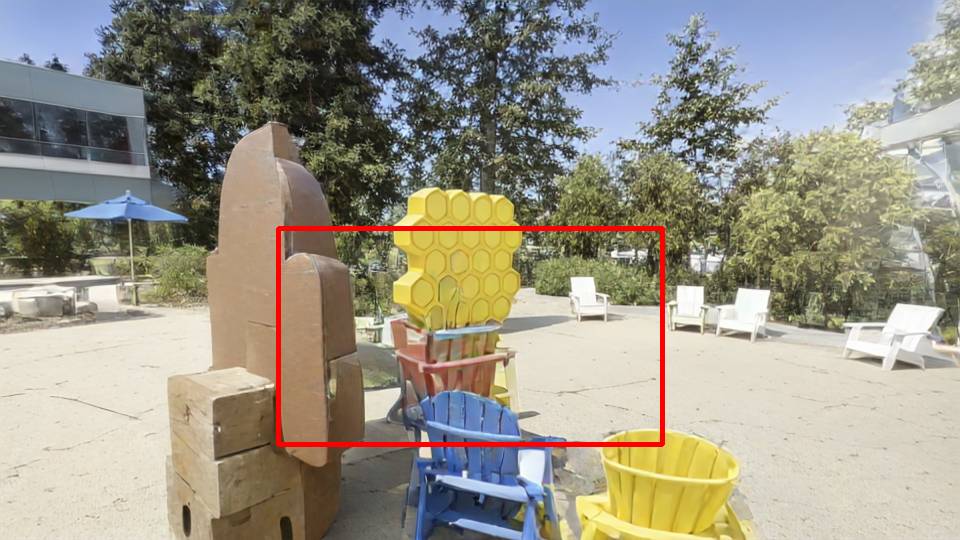}\includegraphics{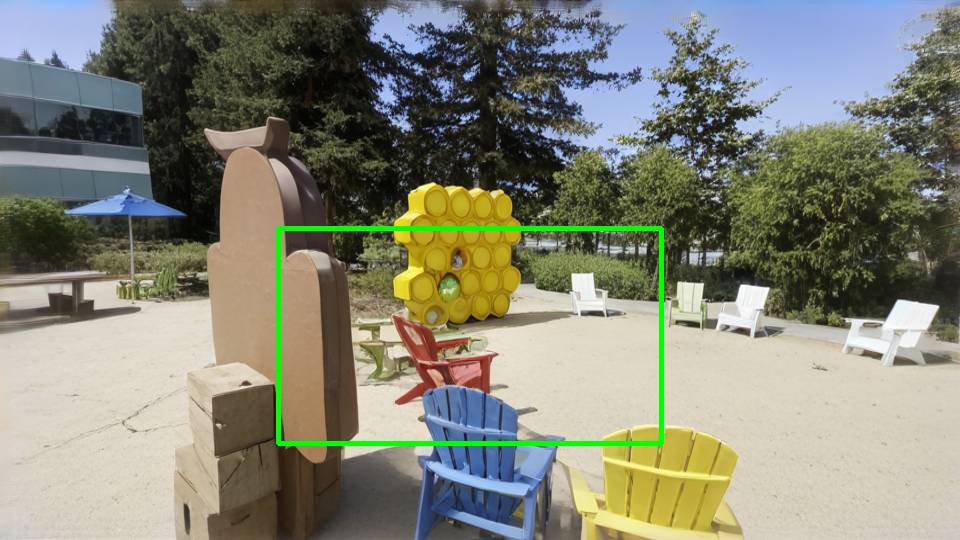}\includegraphics{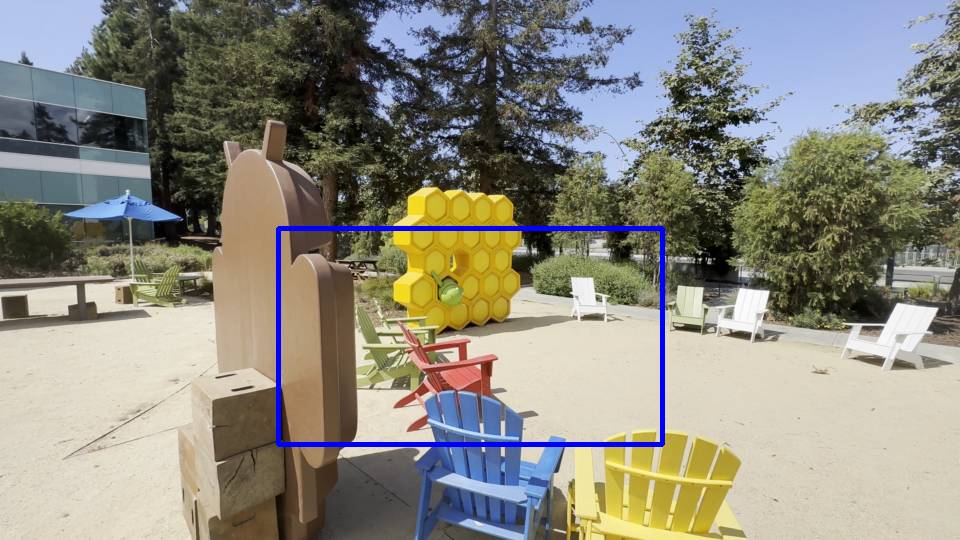}\includegraphics{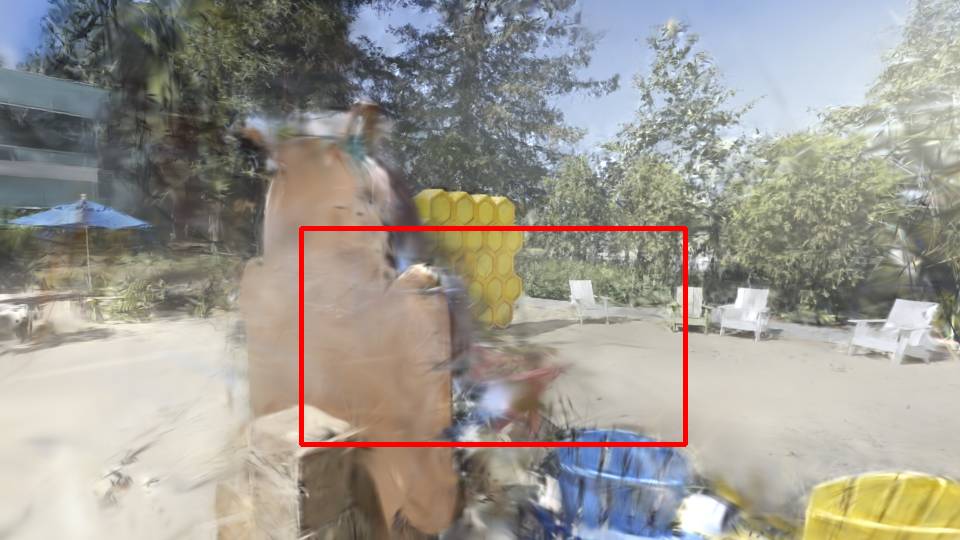}\includegraphics{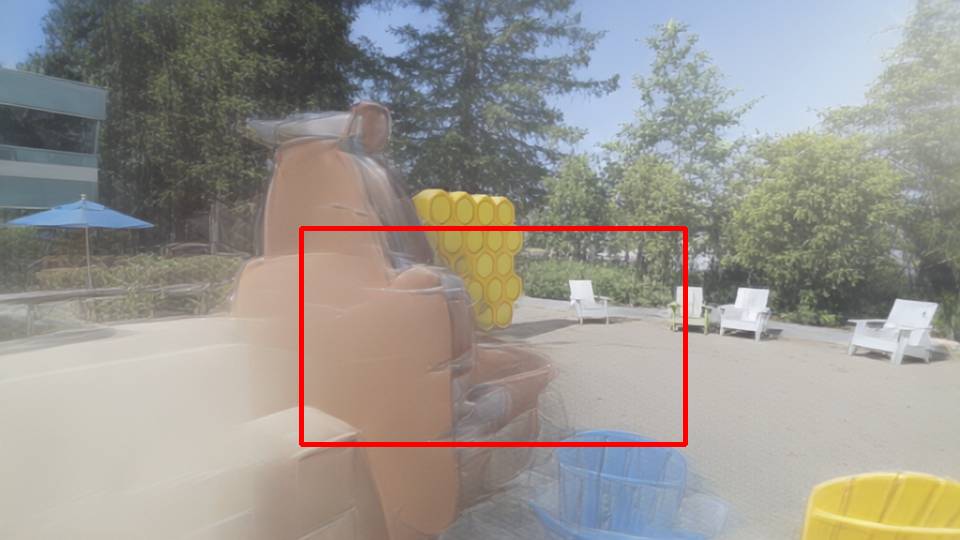}\includegraphics{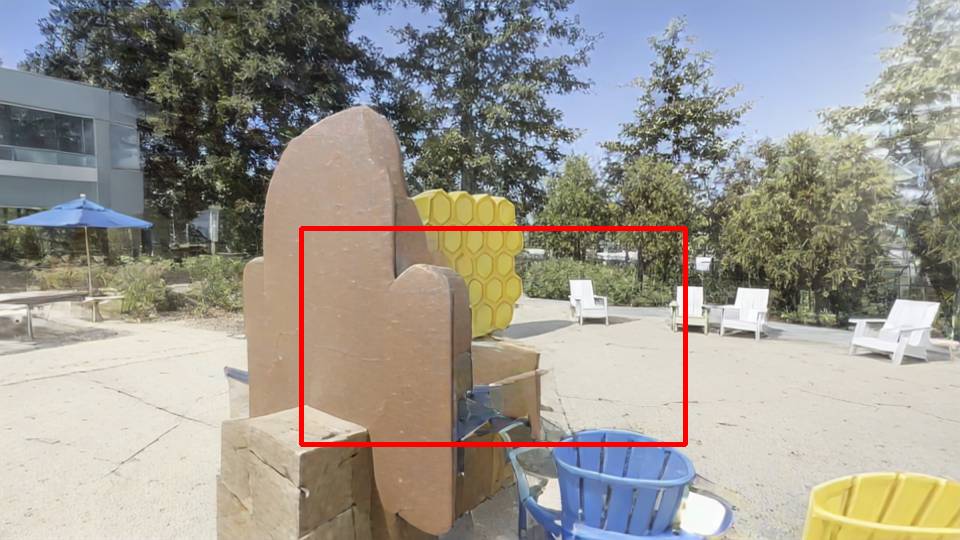}\includegraphics{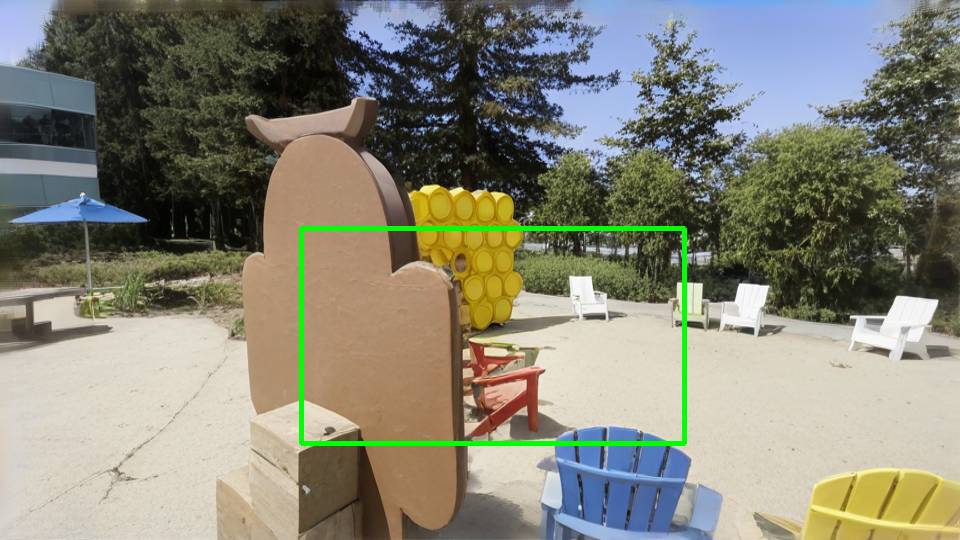}\includegraphics{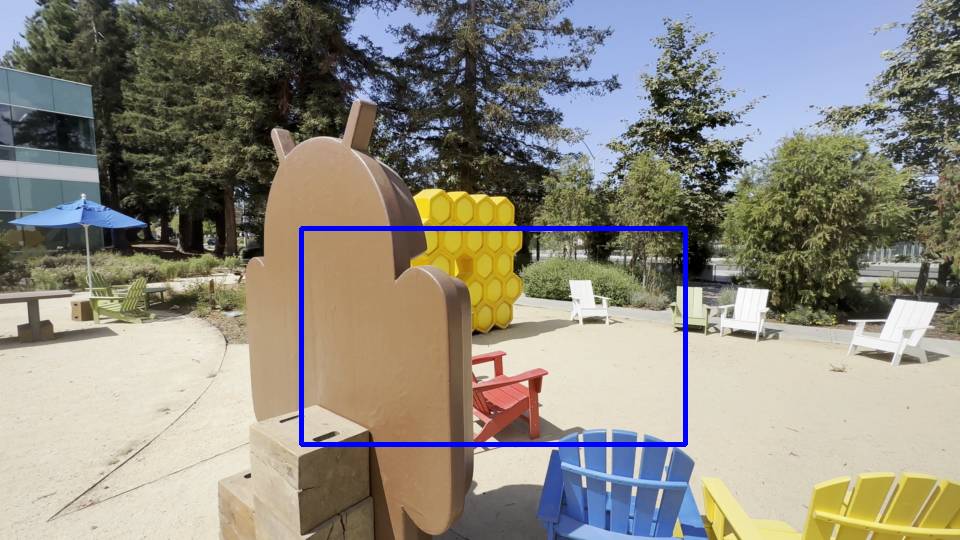}\includegraphics{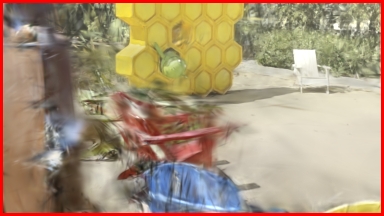}\includegraphics{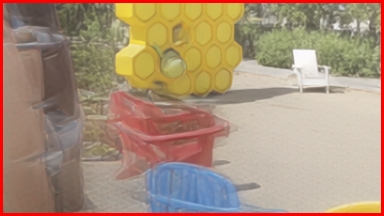}\includegraphics{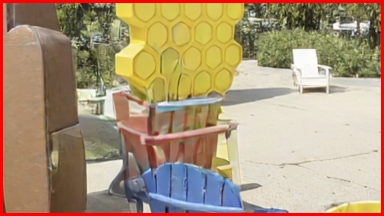}\includegraphics{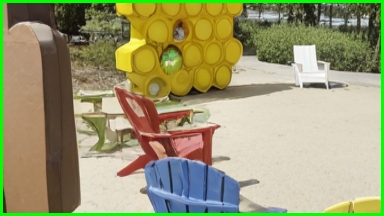}\includegraphics{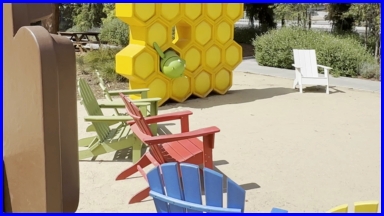}\includegraphics{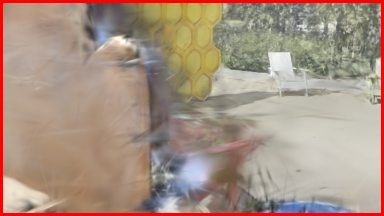}\includegraphics{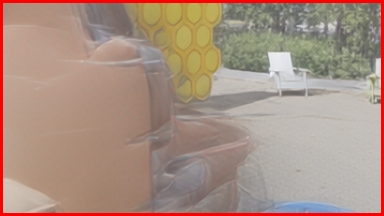}\includegraphics{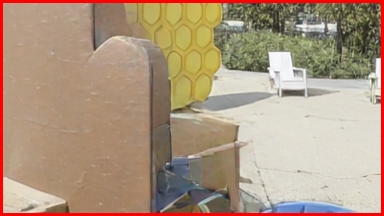}\includegraphics{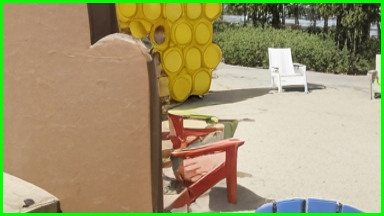}\includegraphics{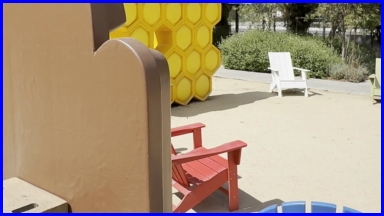}

\includegraphics{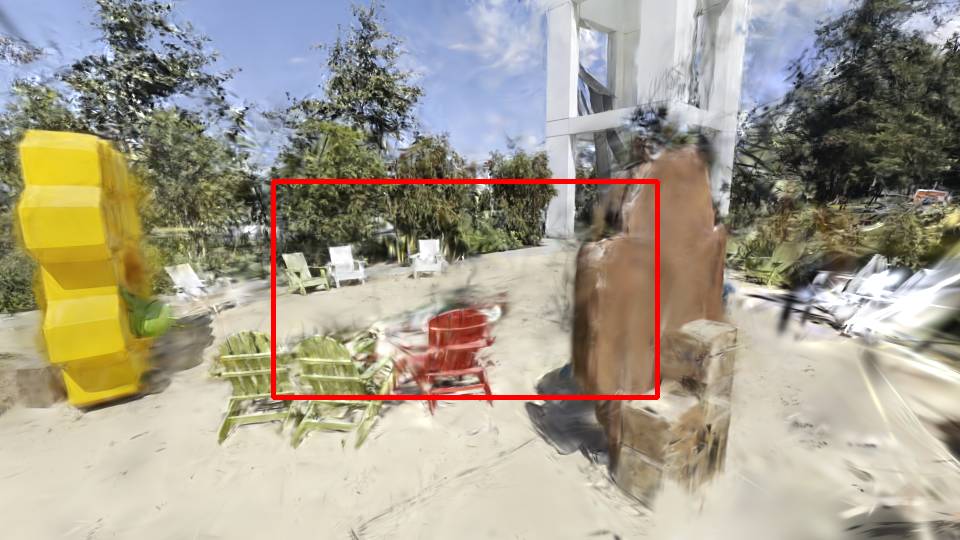}\includegraphics{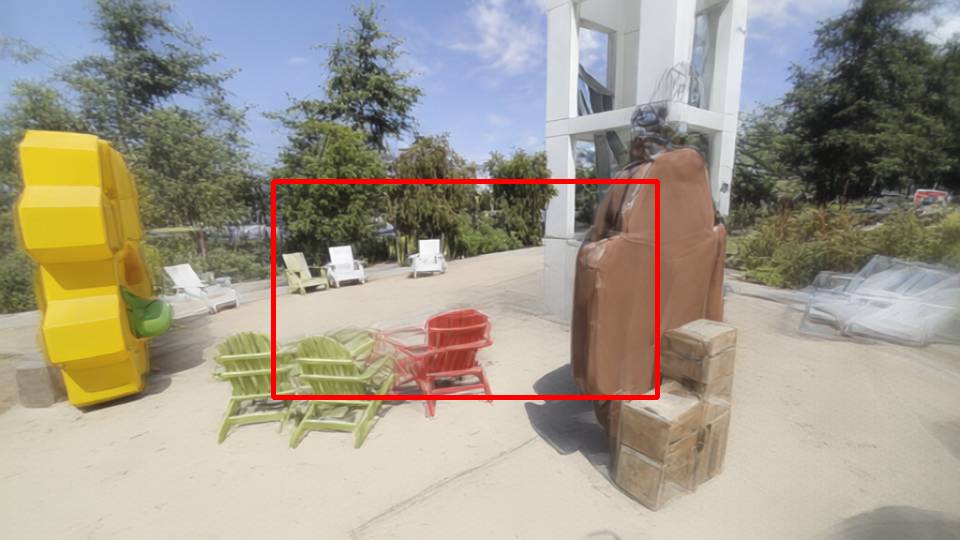}\includegraphics{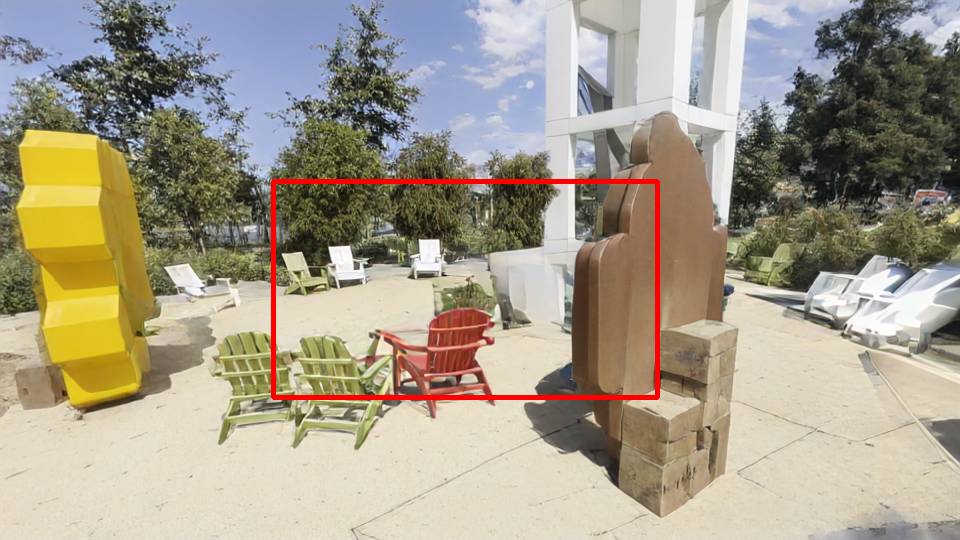}\includegraphics{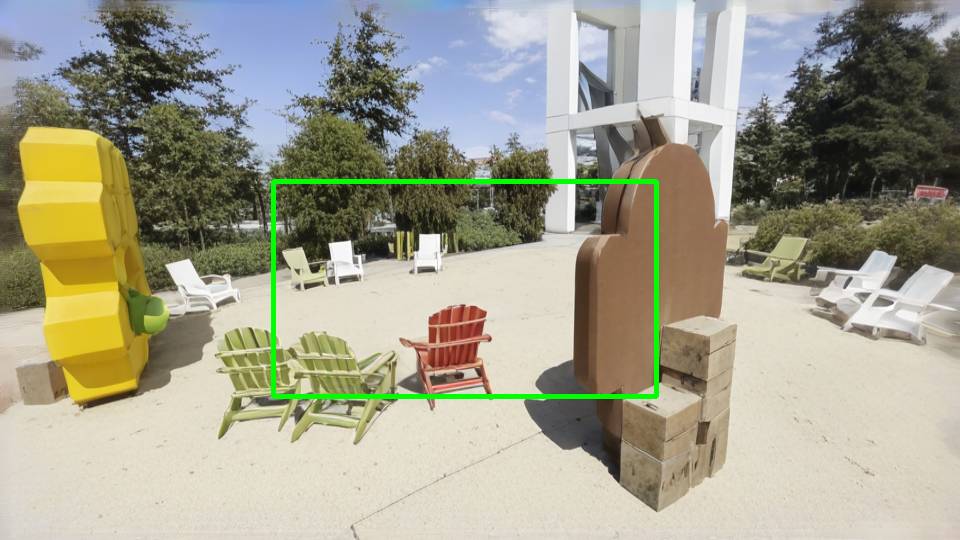}\includegraphics{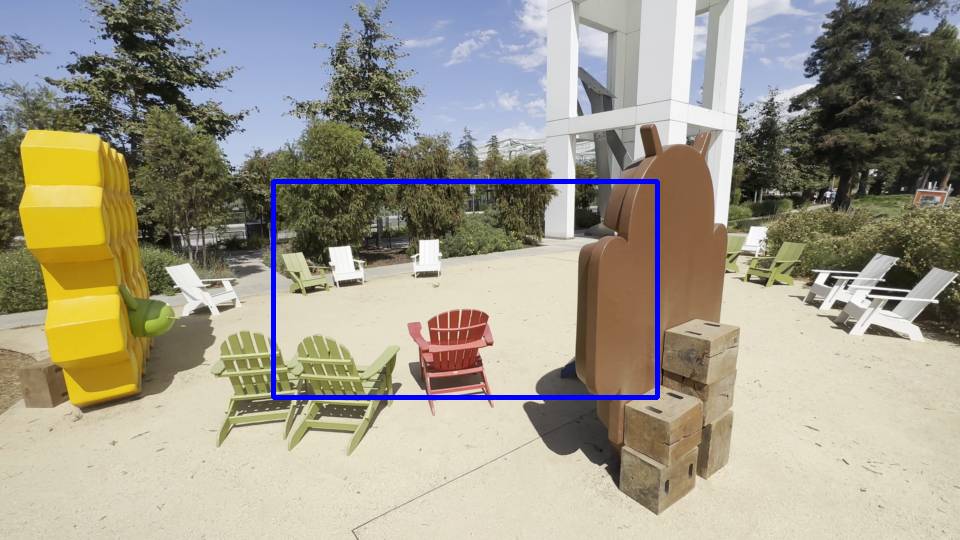}\includegraphics{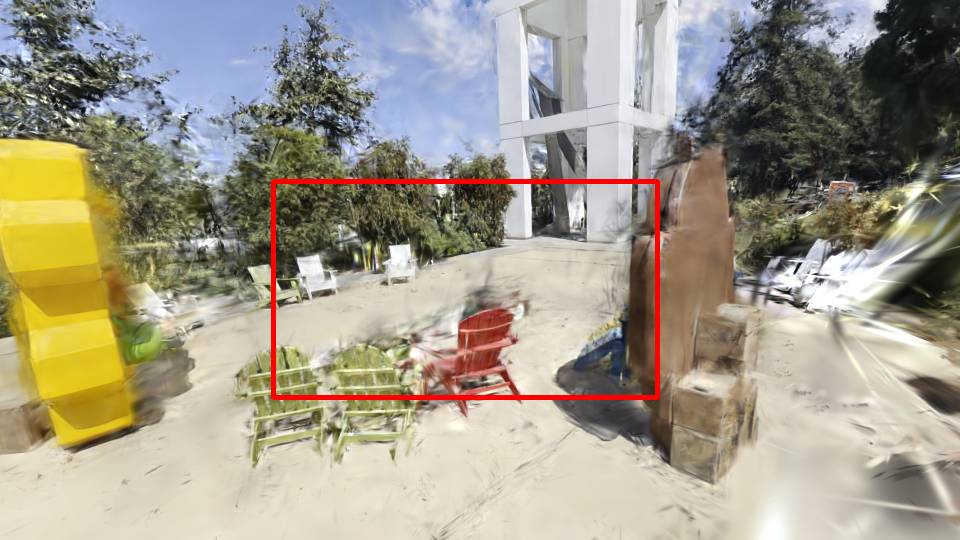}\includegraphics{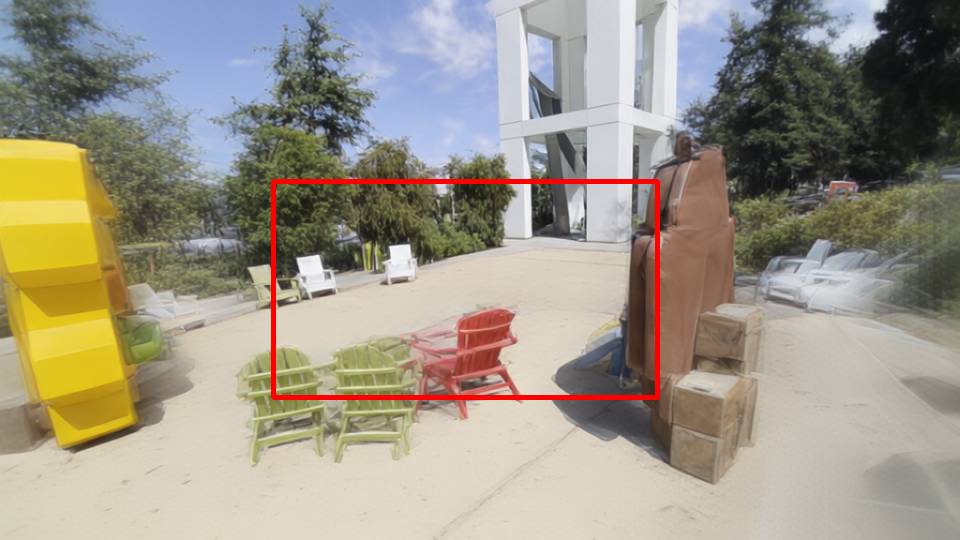}\includegraphics{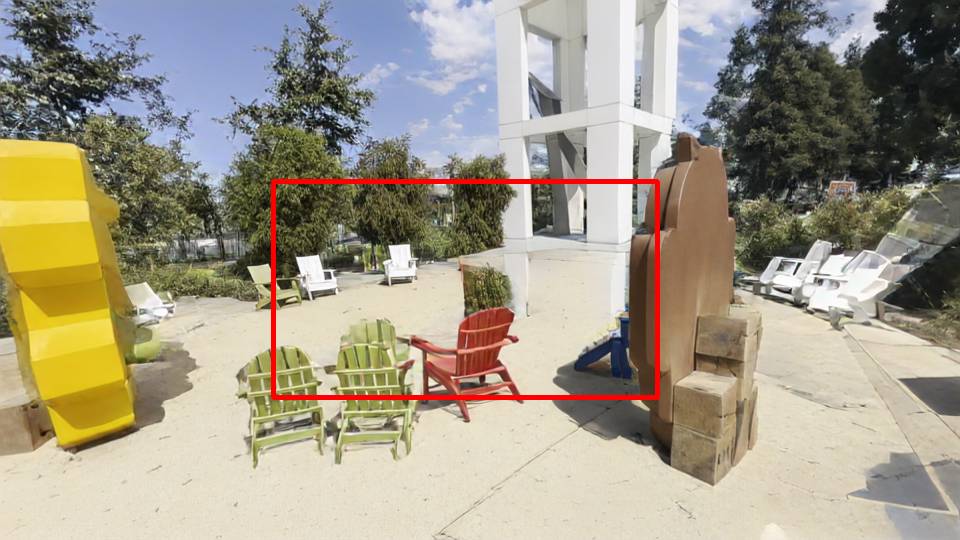}\includegraphics{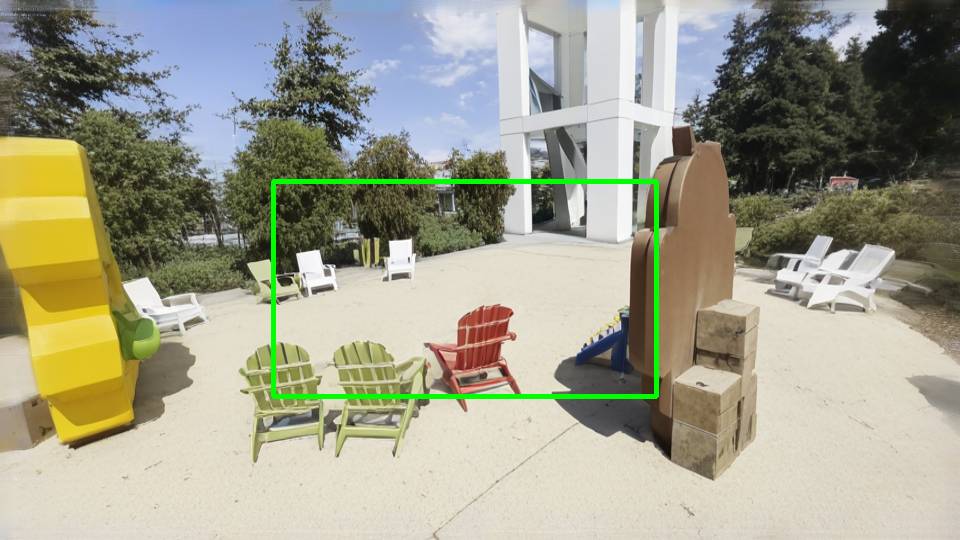}\includegraphics{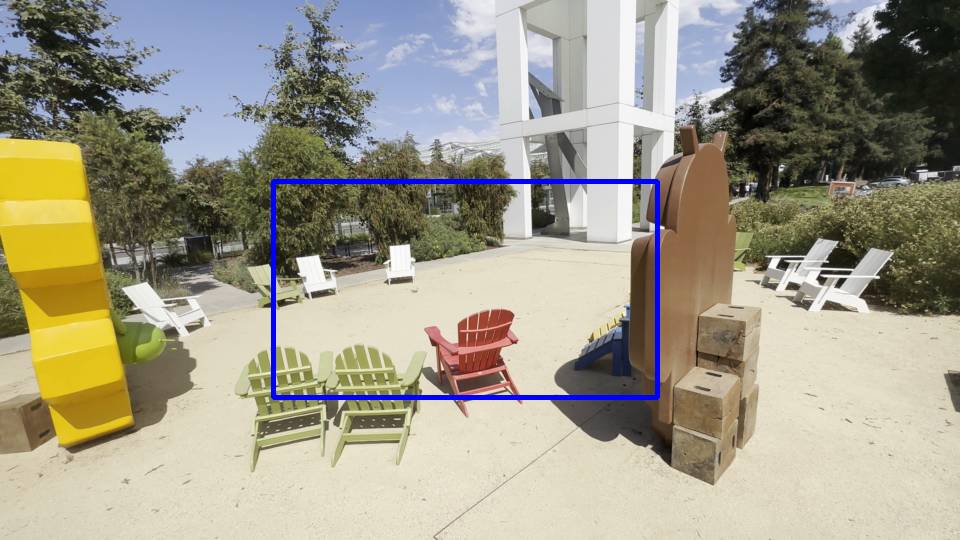}\includegraphics{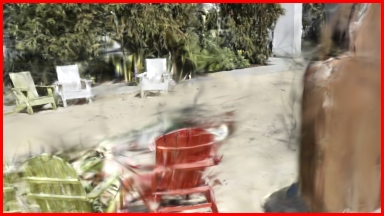}\includegraphics{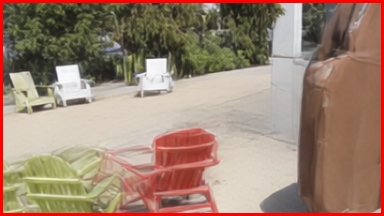}\includegraphics{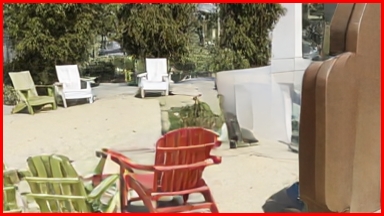}\includegraphics{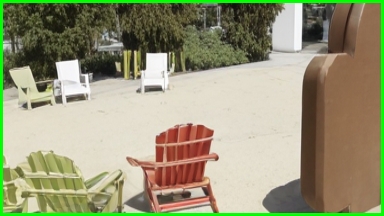}\includegraphics{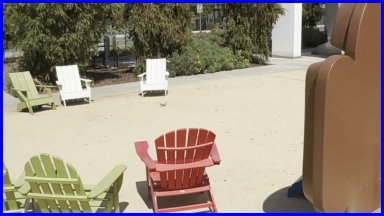}\includegraphics{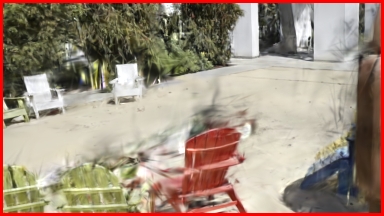}\includegraphics{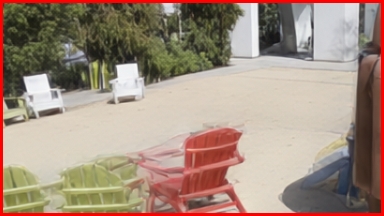}\includegraphics{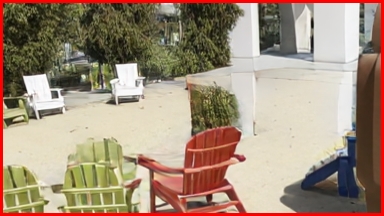}\includegraphics{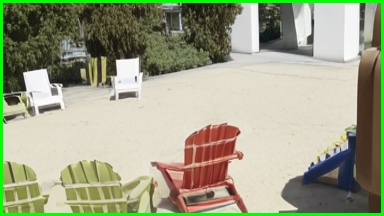}\includegraphics{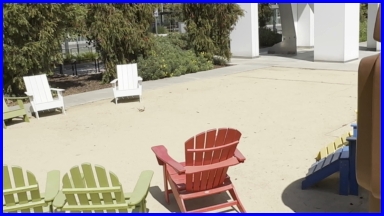}

\includegraphics{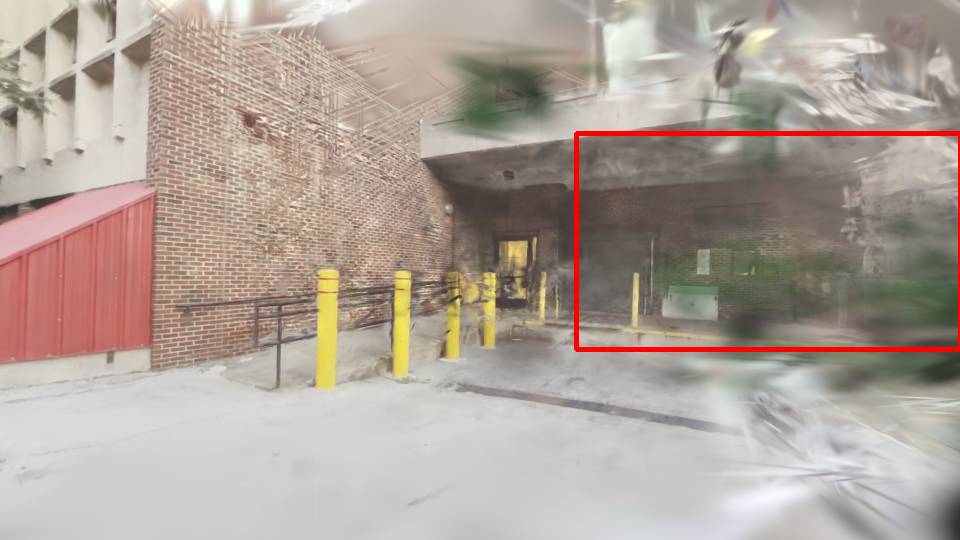}\includegraphics{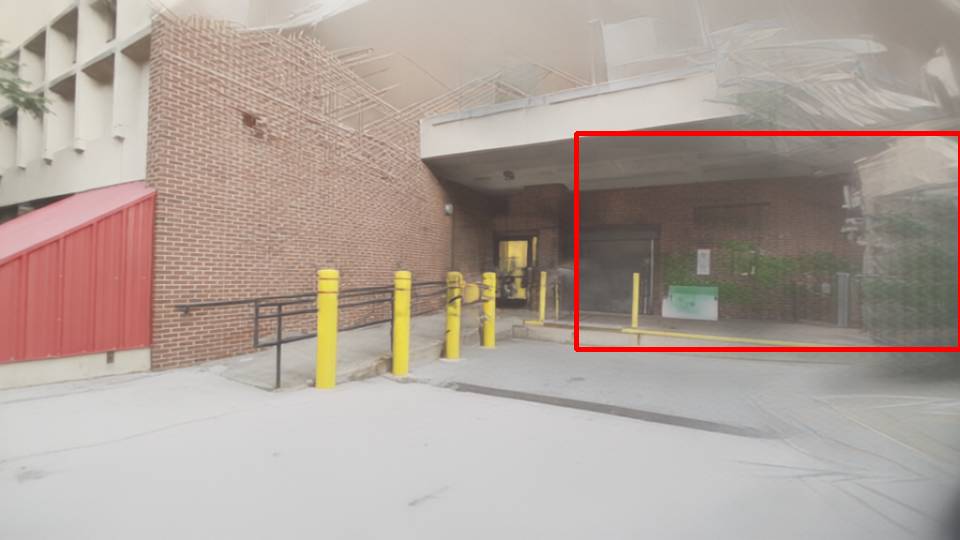}\includegraphics{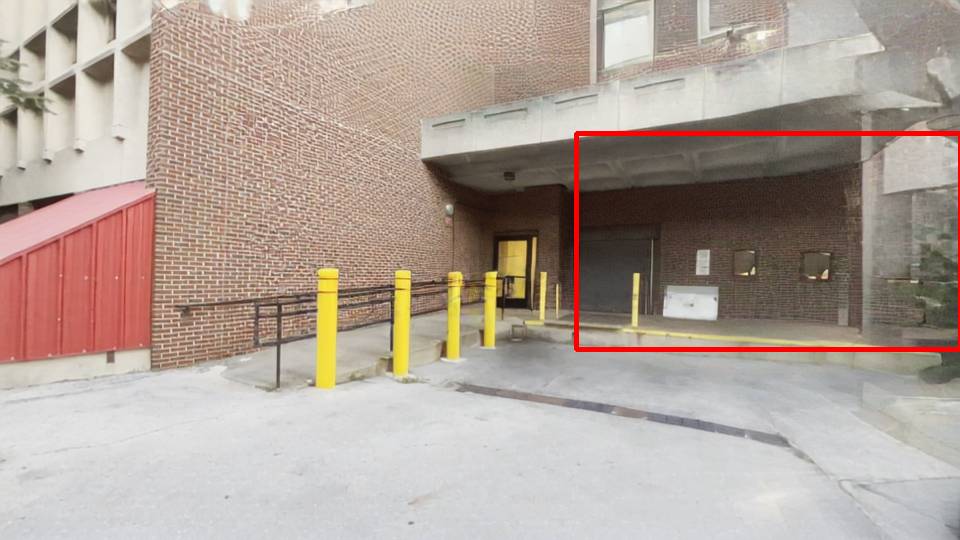}\includegraphics{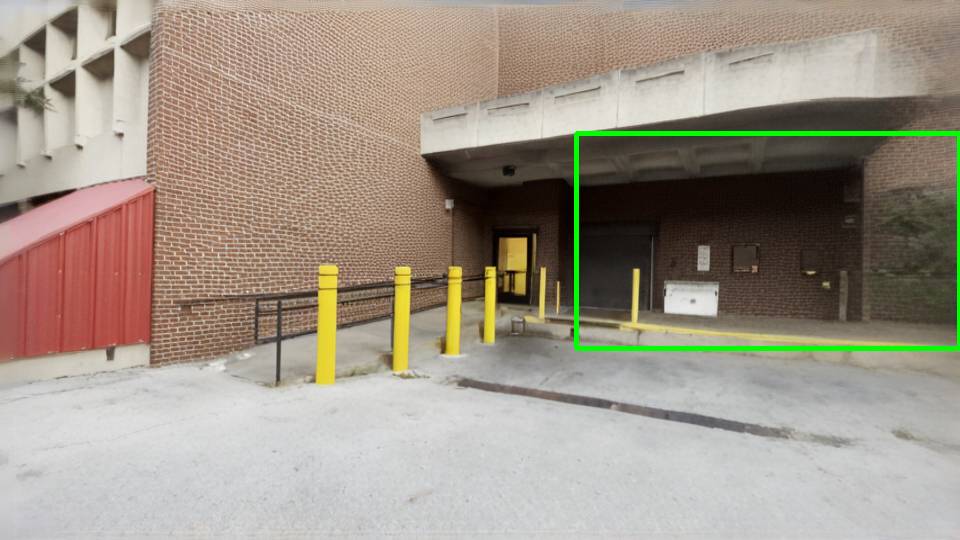}\includegraphics{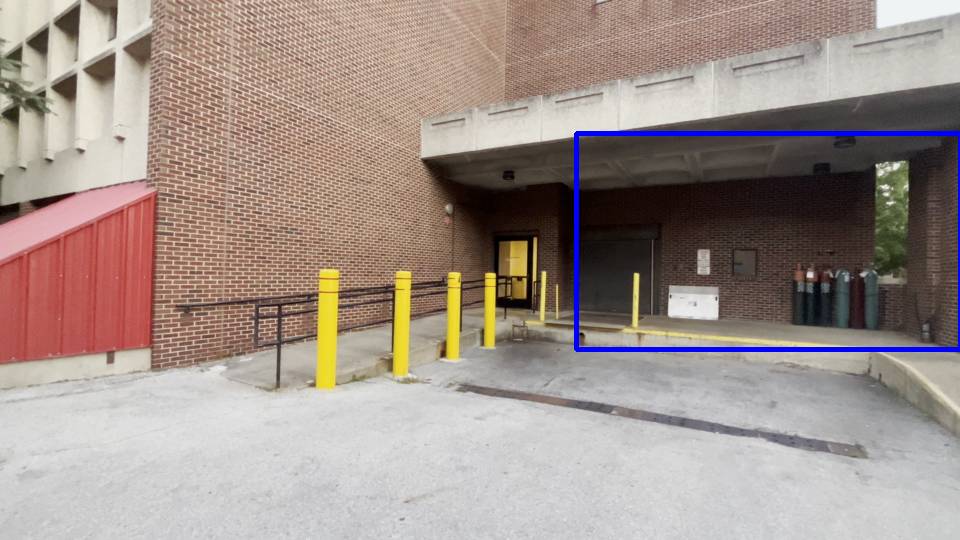}\includegraphics{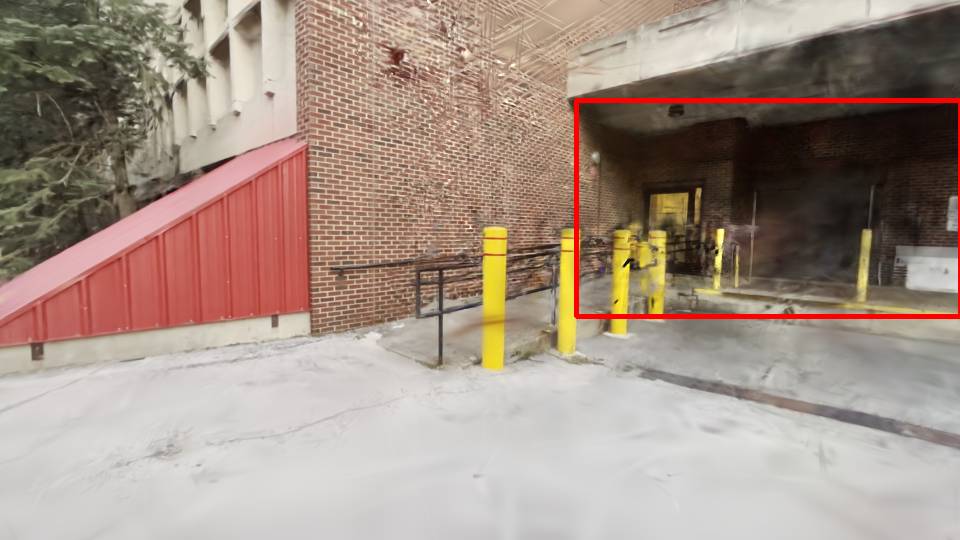}\includegraphics{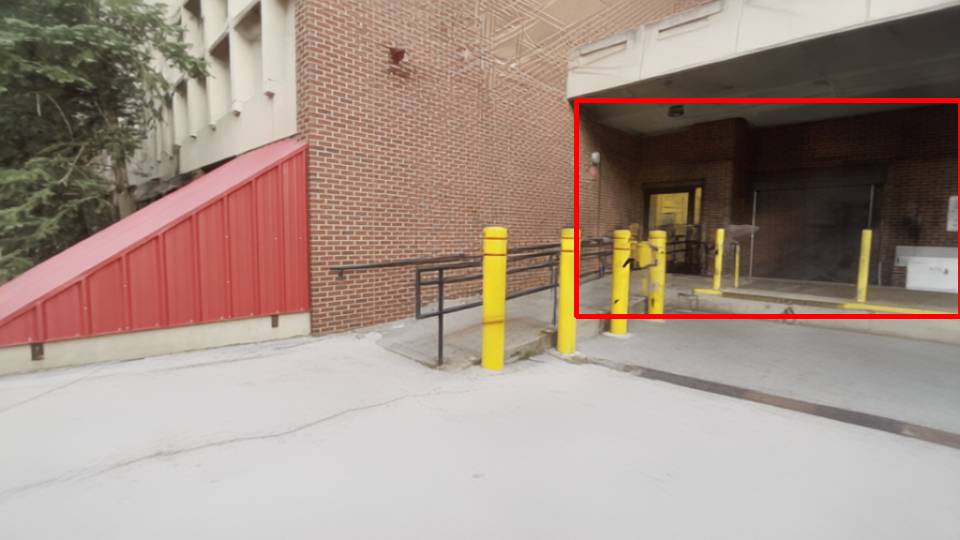}\includegraphics{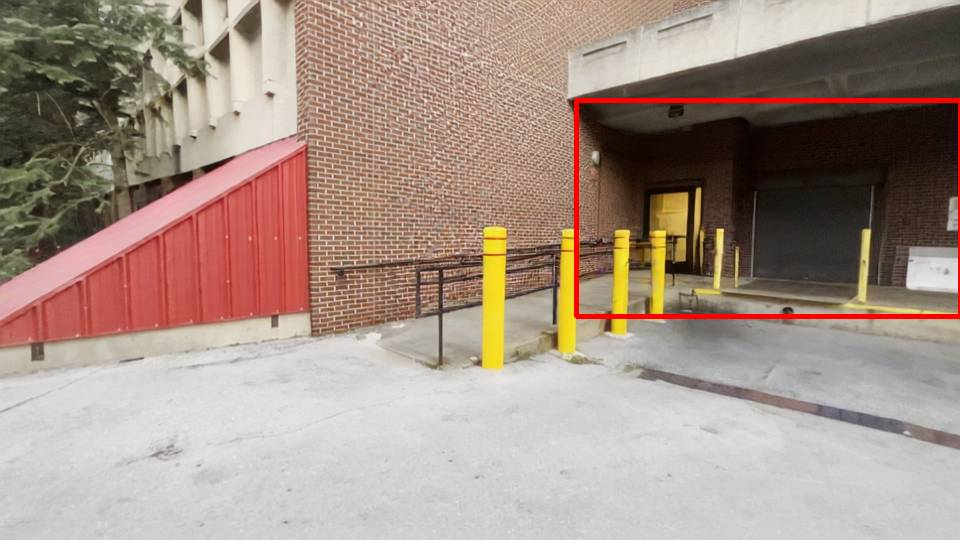}\includegraphics{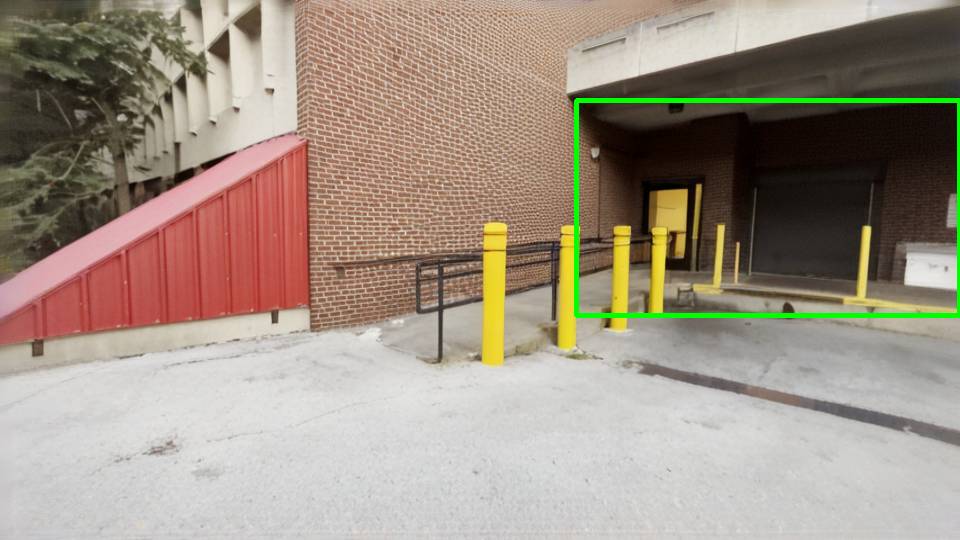}\includegraphics{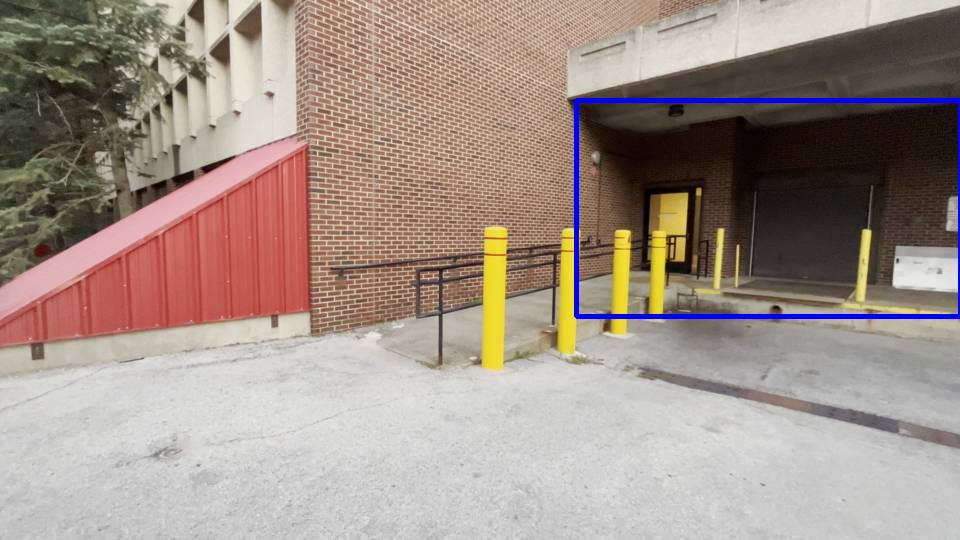}\includegraphics{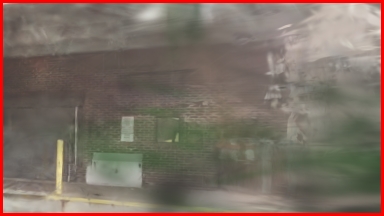}\includegraphics{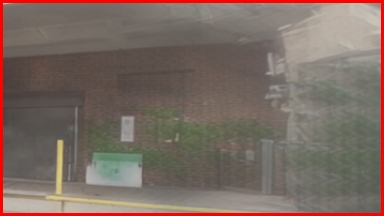}\includegraphics{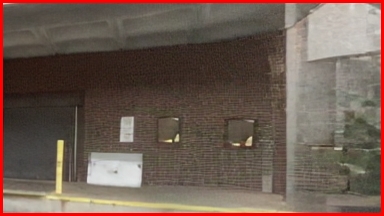}\includegraphics{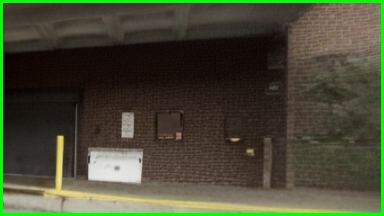}\includegraphics{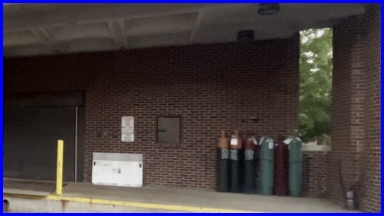}\includegraphics{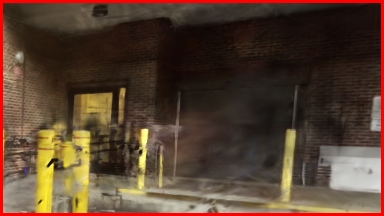}\includegraphics{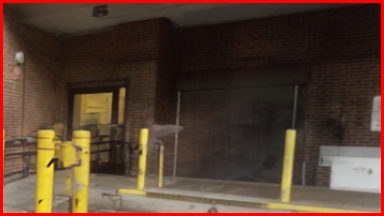}\includegraphics{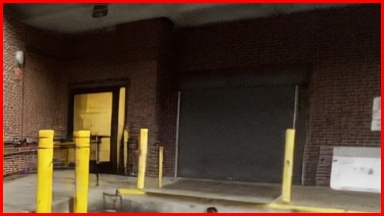}\includegraphics{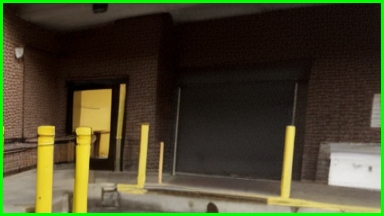}\includegraphics{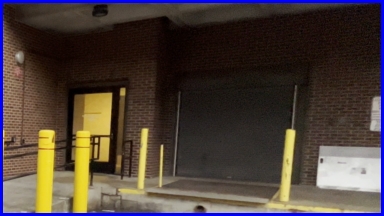}

\includegraphics{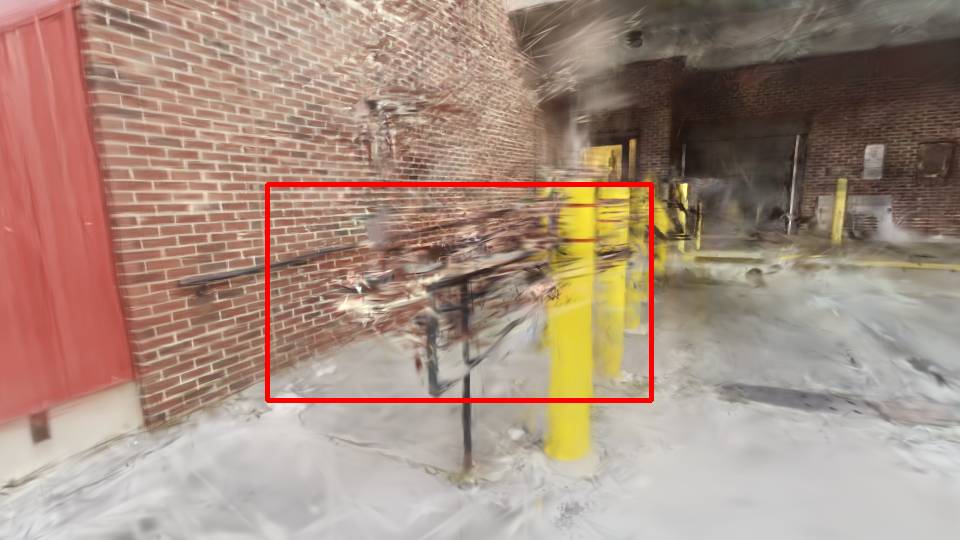}\includegraphics{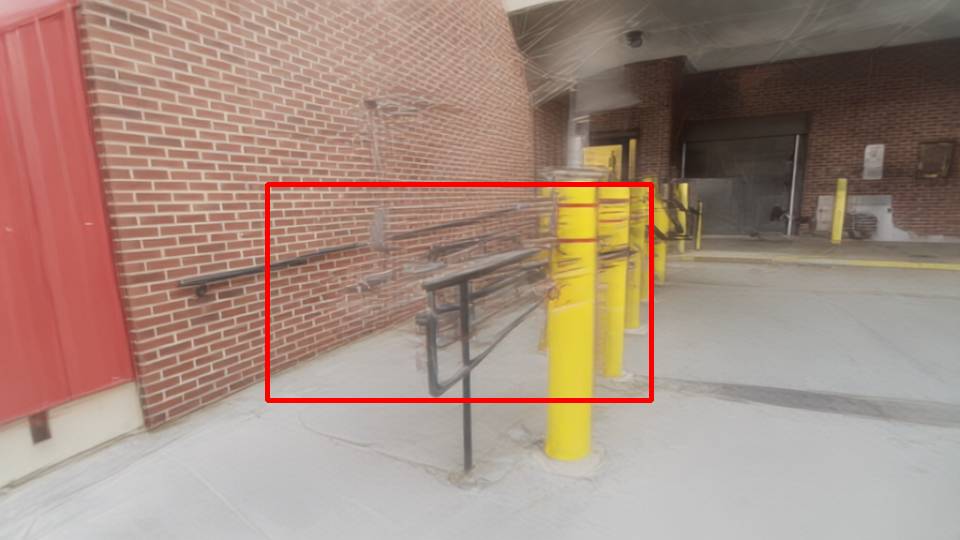}\includegraphics{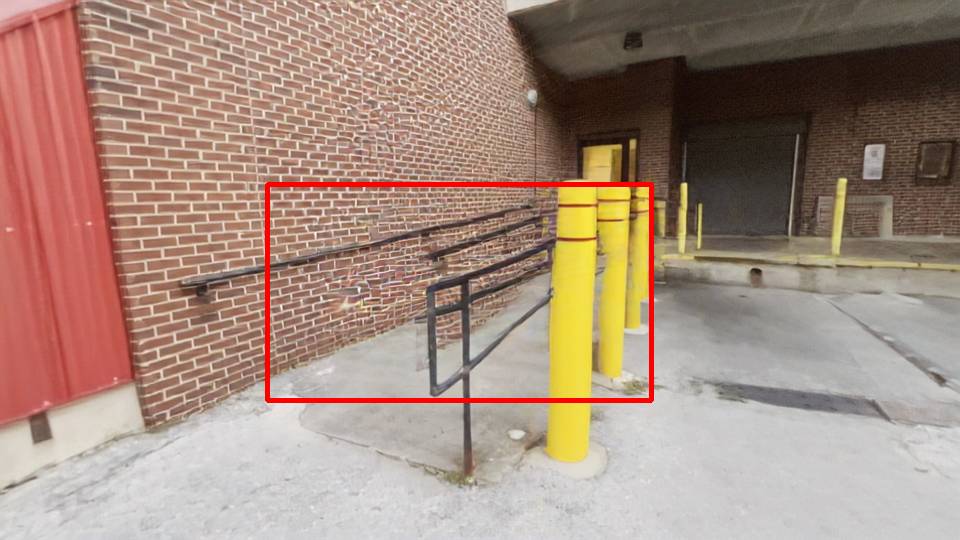}\includegraphics{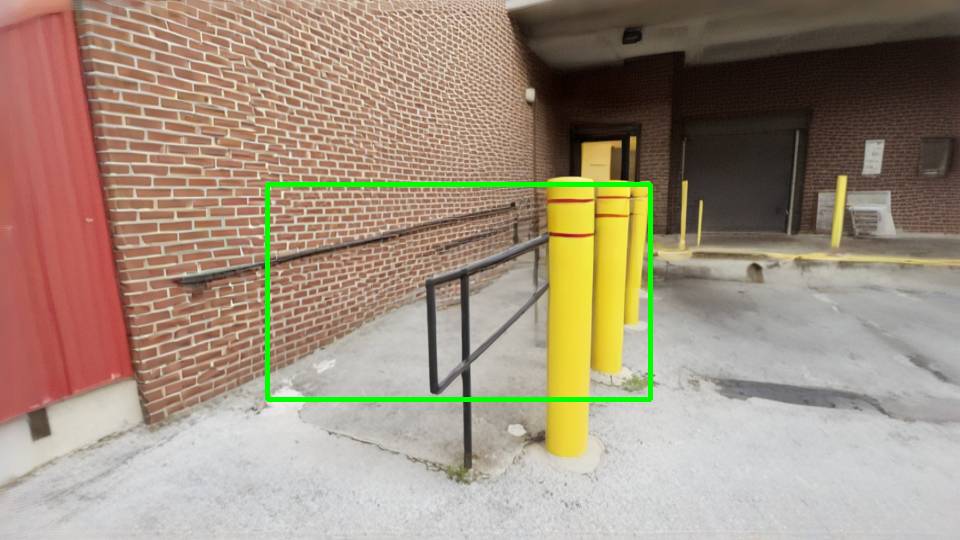}\includegraphics{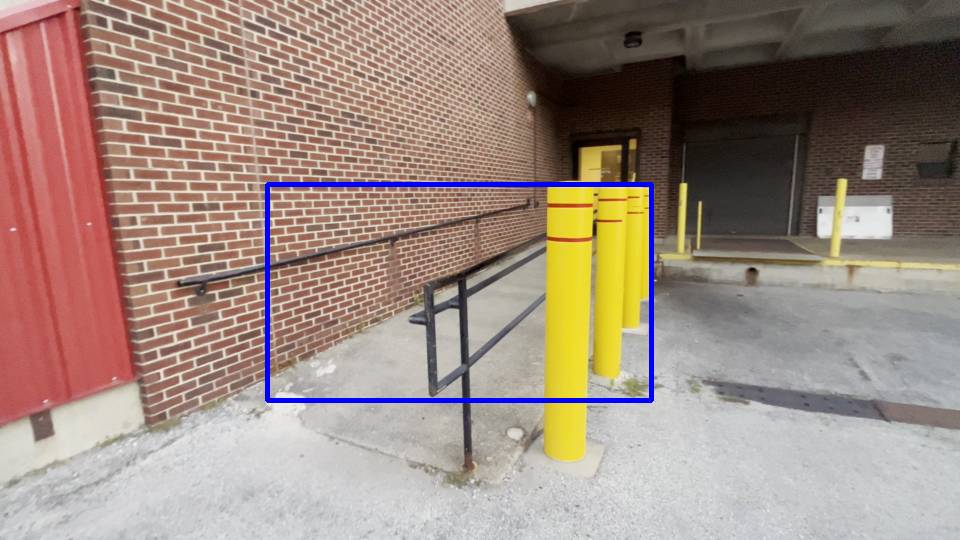}\includegraphics{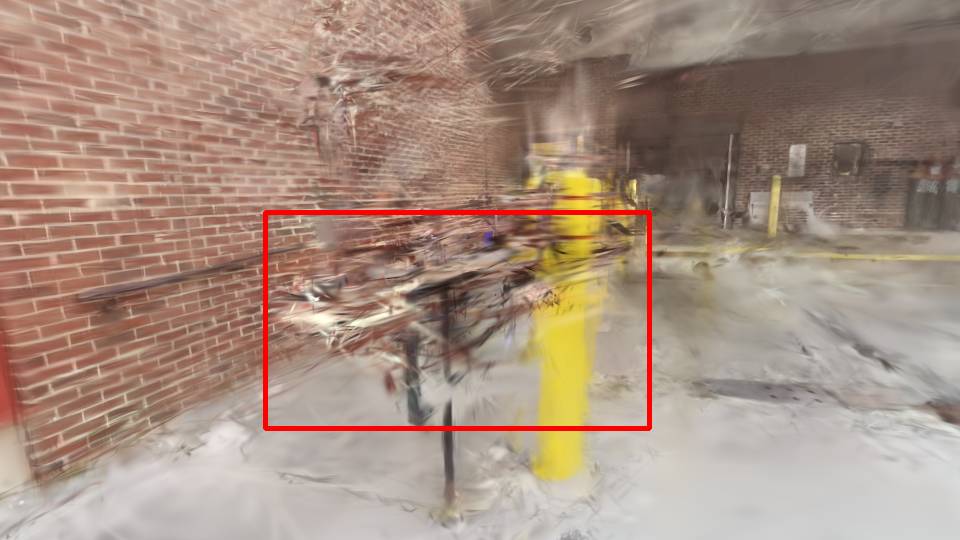}\includegraphics{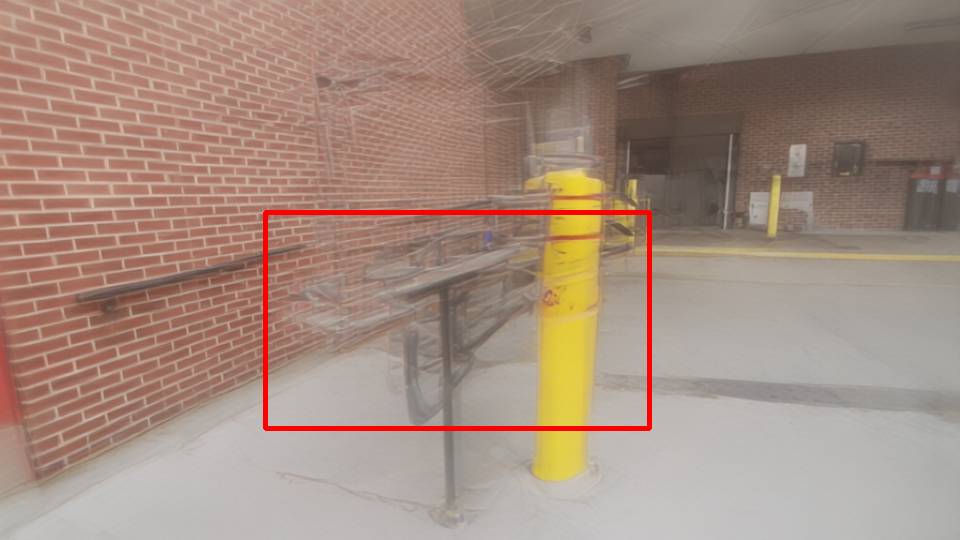}\includegraphics{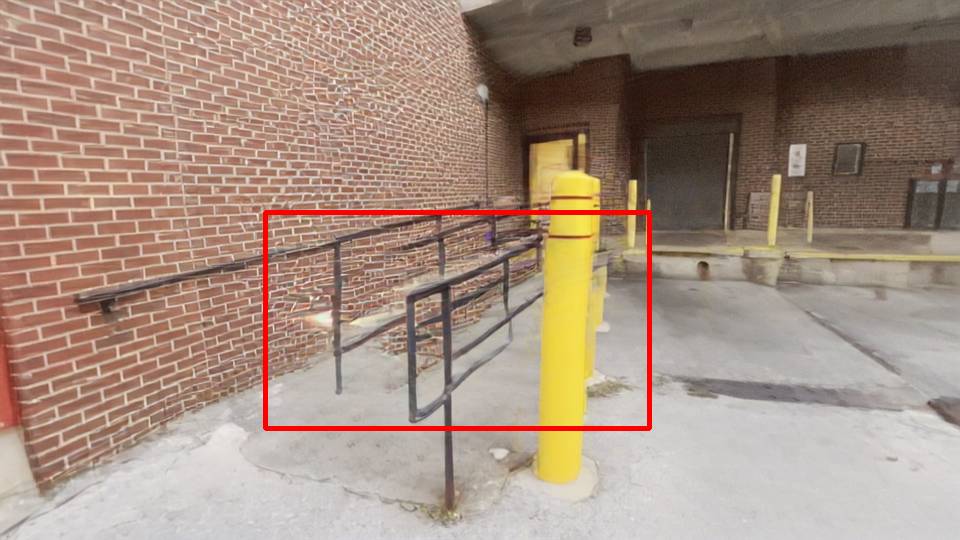}\includegraphics{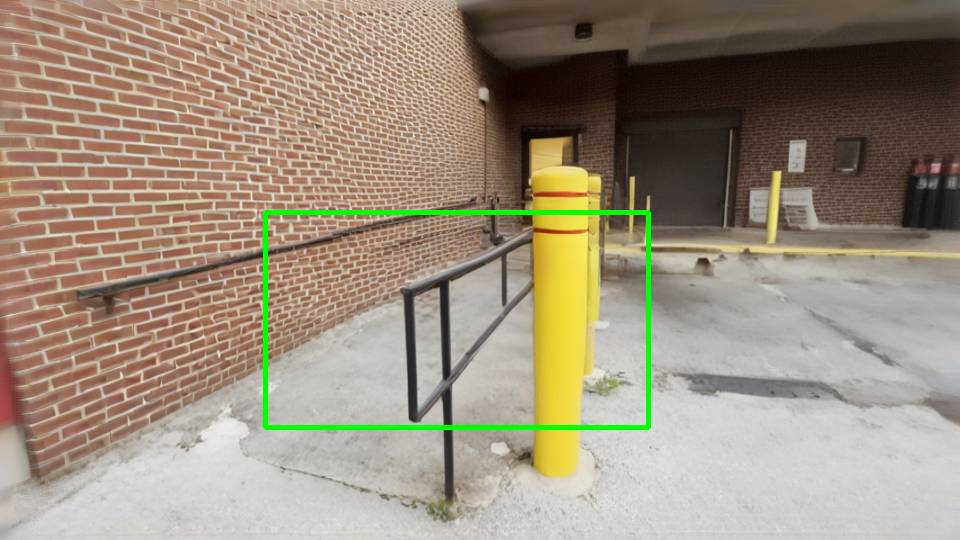}\includegraphics{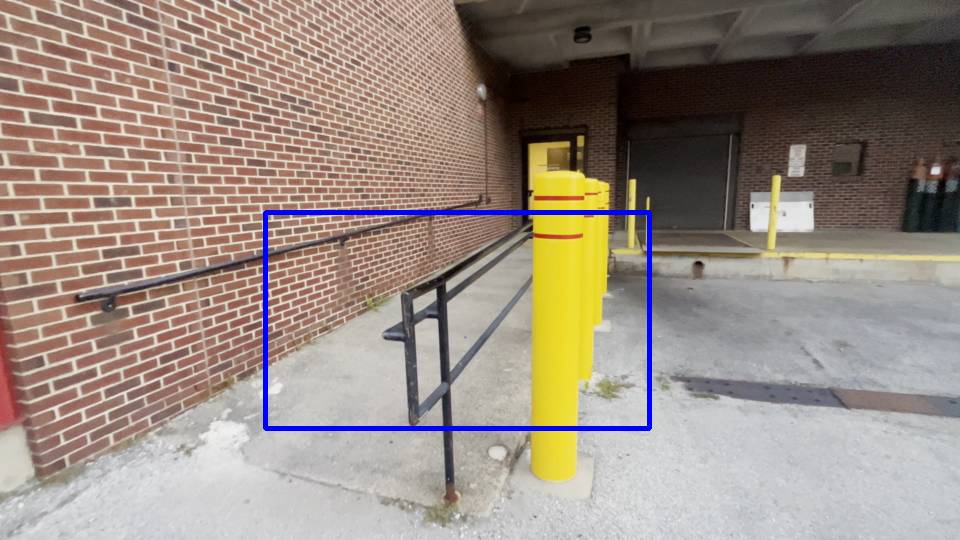}\includegraphics{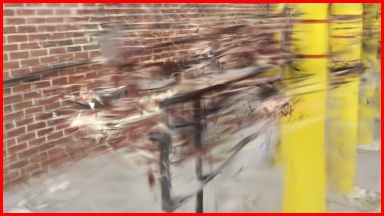}\includegraphics{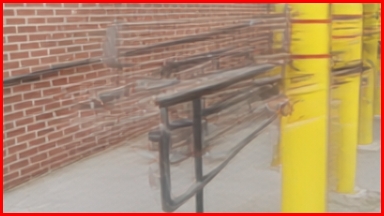}\includegraphics{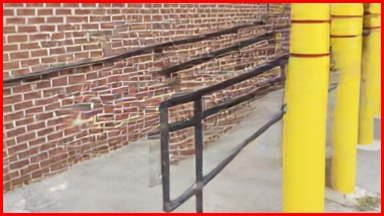}\includegraphics{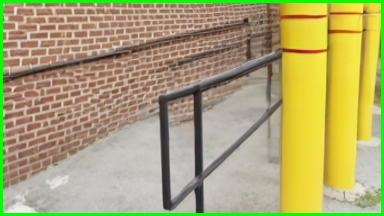}\includegraphics{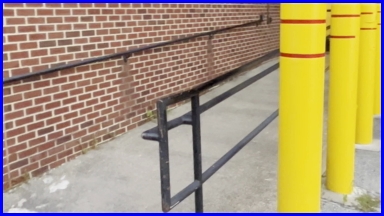}\includegraphics{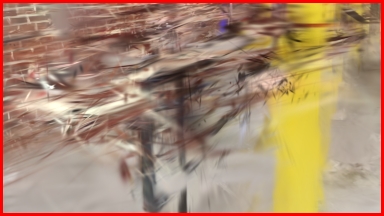}\includegraphics{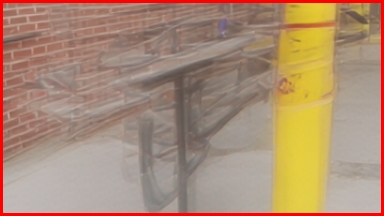}\includegraphics{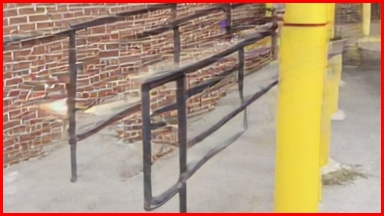}\includegraphics{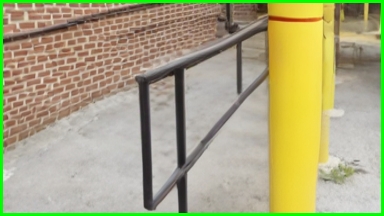}\includegraphics{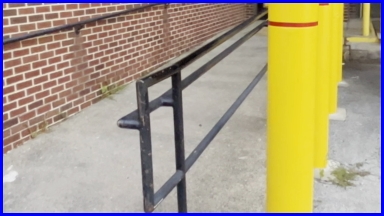}

\includegraphics{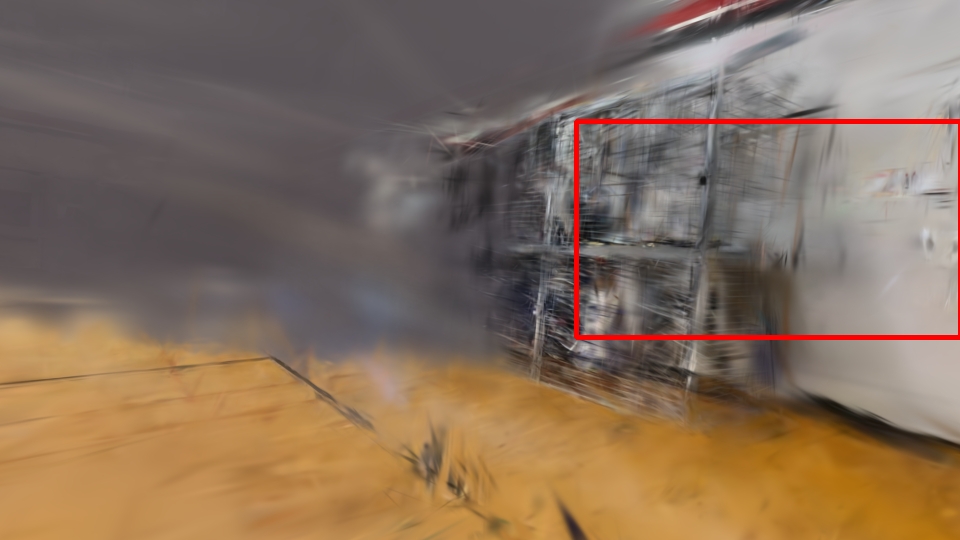}\includegraphics{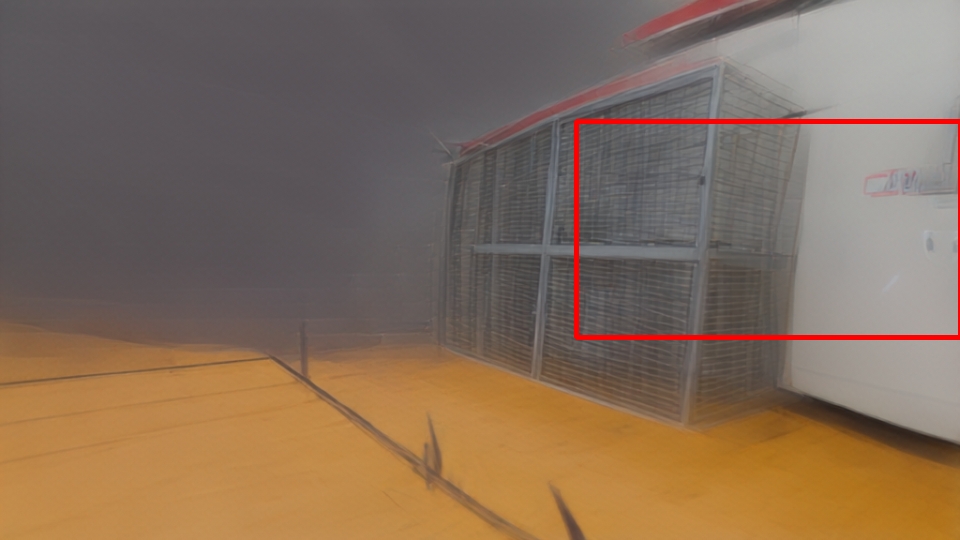}\includegraphics{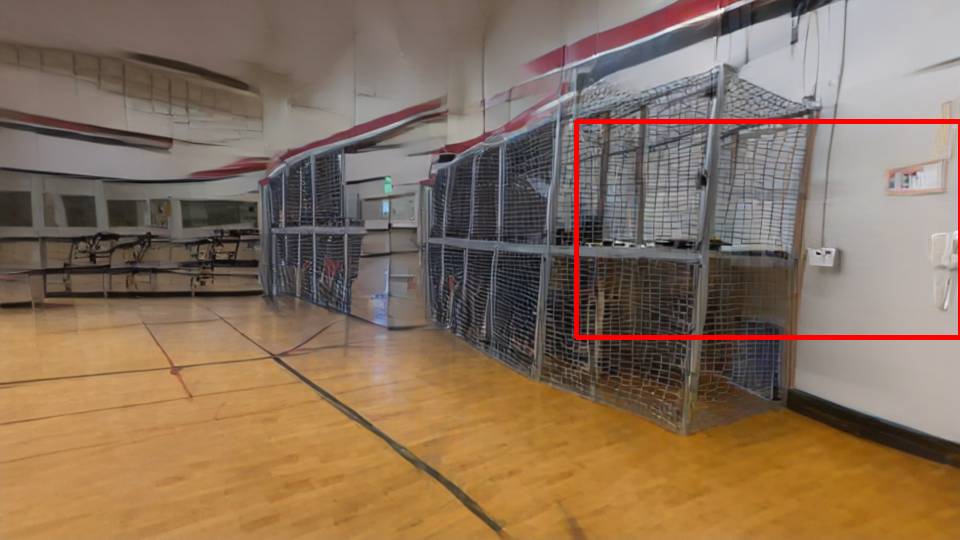}\includegraphics{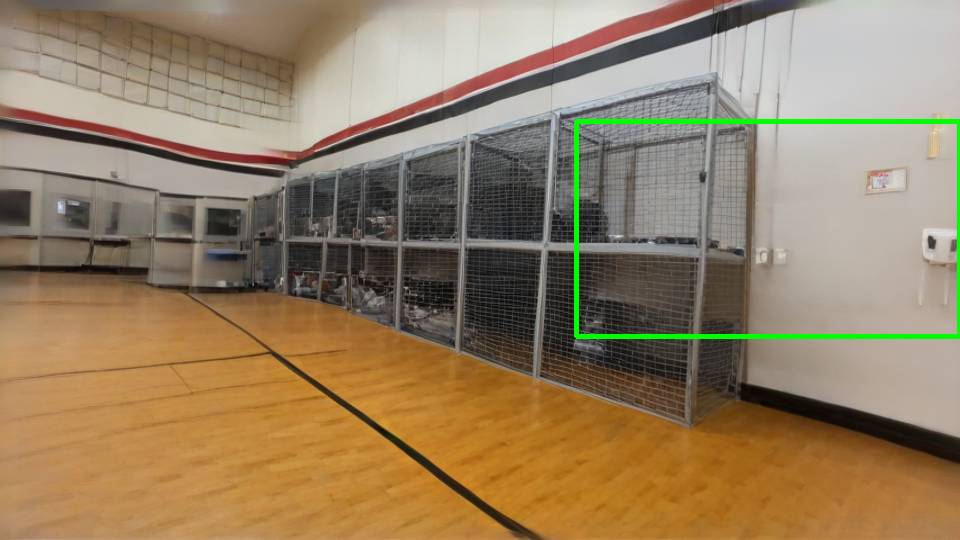}\includegraphics{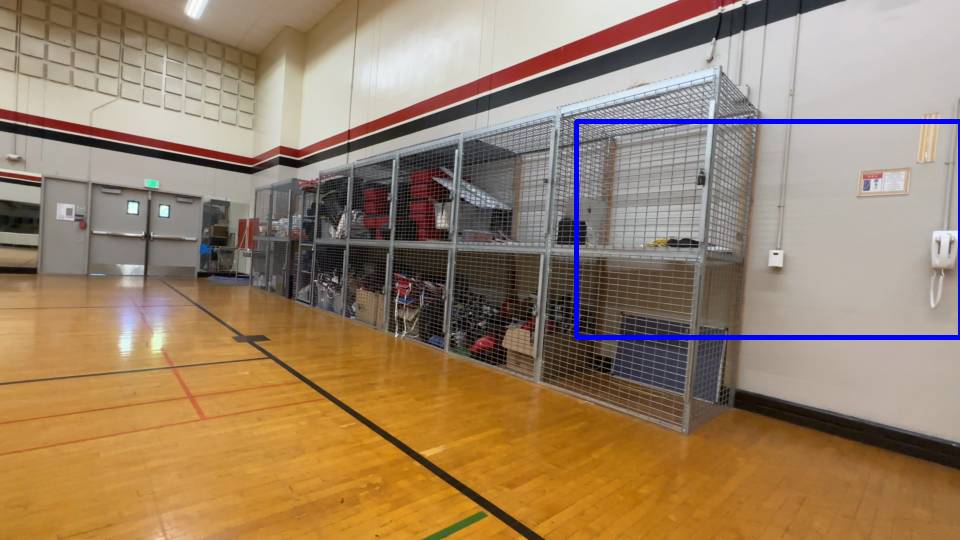}\includegraphics{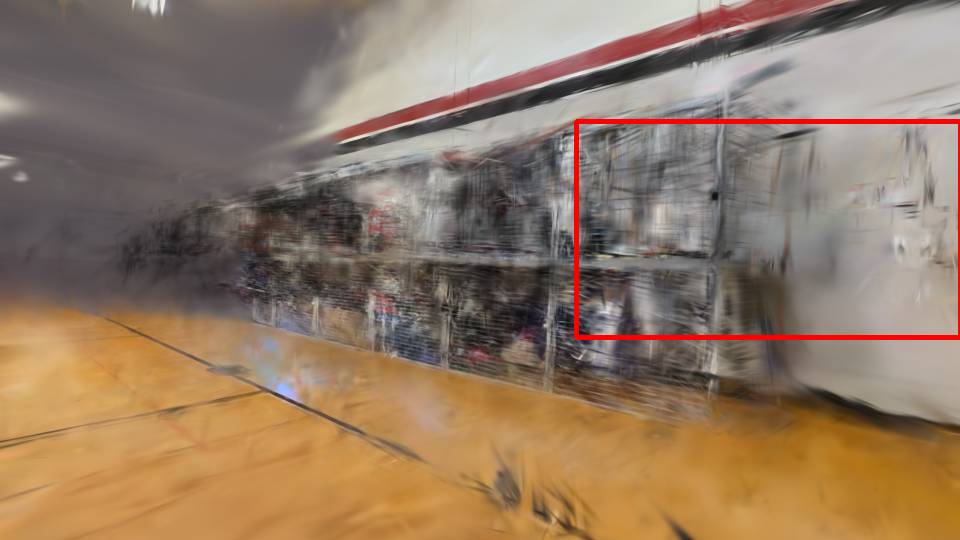}\includegraphics{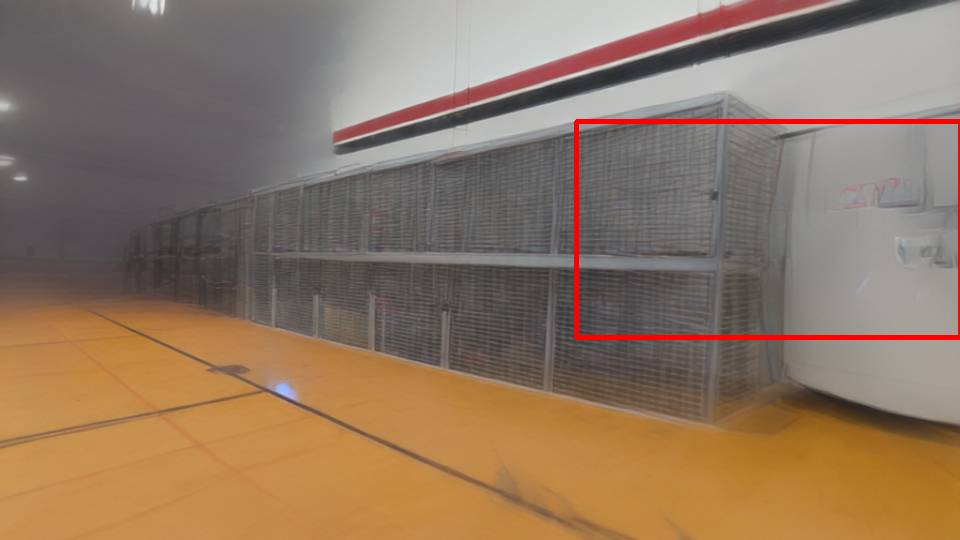}\includegraphics{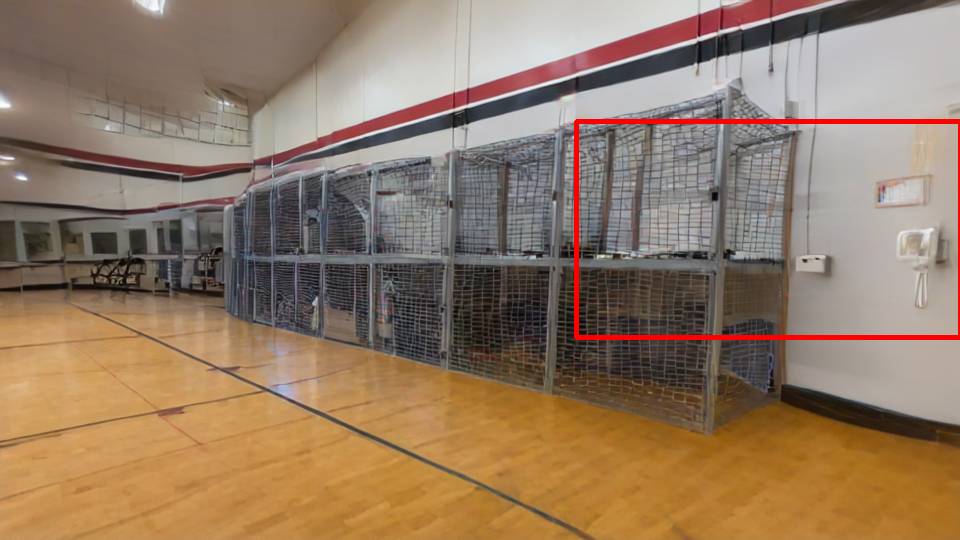}\includegraphics{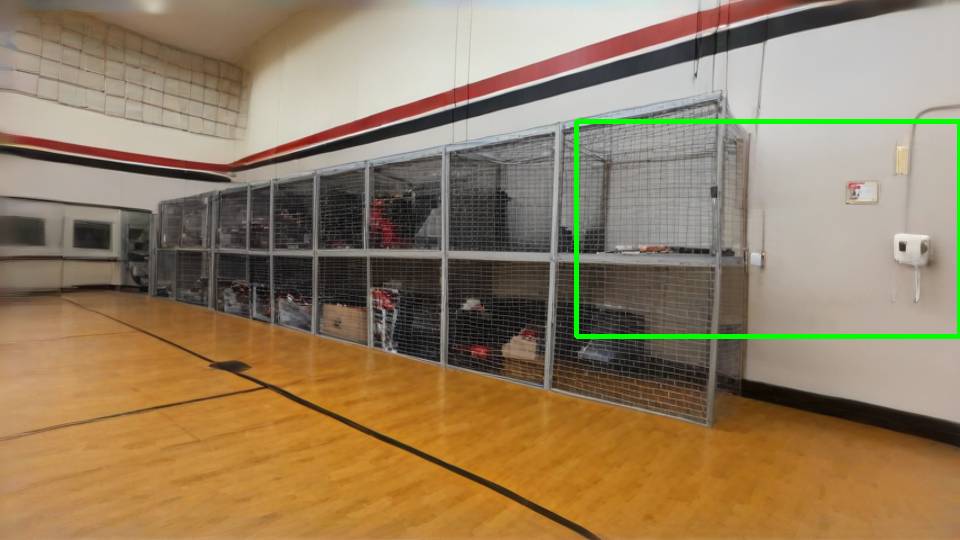}\includegraphics{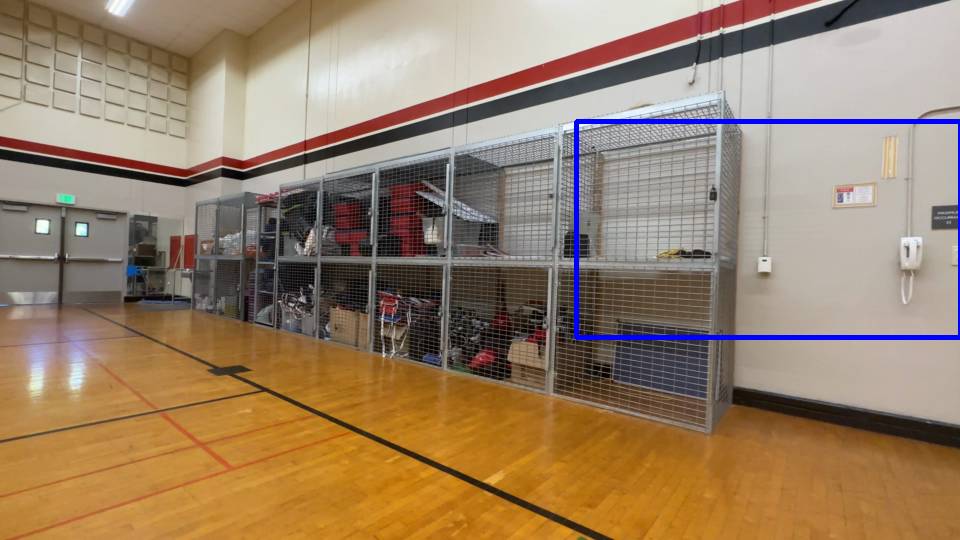}\includegraphics{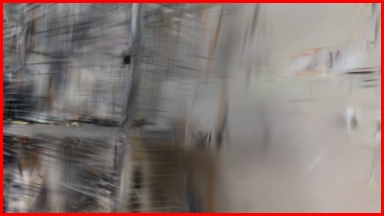}\includegraphics{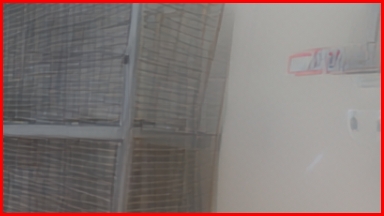}\includegraphics{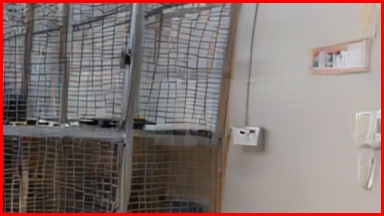}\includegraphics{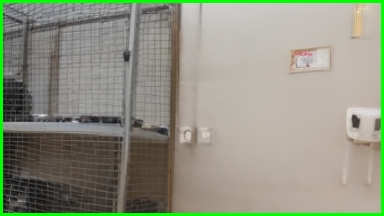}\includegraphics{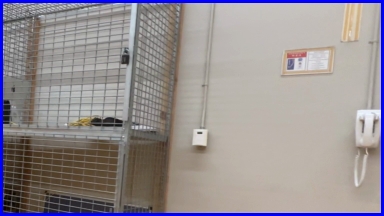}\includegraphics{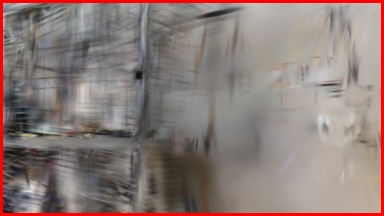}\includegraphics{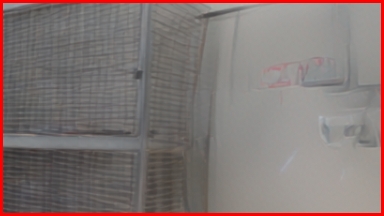}\includegraphics{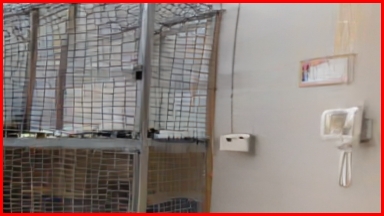}\includegraphics{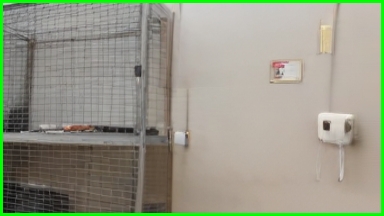}\includegraphics{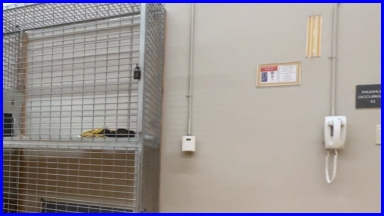}

\includegraphics{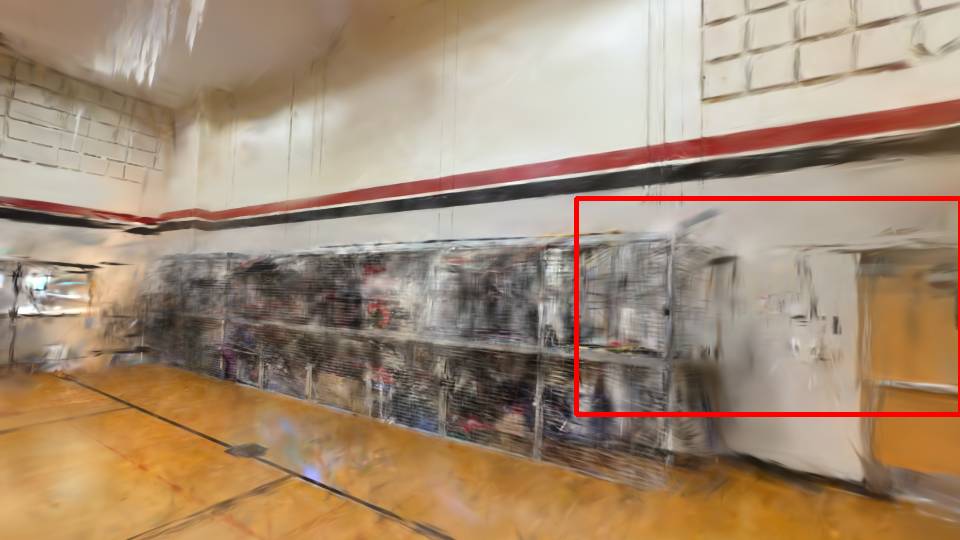}\includegraphics{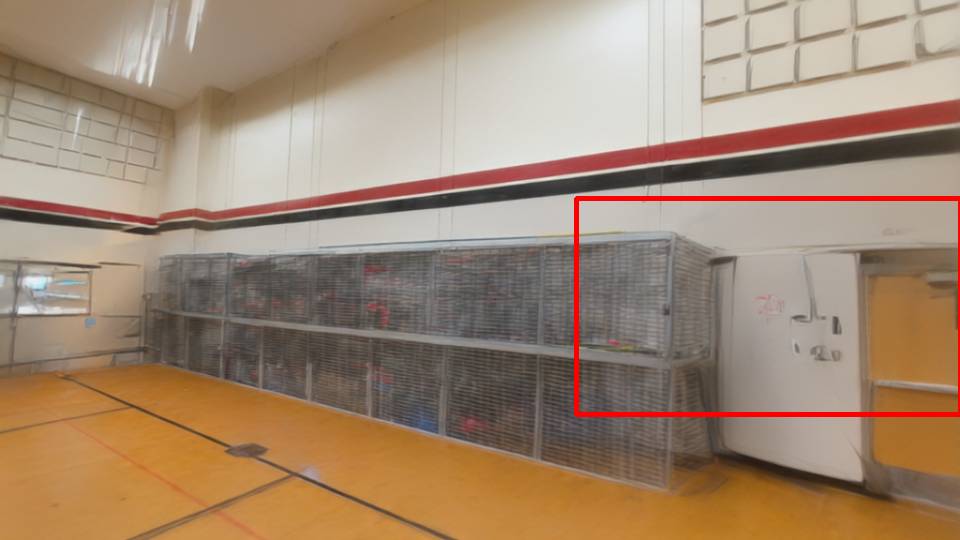}\includegraphics{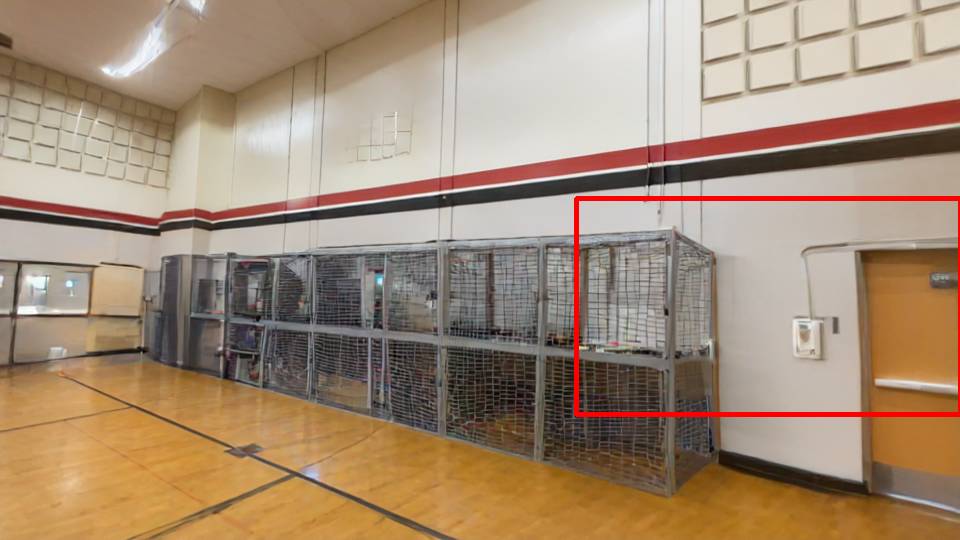}\includegraphics{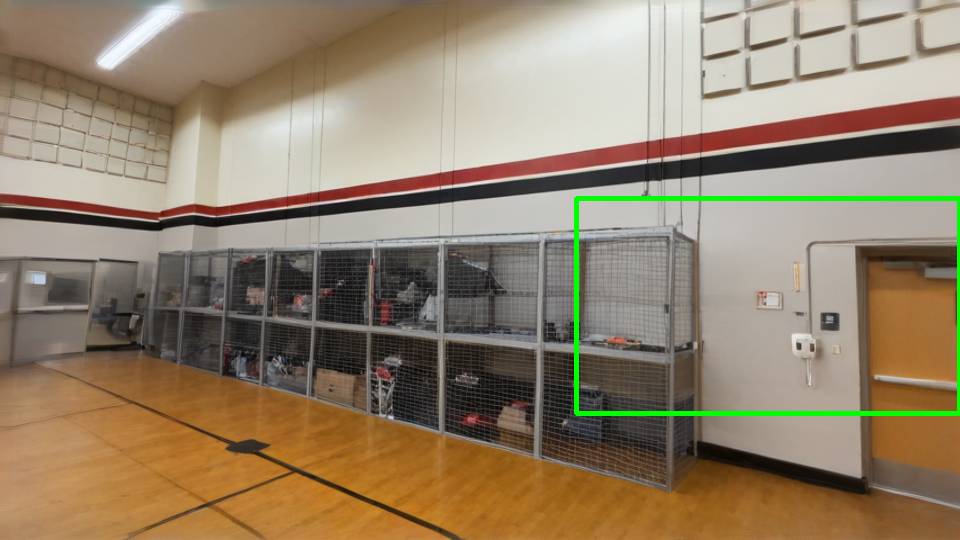}\includegraphics{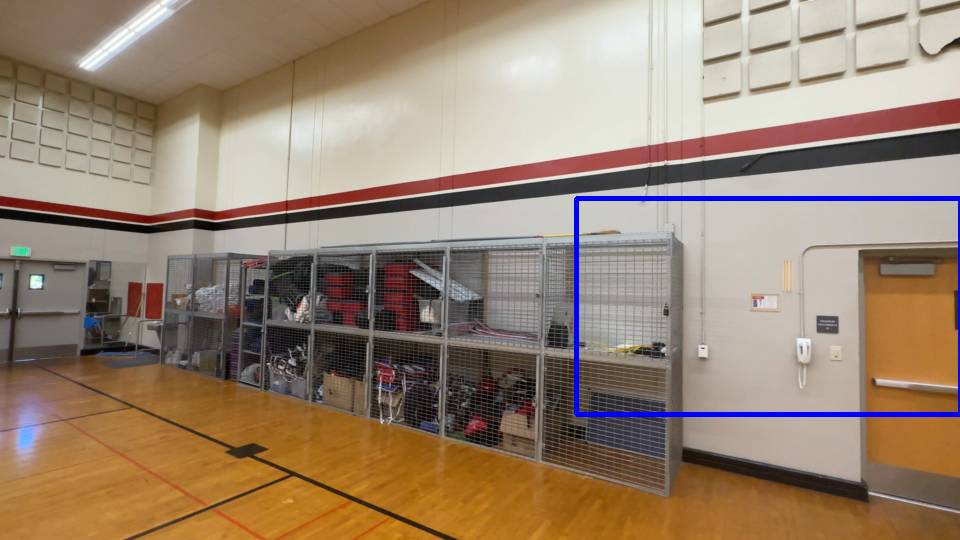}\includegraphics{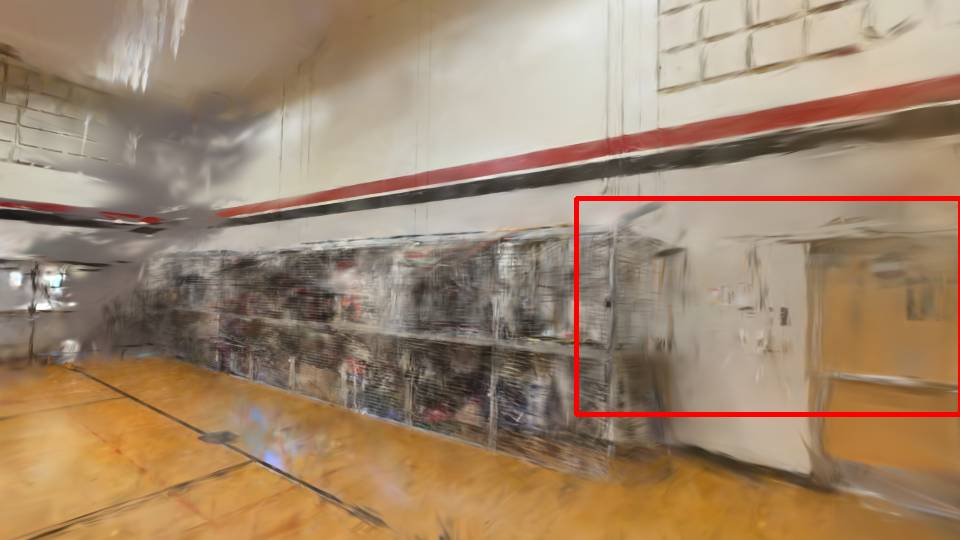}\includegraphics{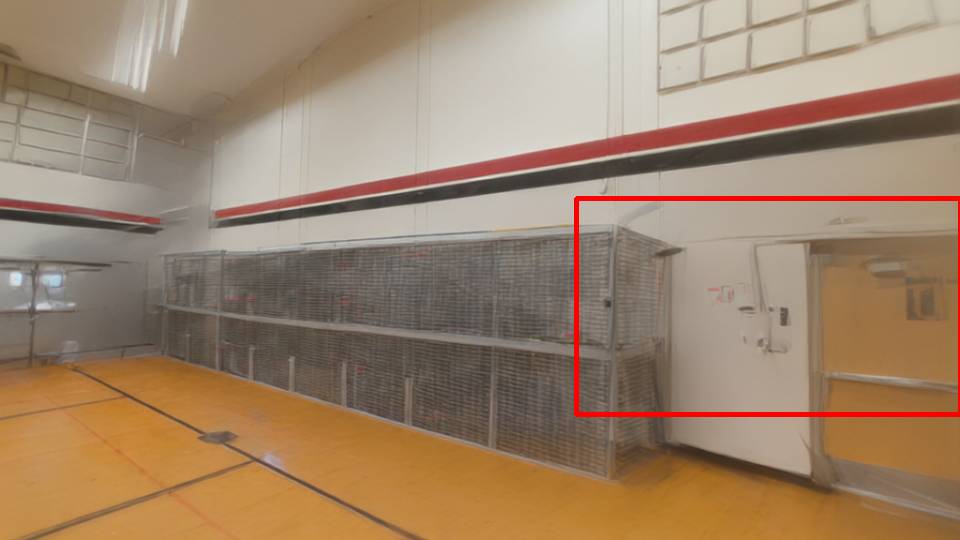}\includegraphics{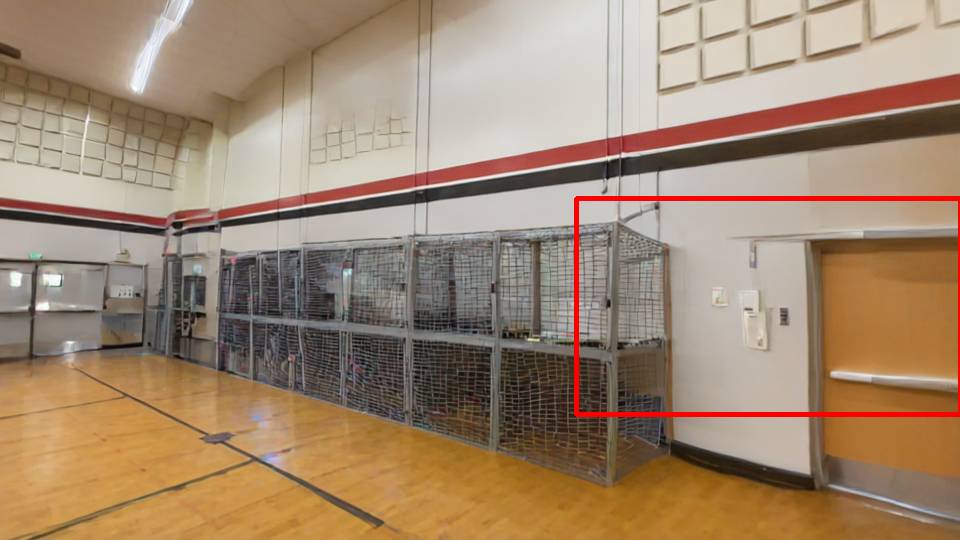}\includegraphics{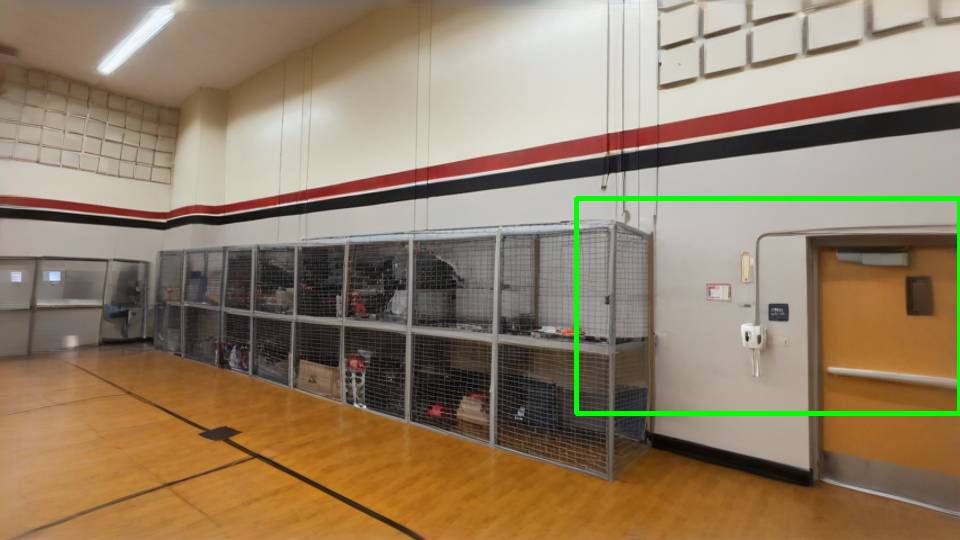}\includegraphics{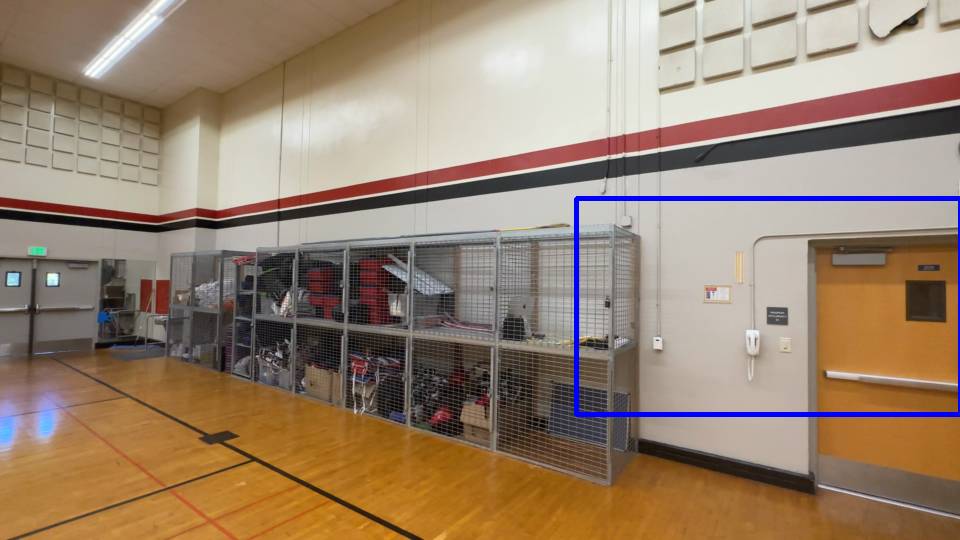}\includegraphics{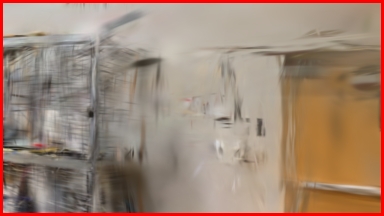}\includegraphics{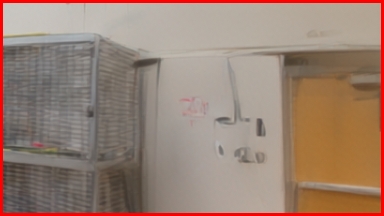}\includegraphics{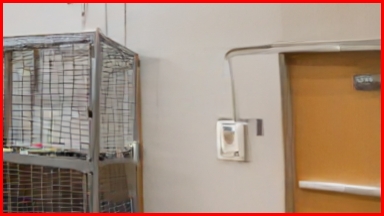}\includegraphics{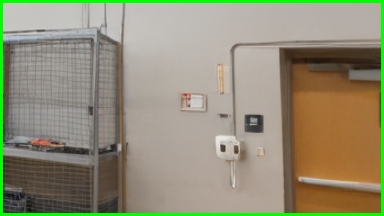}\includegraphics{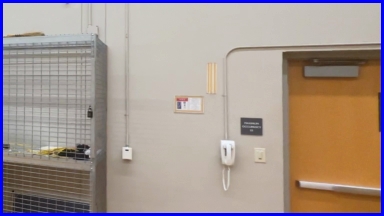}\includegraphics{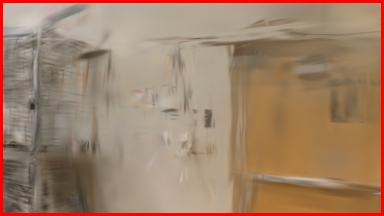}\includegraphics{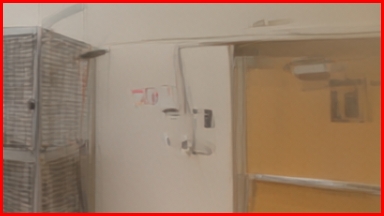}\includegraphics{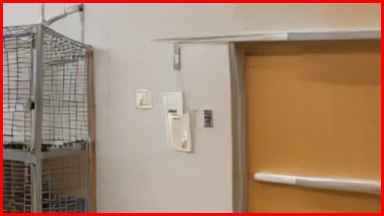}\includegraphics{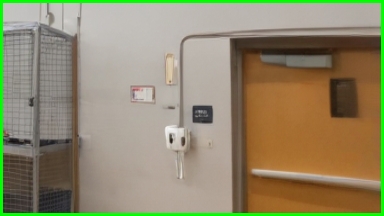}\includegraphics{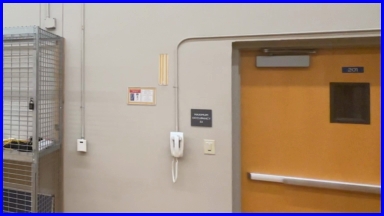}

\FloatBarrier

\bibliographystyle{splncs04}
\bibliography{main}